\documentclass{article}

\usepackage[main,final,preprint]{neurips_2026}

\usepackage[utf8]{inputenc}
\usepackage[T1]{fontenc}
\usepackage{hyperref}
\usepackage{url}
\usepackage{booktabs}
\usepackage{amsfonts}
\usepackage{amsmath}
\usepackage{amssymb}
\usepackage{nicefrac}
\usepackage{microtype}
\usepackage{graphicx}
\usepackage{subfig}
\usepackage{multirow}
\usepackage{tabularx}
\usepackage{array}
\usepackage{xcolor}         
\usepackage{pdflscape}
\usepackage{longtable}

\providecommand{\keywords}[1]{\vspace{0.5em}\noindent\textbf{Keywords:} #1}

\begin{document}

\makeatletter
\providecommand\footref[1]{\textsuperscript{\ref{#1}}}
\makeatother

\title{Robust Ambiguity Detection (RAD)\\
From Model- and Feature-Space Consistency}

%

\author{
  Manya Singh \qquad Mark T. Keane \qquad Arjun Pakrashi \\
  School of Computer Science,\\
  University College Dublin, \\
  Dublin, Ireland \\
  \texttt{manya.singh@ucdconnect.ie, \{mark.keane, arjun.pakrashi\}@ucd.ie} \\
}


\maketitle

\begin{abstract}
 Machine learning models should be \textit{robust}, in the sense of remaining predictively consistent under permissible variations. A model's predictions should ideally remain unchanged when it is replaced by a functionally equivalent one, or when its inputs are subject to minor, admissible perturbations. If such changes alter a prediction significantly, then the prediction is “ambiguous” with respect to the model. Models should abstain from making such ambiguous predictions and/or should flag them for human inspection, especially in high-stakes decision-making scenarios. However, in practice, such ambiguity is not easy to identify once a model is deployed. Here, the \textit{Robust Ambiguity Detection} (RAD) framework is advanced for quantifying predictive ambiguity using two complementary metrics: \textit{Model-Space Consistency} and \textit{Feature-Space Consistency}. These two scores, the RAD Score-Pair, visualised through the RAD Plot, provide an interpretable characterisation of the sources of ambiguity and the actions a user may consider in response. RAD is evaluated on synthetic datasets with systematically controlled overlap, as well as several real-world datasets where the level of ambiguity cannot be directly inspected. Finally, we demonstrate a downstream application of RAD where samples are ranked by their \textit{RAD Pareto-Rank} and the most ambiguous are abstained from prediction, achieving performance comparable to existing rejection-based approaches.
\end{abstract}


\section{Introduction}

Ambiguity occurs when an item has more than one plausible interpretation \cite{klir1987we,frisch1988ambiguity}. In machine learning (ML), ambiguity occurs when a given instance leads to different predictive outcomes across multiple models with nearly identical predictive performance (the so-called Rashomon set \cite{breiman2001statistical}). Several different efforts have been made to measure this \textit{predictive multiplicity} \cite{black2022model,marx2020predictive,ganesh2025systemizing}, casting it as Rashomon capacity \cite{hsu2022rashomon}, Discrepancy \cite{watson2023predictive}, Ambiguity \cite{watson2024predictive} or Arbitrariness \cite{cooper2024arbitrariness}. All these approaches measure \textit{predictive multiplicity} by analysing disagreements among equally accurate \textit{models} on the \textit{same} input instance. We build directly on this formulation, but extend the analysis to the local neighbourhood of an instance, asking whether this disagreement persists under small, permissible variations in the feature instance. In this sense, our contribution is not a redefinition of predictive multiplicity, but an investigation of its interaction with local robustness \cite{zhong2021understanding}.

Consider a concrete example where a university admissions system deploys a set of 10 equally well-performing ML models. For one applicant, Alice, and all $10$ models predict ``accept''. All the models for Alice would still have kept agreeing on ``accept'' even if the GPA was $0.01$ higher or lower. For another applicant, Mike, $5$ models predict ``accept'' while the remaining 5 predict ``reject'', making this instance ambiguous due to disagreement across equivalent models. Model agreements for Mike wouldn't change even if the GPA were a bit higher or lower. Now consider another applicant, Mary, whose credentials differ from Mike’s only by a minimally higher GPA score of $0.01$. For Mary, all 10 models predict ``accept''. Although the two applicants are nearly identical, the collective prediction changes abruptly across an extremely small region of the feature space. Scores based on a single point (e.g., Self-Consistency) cannot distinguish between Mary and Alice, where Mary's predictions are actually unreliable. Also, in the case of Mike, single-point-based scores would not be able to identify if the disagreement was due Mike falling into a "contested" region, or if it is an isolated case. 

In high-stakes settings such as university admissions, if nearly identical applicants receive opposite outcomes solely because the decision boundary passes through a negligibly small region between them, then agreement on the original input alone may be insufficient evidence of predictive reliability. We therefore propose that predictions which fail to remain consistent under small, permissible variations to either the model or the input should be treated as ambiguous and flagged for human review.

In this work, we define ambiguity as a \textit{special case of predictive inconsistency that arises when multiple equivalent models disagree on the prediction of a datapoint, or on equivalent data points in its neighborhood}. This two-dimensional definition captures not only whether models disagree at a point, but whether that agreement (or disagreement) persists under minor, permissible variations to the input (neighbourhood). We use the term \textbf{robust-ambiguity} throughout the text to distinguish our definition from the "ambiguity" metric in \citet{watson2024predictive}. 

\begin{figure}[t!]
    \centering
    \includegraphics[width=0.75\textwidth]{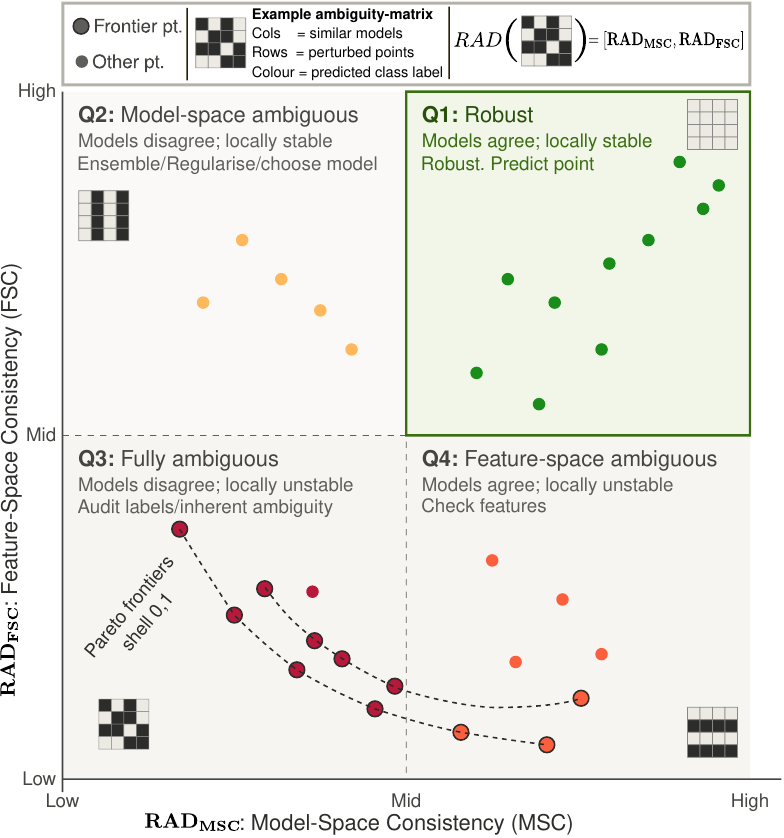}
    \caption{A representation of the RAD Plot, which is a scatter plot of the RAD Score-Pair [$\mathrm{RAD}_{\mathrm{MSC}}$,$\mathrm{RAD}_{\mathrm{FSC}}$] representing \textit{robust-ambiguity} by jointly capturing Model-Space Consistency (MSC) and Feature-Space Consistency (FSC). Each inset pictograph of a matrix represents a typical ambiguity-matrix of a single datapoint $t$, from which the RAD Score-Pair is computed. Each (j,i)$^{th}$ location of an ambiguity matrix is the predicted class of the $i^{th}$ equivalent model on the $j^{th}$ neighbourhood point of $t$. Each quadrant uncovers specific properties, based on which potential actions could be taken. Pareto-optimal frontiers rank points for downstream tasks.}
    \label{fig:demo_RAD}
\end{figure}

To achieve this, we propose a general framework, Robust Ambiguity Detection (RAD), which has the following contributions:

\begin{itemize}
    \item A three-stage \textbf{RAD framework} introduced in Section \ref{sec:rad}, that, for any test datapoint generates a set of equivalent models, a set of local perturbations of the datapoint, and a matrix of predictions across both, named the \textit{ambiguity matrix}.
    
    \item A \textbf{RAD Score-Pair} [$\mathrm{RAD}_{\mathrm{MSC}}$,$\mathrm{RAD}_{\mathrm{FSC}}$] (Section \ref{sec:rad_impl}), which reduces the ambiguity matrix into two chance-corrected agreement scores, namely, Model-Space Consistency (MSC) and Feature-Space Consistency (FSC). The RAD Score-Pair quantifies the disagreement in each space. A ranking method \textbf{RAD-Pareto-Rank} (Section \ref{subsec:radrank}) is also introduced which rank orders datapoints based on robust-ambiguity for downstream tasks.
    
    \item A \textbf{RAD Plot} (Section \ref{sec:rad_interpret}) that visualises the RAD Score-Pair (Fig.~\ref{fig:demo_RAD}) as four interpretable quadrants, each with specific properties, allowing a practitioner not only to flag ambiguous predictions but to diagnose why they are ambiguous.
\end{itemize}

Once a model is trained for a specific problem using a given dataset, the worldview of the model is always with respect to the decision boundary. Therefore, the RAD Score-Pair is always relative to the models. Therefore, RAD scores do not represent absolute properties of the data or its underlying distribution, but rather identify which predictions cannot be made reliably by the specific class of models.

We demonstrate RAD on synthetic datasets with systematically controlled class overlap, as well as apply it to several real-world datasets from the UCI ML Repository (Section \ref{sec:results}). Using these datasets, we also demonstrate a downstream abstention task, where we abstain predicting ambiguous datapoints based on RAD Pareto-Rank (Section \ref{subsec:result_abstain}). Section \ref{sec:deployment}  discusses some potential applications of RAD.  Section \ref{sec:conclusion} concludes the paper.

\section{Related Work}

There is an emerging group of metrics that have been developed to measure \textit{predictive multiplicity}, i.e. the conflicting predictions from multiple equally competent models on a given input. These measures are typically some function of model disagreement, either through entropy, a simple voting mechanism, or others. Broadly, they fall into the umbrella of measuring model-space disagreements. 

\textit{Rashomon Capacity} \cite{hsu2022rashomon} quantifies predictive multiplicity for probabilistic classifiers by measuring the spread of predicted score distributions across the Rashomon set. It is formally defined as the channel capacity (under KL divergence) of the multiplicity set of output distributions for a sample. It applies to probabilistic outputs rather than thresholded class labels, and like other point-based metrics, it does not measure whether the score variation persists in the sample's local neighbourhood. Authors in \cite{watson2023predictive,watson2024predictive} defined the metric \textit{ambiguity} as the proportion of samples with diverging predictions among the equivalent model set, and \textit{obscurity} \cite{cavus2024experimental} extended this to count how many models disagree with the majority prediction. All three operate at a fixed point.

Separately, \textit{local robustness} \cite{zhong2021understanding,leino2021relaxing,summers2021nondeterminism} studies the consistency of a single model’s predictions within a local neighbourhood of an input. A single model can be vulnerable to perturbations as small as one pixel \cite{huang2025one}, consequences of which can be undesirable in high-stakes settings. This captures \textit{feature-space disagreement} with respect to one model, but ignores model-space variation.

\textit{Arbitrariness} \cite{cooper2024arbitrariness} is defined as the variance in predictions across equally good models for a single datapoint, occurring by chance. The associated \textit{self-consistency (SC)} metric is the average pairwise agreement among equivalent models. This aligns with our goal of flagging unreliable predictions, but operates in one dimension (model-space) and at one point. The mechanism in \cite{cooper2024arbitrariness} decomposes how much disagreement is attributable to random chance, while our work instead measures agreement (chance-corrected) jointly across model- and feature-space. 

Because arbitrariness and robust-ambiguity operate in different domains and measure different quantities, they are complementary. A complete reliability assessment requires measuring both, which is what RAD provides through the RAD Score-Pair.

Single model uncertainty scores (softmax probabilities, distance to SVM margin, leaf-node class distribution \cite{hastie2009elements}) are widely used as proxies for reliability, although they neither reflect epistemic uncertainty nor model disagreement \cite{gawlikowski2023survey,hullermeier2021aleatoric}, but only describe the behaviour of a single chosen model. They cannot capture instability arising from predictive multiplicity or local robustness across an equivalent model set.

Ambiguity has been operationalised for abstention in \cite{watson2023predictive,watson2024predictive,cooper2024arbitrariness,rudin2024amazing}. However, a datapoint flagged as locally robust with respect to one model may not be locally robust under another equivalent model. Abstention based on a single model does not guarantee reliability when other equivalent models would respond differently. A joint consideration of all equivalent models across a neighbourhood around the target point is required before flagging for abstention, which RAD provides.

\section{RAD Framework Overview}\label{sec:rad}
The Robust Ambiguity Detection (RAD) framework measures \textit{robust-ambiguity} by computing disagreement among equivalent models, in the model-space, under local perturbations of a test datapoint, in the feature-space. It is model-agnostic and can be adapted to the data type and the data domain at hand. The framework has three stages; we describe them generically here with a concrete instantiation in Section \ref{sec:rad_impl}.

\paragraph{Stage 1: Generating a set of similar models.}
A set of $n$ models $\mathcal{M}^{(m)}$ with nearly equal prediction performance, produced by a method $m$, but varying in model hyperparameters, bootstrapped training dataset, using different ML algorithms, different seeds, dropout masks, different algorithm families, etc. This is referred to as \textit{equivalent models} (a sample of the broader    Rashomon set), because there is no clear basis to prefer one over another.

\paragraph{Stage 2: Generating a set of similar data points.}
For a test datapoint $\mathbf{t}\in\mathbb{R}^{d}$, generate a set  $\mathcal{T}_{\mathbf{t}}=\{\mathbf{t}_{1},\mathbf{t}_{2},\mathbf{t}_{3},\ldots,\mathbf{t}_{p}\}$ of $(p-1)$ synthetic \textit{neighbourhood points} (with $\textbf{t}_{1}=\textbf{t}$) in its local neighborhood so that the stability of predictions in that region can be quantified.

\paragraph{Stage 3: Quantifying robust-ambiguity.}
Every model in $\mathcal{M}^{(m)}$ will make predictions on $\mathcal{T}_{\mathbf{t}}$, resulting in a two-dimensional \textit{ambiguity matrix}, $\mathcal{A}_{\mathcal{T}_{\mathbf{t}}}^{\mathcal{M}^{(m)}}$ of size $p \times n$ (where $|\mathcal{M}^{(m)}| = n$ and $|\mathcal{T}_{\mathbf{t}}| = p$) with entry $(j,i)$ containing $\mathcal{M}^{(m)}_{i}(\mathbf{t}_{j})$, the prediction of model $i$ on perturbation $j$:

\begin{equation}
\mathcal{A}_{\mathcal{T}_{\mathbf{t}}}^{\mathcal{M}^{(m)}} =
\begin{bmatrix}
\mathcal{M}^{(m)}_1(\mathbf{t}_1) & \mathcal{M}^{(m)}_2(\mathbf{t}_1) & \dots & \mathcal{M}^{(m)}_n(\mathbf{t}_1) \\
\mathcal{M}^{(m)}_1(\mathbf{t}_2) & \mathcal{M}^{(m)}_2(\mathbf{t}_2) & \dots & \mathcal{M}^{(m)}_n(\mathbf{t}_2) \\
\vdots & \vdots & \ddots & \vdots \\
\mathcal{M}^{(m)}_1(\mathbf{t}_p) & \mathcal{M}^{(m)}_2(t_p) & \dots & \mathcal{M}^{(m)}_n(\mathbf{t}_p)
\end{bmatrix}
\label{eq:amb_matrix}
\end{equation}

The ambiguity matrix for $\mathbf{t}$ is then reduced to the \textbf{RAD Score-Pair [$\mathrm{RAD}_{\mathrm{MSC}}$,$\mathrm{RAD}_{\mathrm{FSC}}$]}, where $\mathrm{RAD}_{\mathrm{MSC}}$ (Model-space Consistency) measures inter-model agreement across the neighbourhood, and $\mathrm{RAD}_{\mathrm{FSC}}$ (Feature-Space Consistency) measures intra-model stability across the neighbourhood. For this work, the specific reduction we use is given in Section \ref{subsec:stage3}. This RAD Score-Pair can then be used to decide the degree of ambiguity of a test datapoint, used in downstream tasks, or the nature of the ambiguity can be inspected using the \textbf{RAD Plot} (an example plot is shown in Figure \ref{fig:demo_RAD}), which is a scatter plot of the RAD Score-Pair interpreted based on which quadrant the datapoints fall.

\section{RAD Framework Implementation}\label{sec:rad_impl}

To implement RAD for a specific problem, we need to decide, (Stage 1) how to generate a the set of \textit{equivalent models}, $\mathcal{M}^{(m)}$, (Stage 2) how to generate the \textit{neighbourhood points} set, $\mathcal{T}_{\mathbf{t}}$, of a given test datapoint $\mathbf{t}$, and (Stage 3) how to compute the ambiguity Score-Pair by measuring the degree of disagreement in the ambiguity matrix $\mathcal{A}_{\mathcal{T}_{\mathbf{t}}}^{\mathcal{M}^{(m)}}$ of a datapoint $\mathbf{t}$.
In this paper, we focus on demonstrating RAD on three types of datasets with real-numbered attributes. Synthetic datasets, real-world datasets from the UCI ML Repository \cite{asuncion2007uci}, and one image dataset  (MNIST  \cite{deng2012mnist}).
This section will show an implementation of all three stages, and also introduce the usefulness of the RAD Plot as an interpretable diagnostic tool.

\subsection{Stage 1: Generate Equivalent Models in the Model-space}

In our work, \textit{robust-ambiguity} is defined with respect to both, the model-space and feature-space. Therefore, the ambiguity of a datapoint depends on how the models are being trained. In this implementation we select the Decision Tree (DT) \cite{breiman2017classification}  algorithm. Firstly, the best hyperparameter is found for the algorithm (DT) (through cross-validation experiments). This chosen hyperparameter is used as a "blueprint" to generate the set of similar models $\mathcal{M}^{(m)}$.

Next, using the chosen hyperparameters, we train the set of similar models $\mathcal{M}^{(DT+BS)}$, which is generating $b$ number of DT models, each trained on a bootstrap sample (BS) of the training partition of the dataset.

The method to generate $\mathcal{M}^{(m)}$ depends on the specific application. If a problem demands simulation of a set of models that can allow higher levels of disagreement, i.e. find a more sensitive measure of ambiguity, bootstrap sampling can be used. Otherwise, if minimal stochastic level disagreement is to be measured, seed sampling can be used (\textit{underspecification} set\cite{rudin2024amazing}). That is, when the only source of variation among the models is the initial seed value used to train them.  Alternatively, when using neural networks or other deep learning networks, a unique dropout mask \cite{kendall2017uncertainties} can be used at the final layer to create the set of equivalent models, or Monte-Carlo Dropout \cite{gal2016dropout} can be used to generate multiple predictions. Ideally, the model set should effectively contain multiple models with similar performance that cannot be clearly differentiated.

\subsection{Stage 2: Generating Neighbourhood Datapoints in the Feature-space}\label{subsec:stage2}

Local robustness is the property of a model to produce consistent, robust predictions in the local neighbourhood of a query data point, or when the input is slightly modified \cite{zhong2021understanding}. If a sample lies close to the decision boundary, such a consistent prediction is less likely, as a minor change in the input features can push it past the decision boundary, making that data point vulnerable \cite{leino2021relaxing}. It is beneficial to then measure the prediction stability around a query data point, either within a hyper sphere or the underlying data manifold.
It is important to note here that the type of perturbations of $\mathbf{t}$ we use to assess the models' local robustness will inform the type of robustness we measure. For example, in this instance, we do not specifically use adversarial perturbations, therefore we intend to measure robustness under any perturbation, and not specifically adversarial robustness \cite{gojic2023non}.

For a test datapoint $\mathbf{t}$, a SMOTE-style linear interpolation sampling rule was implemented to generate similar synthetic data points to generate $\mathcal{T}_{\mathbf{t}}$ as follows
\begin{multline}
\label{eq:smote_like}
    \mathcal{T}_{\mathbf{t}} = \{S(\mathbf{t}, \mathbf{x}_{r_{i}}, \lambda_{i}) \;|\; r_{i} \sim \mathcal{U}\{1,\ldots,k\}, \mathbf{x}_{r_{i}} \in \mathrm{NN}_{k}(\mathbf{t},\mathbf{D}),\; \lambda \sim \mathcal{U}(0,1)\}_{i=1}^{p}
\end{multline}

where $S(\mathbf{t}, \mathbf{x}, \lambda) = \mathcal{D}(\mathcal{E}(\mathbf{t}) + \lambda(\mathcal{E}(\textbf{x})-\mathcal{E}(\mathbf{t})))$. Here, $\mathcal{E}$ is an encoder which transforms the feature space into a latent space, and $\mathcal{D}$ is a decoder which brings back the latent representation to the original feature space. $\mathrm{NN}_{k}(\mathbf{t},\mathbf{D})$ is the $k$ nearest neighbours of $\mathbf{t}$ with respect to the training dataset $\mathbf{D}$. If sampling directly from the original feature space, then $\mathcal{E}(\mathbf{x})=\mathcal{D}(\mathbf{x})=\mathbf{x}$, which we do for the synthetic dataset and UCI ML datasets. For the MNIST dataset, we perform this sampling in a latent space learned by an autoencoder in this case  \cite{rumelhart1985learning}.

Unlike the original SMOTE \cite{chawla2002smote} algorithm, which generates synthetic samples within the same minority class to oversample that class, our approach considers nearest neighbours regardless of class. This allows perturbations to occur either between classes or within the same class, depending on the location of the test point. If $\textbf{t}$ lies near a decision boundary, then the perturbations are more likely to span multiple classes, increasing the neighbourhood disagreement in the ambiguity-matrix. If it lies far from the boundary, the perturbations are more likely to remain within the same class. Furthermore, models with complex decision boundaries are more sensitive to minor perturbations of datapoints near the decision boundary and, consequently, more prone to ambiguity \cite{black2022model}.

\subsection{Stage 3: Compute RAD Score-Pair from the Ambiguity-Matrix}\label{subsec:stage3}
In this stage, the predictions of $\mathcal{T}_{\mathbf{t}}$ by the set of models $\mathcal{M}^{(m)}$ are used to generate the ambiguity matrix $\mathcal{A}_{\mathcal{T}_{\mathbf{t}}}^{\mathcal{M}^{(m)}}$ (Eq.~\eqref{eq:amb_matrix}).   The matrix is then reduced to the \textit{RAD Score-Pair}, which captures robust-ambiguity across two dimensions.
\subsubsection{Choosing Disagreement Metric}
To quantify the level of agreement among the \textit{equivalent models} in the \textit{neighbourhood} of $\mathbf{t}$ in the ambiguity-matrix, we use Gwet's Agreement Coefficient AC1~\cite{gwet2014handbook}. There are several other choices to compute such agreement scores. Cohen's Kappa \cite{cohen1960coefficient} is limited to two raters, averaging pairwise scores compounds chance-correction errors. Fleiss' Kappa \cite{fleiss1971measuring} handles multiple raters but suffers from the high-agreement paradox - paradoxically low scores under skewed label distributions, common when most data points are unambiguous. Gwet's AC1 \cite{gwet2014handbook} is robust to this paradox and is also chance-corrected, making it the most appropriate choice for our setting.

Two assumptions must hold to use Gwet's AC1: (i) each rater makes judgments independently, and (ii) every rater judges every question (fully crossed design). Both conditions are satisfied here: models are trained independently, and there is no ordering of the models with no meaningful distinctions between the models in $\mathcal{M}^{(m)}$. Also, every model predicts on every query point and its perturbations. The Gwet's AC1 score ranges from $-1$ to $1$, where $1$ indicates perfect agreement, $0$ indicates chance-level agreement, and negative values indicate systematic disagreement beyond chance.

The Gwet's AC1 score is shown here in the context of RAD Score-Pair computation. Let $\mathbf{G} \in \mathcal{Y}^{N \times K}$ be any matrix whose entry $G_{n,k}$ records the categorical label assigned by \textit{rater} $k \in \{1, \ldots, K\}$ (columns) to \textit{question} $n \in \{1, \ldots, N\}$ (rows), with labels drawn from a finite set $\mathcal{Y}$. Gwet's Agreement Coefficient AC1 on $\mathbf{G}$ is

\begin{equation}
\mathrm{AC1}(\mathbf{G}) \;=\;
\frac{p_o(\mathbf{G}) - p_e(\mathbf{G})}{1 - p_e(\mathbf{G})}
\label{eq:ac1}
\end{equation}

\noindent where

\begin{align*}
p_e(\mathbf{G}) &\;=\; \sum_{y \in \mathcal{Y}}
  \hat{\pi}_{\mathbf{G}}(y)\,\bigl(1 - \hat{\pi}_{\mathbf{G}}(y)\bigr)
\end{align*}
\begin{align*}
\hat{\pi}_{\mathbf{G}}(y) &\;=\; \frac{1}{N \, K}
  \sum_{n=1}^{N} \sum_{k=1}^{K}
  \mathbf{1}\!\left[G_{n,k} = y\right], \quad y \in \mathcal{Y}
\end{align*}
\begin{align*}
p_o(\mathbf{G}) &\;=\; \frac{1}{N} \sum_{n=1}^{N} \frac{1}{K(K-1)}
  \sum_{k=1}^{K} \sum_{\substack{k'=1 \\ k' \neq k}}^{K}
  \mathbf{1}\!\left[G_{n,k} = G_{n,k'}\right]
\end{align*}

\noindent here $p_e (\mathbf{G})$ is chance agreement probability of $\mathbf{G}$, $p_o(\mathbf{G})$ is the observed agreement in $\mathbf{G}$, averaged over all pairs of models across all points in the ambiguity matrix $\mathcal{A}_{\mathcal{T}_{\mathbf{t}}}^{\mathcal{M}^{(m)}}$. The marginal probability of the class label $y$ across all models, and all datapoints in the ambiguity matrix is denoted as $\hat{\pi}(y)$.

\subsubsection{RAD Score-Pair Computation: $[\mathrm{RAD}_{\mathrm{MSC}}$,$\mathrm{RAD}_{\mathrm{FSC}}]$}
We apply Gwet's AC1 score in two complementary ways: (1) treating models as independent raters to capture inter-model ambiguity (Model-Space Consistency), and (2) treating synthetic samples as independent raters to capture the intra-model ambiguity (Feature-Space Consistency).

\noindent \textbf{Model-Space Consistency ($\mathrm{RAD}_{\mathrm{MSC}}$)} captures the ambiguity arising from variations in the equivalent models. It is computed on the ambiguity matrix by treating models as raters, perturbed datapoints as questions, and predicted class labels as answers.
\begin{align*}
\mathrm{RAD}_{\mathrm{MSC}} = \mathrm{AC1} (\mathcal{A}_{\mathcal{T}_{\mathbf{t}}}^{\mathcal{M}^{(m)}})
\end{align*}
 This reveals inter-model disagreement over the neighbourhood.

\begin{itemize}
  \item $\mathrm{RAD}_{\mathrm{MSC}}$ $\sim$ high: Equivalent models mostly agree on all points in the neighbourhood.
  \item $\mathrm{RAD}_{\mathrm{MSC}}$ $\sim$ mid: Equivalent models agree on some points in the neighbourhood but disagree on others.
  \item $\mathrm{RAD}_{\mathrm{MSC}}$ $\sim$ low: Equivalent models disagree on most or all synthetic samples in the neighbourhood. The decision boundaries of equivalent models do not agree at all.
\end{itemize}

\noindent\textbf{Feature-Space Consistency ($\mathrm{RAD}_{\mathrm{FSC}}$)} captures ambiguity arising from variations in the datapoints in the neighbourhood of $\mathbf{t}$. It is computed on the transposed ambiguity matrix by treating synthetic samples as raters and models as questions.  

\begin{align*}
\mathrm{RAD}_{\mathrm{FSC}} = AC1 ((\mathcal{A}_{\mathcal{T}_{\mathbf{t}}}^{\mathcal{M}^{(m)}})^{T})
\end{align*}
This reveals how stable each model is across the neighbourhood.

\begin{itemize}
  \item $\mathrm{RAD}_{\mathrm{FSC}}$ $\sim$ high: Each equivalent model is internally stable in the neighbourhood.
  \item $\mathrm{RAD}_{\mathrm{FSC}}$ $\sim$ mid: Some equivalent models are stable across the neighbourhood, while others are not.
  \item $\mathrm{RAD}_{\mathrm{FSC}}$ $\sim$ low: Most equivalent models flip their prediction within the neighbourhood. Model class is locally unstable.
\end{itemize}

When the ambiguity-matrix has a single row, $\mathrm{RAD}_{\mathrm{MSC}}$ reduces to a pairwise inter-model agreement at one single point (similar to single-point based metrics such as Rashomon Capacity \cite{hsu2022rashomon}). The integration over $p$ neighbourhood points allows us to distinguish disagreement that persists within the neighbourhood from disagreement that exists only because $\mathbf{t}$ happens to be on an overlapping decision boundary.

On the other hand, if the ambiguity matrix has a single column, $\mathrm{RAD}_{\mathrm{FSC}}$ becomes similar to a local-robustness measurement \cite{leino2021relaxing,zhong2021understanding}. The integration over the $n$ equivalent models allow to distinguish a region where the model alone is unstable from a region where the entire model class is locally unstable.

Together, the RAD Score-Pair [$\mathrm{RAD}_{\mathrm{MSC}}$,$\mathrm{RAD}_{\mathrm{FSC}}$] reveals the source of robust-ambiguity. The analysis, interpretation and use of the RAD Score-Pair are discussed in the next section.

\section{Interpreting RAD Score-Pair and RAD Plot}\label{sec:rad_interpret}

This section describes how the RAD Score-Pair is interpreted, visualised, and ranked. Section \ref{subsec:radplot} introduces the RAD Plot and its four quadrants. Section \ref{subsec:readingplot} gives the aggregate distribution of points in the plot. Section \ref{subsec:radrank} introduces RAD-Pareto-Rank, a method to rank ambiguous datapoints for downstream tasks.

\subsection{RAD Plot and Quadrants}\label{subsec:radplot}
The RAD Score-Pair [$\mathrm{RAD}_{\mathrm{MSC}}$,$\mathrm{RAD}_{\mathrm{FSC}}$] provides a two-dimensional view of how confusing a datapoint is for the deployed model class. For multi-class problems with $c$ classes under a one-vs-all decomposition, every datapoint will have $c$ such pairs; one per class, thus enabling per-class ambiguity analysis. We visualise the RAD Score-Pairs as a scatterplot in 2D space as shown in Fig. \ref{fig:demo_RAD}.

The RAD Score-Pair space can be divided into four quadrants using a threshold on each axis. The general idea of RAD Plot is captured in Fig. \ref{fig:demo_RAD}. This threshold can be calibrated based on the application's tolerance for ambiguity, the specific way the scores were computed (in this case Gwet's AC1 score) and the budget for human review: higher thresholds (e.g., 0.7) flag more datapoints as ambiguous, lower thresholds flag fewer. In all our experiments, we use a threshold of $0.5$ on each axis as it splits right in the middle of the "moderate agreement" level \cite{gwet2014handbook}. Each quadrant corresponds to a qualitatively distinct cause of predictive ambiguity. The properties of each quadrant (\textbf{Q1-Q4}) of the RAD Plot are discussed next.

\begin{flushleft}
    \textbf{Q1: [$\mathrm{RAD}_{\mathrm{MSC}},\mathrm{RAD}_{\mathrm{FSC}}$] = [High, High]}. Models agree with each other, and each model is stable across the neighbourhood. This is the \textbf{robust region} in the model- and feature-space, where models are able to retain their predictions despite minor permissible variations to both the model and data.
\end{flushleft}

\begin{flushleft}
    \textbf{Q2: [$\mathrm{RAD}_{\mathrm{MSC}},\mathrm{RAD}_{\mathrm{FSC}}$] = [Low, High]}. Each individual model is internally stable; it keeps predicting the same class throughout the neighbourhood, but the models disagree with each other. The models have learned different but internally coherent boundaries in this region. The ambiguity is a global model-level disagreement, not a local spatial one. The neighbourhood itself is not ambiguous, as it sits cleanly on one side of each model's boundary, but the models placed those boundaries in different places. This is \textbf{epistemic uncertainty} \cite{hullermeier2021aleatoric} about which model is right, not about where the boundary is.
\end{flushleft}

\begin{flushleft}
    \textbf{Q3: [$\mathrm{RAD}_{\mathrm{MSC}},\mathrm{RAD}_{\mathrm{FSC}}$] = [Low, Low]}. Models disagree with each other, and each individual model also changes its prediction across the neighbourhood. This is a strong indication to abstain, since ambiguity can be attributed to both sources.
\end{flushleft}

\begin{flushleft}
    \textbf{Q4: [$\mathrm{RAD}_{\mathrm{MSC}},\mathrm{RAD}_{\mathrm{FSC}}$] = [High, Low]}. Models broadly agree with each other, but each individual model is unstable across the neighbourhood. The models have placed their boundaries in roughly the same place, and that boundary runs directly through the neighbourhood. The models are consistent with each other about where the boundary is, but the test point is sitting right on top of it.
\end{flushleft}

\subsection{Overall Interpretation of RAD Plot}\label{subsec:readingplot}

The quadrant interpretation above describes individual datapoints. The \emph{aggregate distribution} of points across the plot is also informative, since it indicates whether the dominant source of unreliability is the model-space, feature-space, or both, and what kind of intervention is likely to help. We identify four characteristic patterns, summarised in Table~\ref{tab:aggregate-interpretation}.

\textbf{Right skew} (mass in Q1 + Q4) is feature-space dominant: the model class has converged on a coherent boundary, but the data sits in fragile regions relative to it. \textbf{Upward skew} (Q1 + Q2) is model-space dominant: each model is internally stable, but different models placed their boundaries elsewhere. \textbf{Diagonal spread} (Q1 to Q3) couples both instabilities and is a typical signature of label noise or inconsistent annotation. Concentration in Q3 is the worst case, where neither space is stable.
\begin{table}[t]
\centering
\caption{Aggregate interpretation of RAD Plot. Each row describes a characteristic pattern in the distribution of points across the plot. The dominant source of unreliability it implies, and some model-side and data-side possible actions are listed.}
\label{tab:aggregate-interpretation}
\small
\setlength{\tabcolsep}{6pt}
\renewcommand{\arraystretch}{1.2}
\resizebox{1.0\textwidth}{!}{
\begin{tabular}{@{}l p{0.15\textwidth} p{0.35\textwidth} p{0.25\textwidth}@{}}
\toprule
\textbf{Pattern} & \textbf{Dominant cause} & \textbf{Model-Space action} & \textbf{Feature-Space action} \\
\midrule
Right skew (Q1 + Q4)
  & Feature-space
  & Model side action is unlikely to help, since the models already agree.
  & Feature engineering, or targeted data collection in the affected sub-region. The ambiguity can also be irreducible (aleatoric) if none of the previous actions helps. \\
\addlinespace[2pt]
Upward skew (Q1 + Q2)
  & Model-space
  & Better regularisation. Ensembling can reduce the variance, but not the underlying disagreement.
  & More labels in the disputed region. Active learning rather than uniform collection. \\
\addlinespace[2pt]
Diagonal spread (Q1 $\to$ Q3)
  & Both spaces are unstable. Can be label-noise, low model capacity, or sharp boundary transitions.
  & Increase model capacity (if underfitting). Model family change is unlikely to help.
  & Audit affected points for labelling errors, inter-annotator disagreement, or ambiguous class definitions. \\
\addlinespace[2pt]
Concentrated in Q3
  & Both spaces are unstable.
  & Flag for human review. Increasing model capacity, regularisation, alternative model family. If still in Q3, ambiguity is consistent with irreducible uncertainty.
  & Audit features and labels. Consider whether the task definition (class taxonomy) is well-posed. \\
\bottomrule
\end{tabular}}
\end{table}

\subsection{Ranking Robust-ambiguity with Pareto Frontier}\label{subsec:radrank}
Finding the most ambiguous datapoints requires minimising both $\mathrm{RAD}_{\mathrm{MSC}}$ and $\mathrm{RAD}_{\mathrm{FSC}}$, which can be treated as a bi-objective problem solved by the Pareto-optimal frontier \cite{deb2011multi}. A point is \textit{Pareto-optimal} if no other point has lower $\mathrm{RAD}_{\mathrm{MSC}}$ and lower $\mathrm{RAD}_{\mathrm{FSC}}$. Successive \textit{shells}, obtained by "peeling", rank datapoints from most to least ambiguous (Fig.~\ref{fig:demo_RAD}). We combine the Pareto-frontier selection, with the quadrant interpretation in two passes. First, we peel shells restricted to points outside Q1 (i.e., in Q2/Q3/Q4), ordering them from most to least ambiguous. Once these are exhausted, the remaining points all lie in Q1, and we rank them by Pareto shell as well, so that the least ambiguous robust points appear last. Within any shell, ties are broken by Euclidean distance from the perfect-agreement point $(1, 1)$ in the RAD Plot.

This shell-by-shell ordering, which also accounts for the RAD Plot quadrants, is named \textbf{RAD-Pareto-Rank}. It is used in the abstention experiments in Section~\ref{subsec:result_abstain}, and is one mechanism by which the RAD Score-Pair can be deployed in a downstream task.

\section{Experiments}\label{sec:experiments}

Experiments were conducted to evaluate how RAD can be used to investigate the ambiguity levels on a wide range of datasets, both synthetic and real. The experiments demonstrate the usefulness of the RAD Score-Pairs and RAD Plot first. Next we also demonstrate an example downstream task of abstention \cite{hendrickx2024machine,zhang2023survey} based on the RAD-Pareto-Ranks of test datapoints.

\begin{flushleft}
    \textbf{Synthetic Datasets.} RAD was applied to $5$ binary synthetic datasets - blobs, spirals, concentric circles, moons, and checkerboard; each at three class overlap levels, low, medium and high ($5\times3 = 15$ synthetic configurations). Each dataset contained 1000 datapoints (80/20 train/test split). 
\end{flushleft}

\begin{flushleft}
    \textbf{Real-world datasets.} 
    From UCI ML Repository \cite{asuncion2007uci}, 16 datasets were used to demonstrate ambiguity visualization when decision boundaries cannot be visually inspected (Binary class: \textit{Magic Gamma Telescope}, \textit{Default of Credit Card Clients}, \textit{Rice Bin}, \textit{Banknote Authentication}, \textit{Heart Failure Risk}, \textit{Mammographic Mass}; Multi-class: \textit{Statlog}, \textit{Glass Identification}, \textit{E-Coli}, \textit{Iris}, \textit{Optical Digit Recognition}, \textit{Handwritten Digit Recognition}, \textit{Heart Disease}, \textit{Breast Cancer (Wisconsin)}, \textit{COMPAS}, \textit{Wine Quality}). The MNIST dataset \cite{deng2012mnist} was also used. In both cases, we make 80/20 train/test stratified splits.
\end{flushleft}

 \begin{flushleft}
     \textbf{Abstention Experiment.} To demonstrate a downstream task, we perform an abstention experiment by ranking datapoints following \textbf{RAD-Pareto-Ranks} as described in Section \ref{subsec:radrank}. We train a single hyperparameter optimised decision tree for prediction only, for each dataset. Based on the best hyperparameter set, we train the RAD framework, which we use to get the RAD Score-Pairs. In the test set, we then progressively reject samples using the RAD-Pareto-Rank order, measuring accuracy on the remaining test set. This produces accuracy at varying coverage levels (100\% to 0\% of data). We report the Area Under the Rejection Curve (AURC): accuracy vs.\ coverage \cite{hendrickx2024machine}.
    
    We compare RAD-Pareto rejection against the following baselines:
    \begin{itemize}
      \item \textbf{Self-Consistency (SC)}: Average pairwise agreement across the 25 models on a single test datapoint. Samples with the lowest SC are rejected first \cite{cooper2024arbitrariness}.
      \item \textbf{Entropy}: Class membership entropy from individual model predictions. Standard uncertainty-based rejection \cite{gal2016dropout}.
      \item \textbf{Random}: A baseline, where samples are selected randomly.
    \end{itemize}
 \end{flushleft}

The RAD Framework stages are configured as follows.
\begin{flushleft}
    \textbf{Model for Prediction.} We have chosen decision tree algorithm \cite{breiman2001statistical} for all experiments. For each dataset, the best hyperparameter is tuned through a 5-fold stratified cross-validation experiment, and the best model is fit on the full training set. Note that this model is only used for the prediction, and not a part of RAD.
\end{flushleft}

\begin{flushleft}
    \textbf{RAD Stage 1 Setup.} Using the best hyperparameter configuration, $n=25$\footnote{Bootstrap-model sampling and neighbourhood sampling are both stochastic, so the RAD Score-Pair for a test point will have variance that shrinks as $n$ and $p$ grow. Preliminary runs on a subset of our datasets showed that the per-point Score-Pair varied negligibly under independent re-runs at $n = 25$ and $p = 100$.\label{foot:n_and_p_select}} decision tree models were trained on bootstrap samples of the training set to build the set of equivalent models. This is the $\mathcal{M}^{(m)}$ set.
\end{flushleft}

\begin{flushleft}
    \textbf{RAD Stage 2 Setup.} For a given test point $\mathbf{t}$, a total of $p=100$\footref{foot:n_and_p_select} points (inclusive of $\mathbf{t}$) from the neighbourhood $\mathcal{T}_{\mathbf{t}}$ is generated following Eq.~\eqref{eq:smote_like}. For the MNIST dataset, an autoencoder was trained first, and the perturbations were made in the latent space, as described in Section \ref{subsec:stage2}. For all the experiments, we have used $k=10$ nearest neighbours to generate the local synthetic samples (details in Section \ref{subsec:result_k_select}).
\end{flushleft}

 \begin{flushleft}
     \textbf{RAD Stage 3 Setup.} The details of this stage are described in Section \ref{subsec:stage3}. In addition to this, we generate per-class one-vs-all ambiguity-matrices to get per-class RAD Plots and RAD Score-Pairs ($c$ classes will lead to $c$ Rad Score-Pairs for a $\mathbf{t}$). This will enable tracing the per-class properties. Instead of the original Gwet's AC1 recommended seven binned levels \cite{gwet2014handbook}, we simplify and choose the midpoint of the bins and set $0.5$ to be the "mid" point along both the axes of RAD Plot to indicate the threshold to split the quadrants.
 \end{flushleft}

\section{Results and Discussion}\label{sec:results}
In this section, we analyse and discuss the results for the synthetic dataset (Section \ref{subsec:result_syn}) and real-world (Section \ref{subsec:result_real}) datasets, with respect to RAD Plots. In  Section \ref{subsec:result_abstain}, we present the results of the abstention experiment.

\begin{figure}[!t]
\centering

\subfloat[Low overlap]{\includegraphics[width=0.33\textwidth]{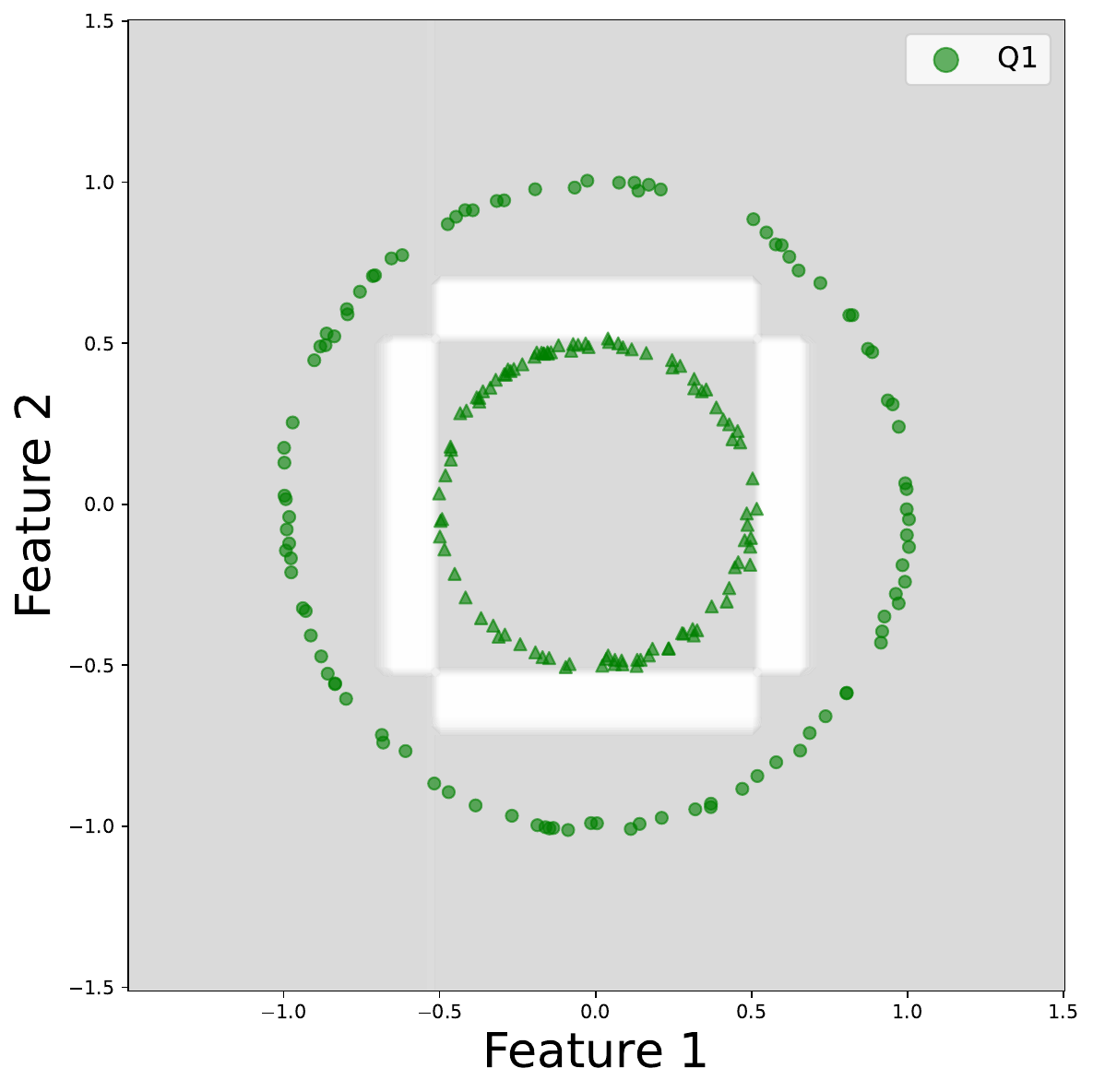}}
\subfloat[Medium overlap]{\includegraphics[width=0.33\textwidth]{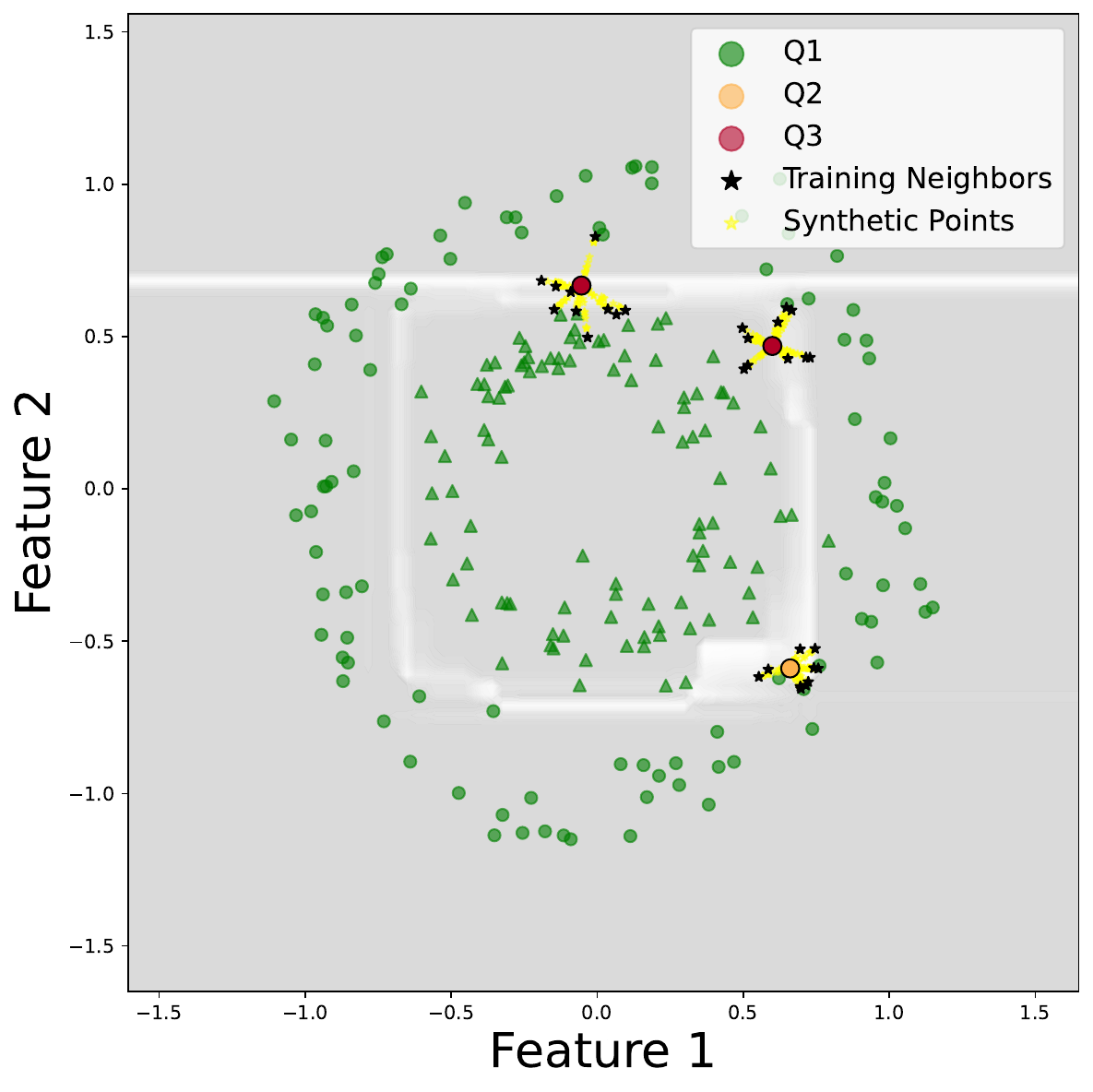}}
\subfloat[High overlap]{\includegraphics[width=0.33\textwidth]{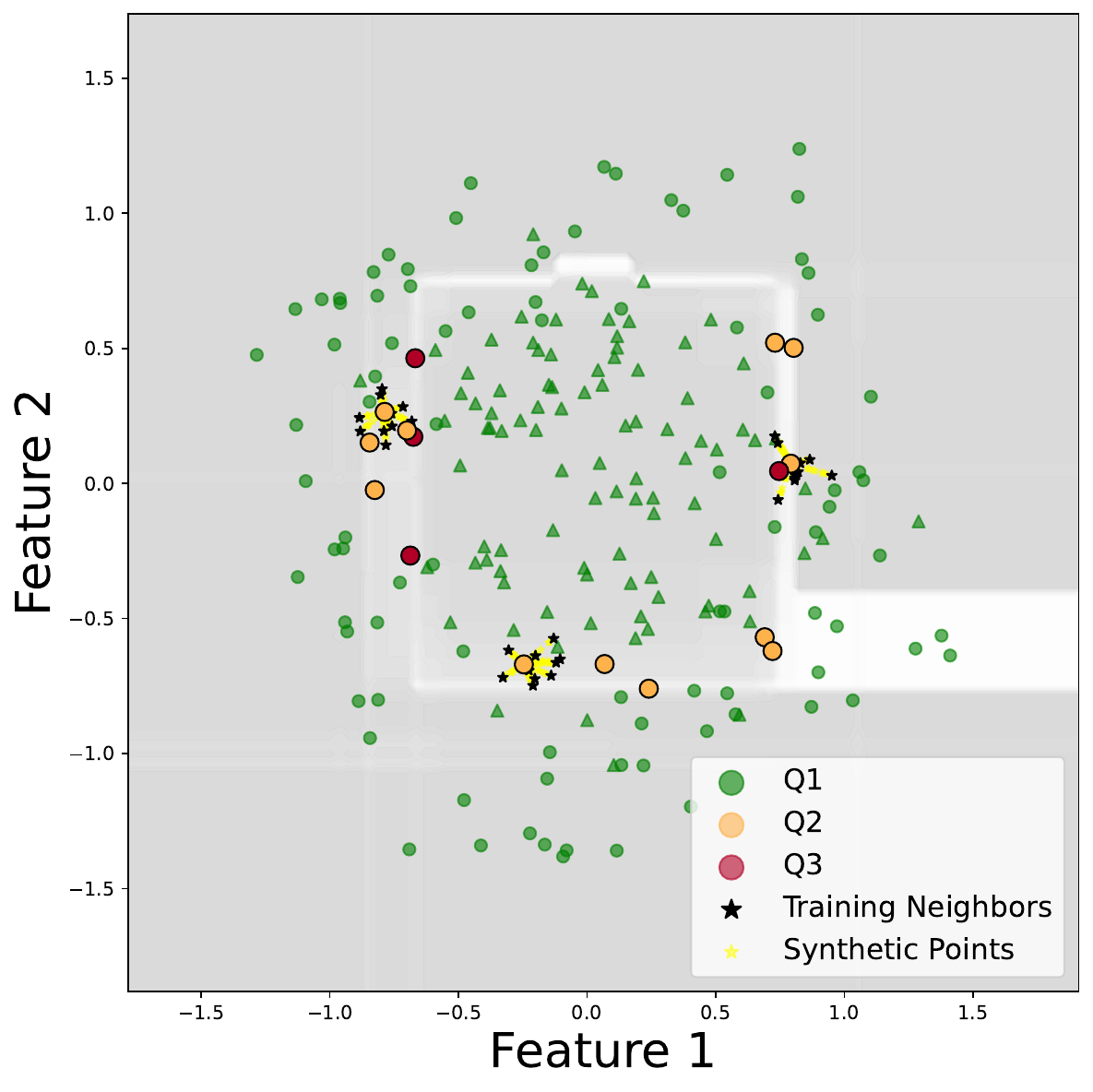}}

\subfloat[Low overlap RAD Plot]{\includegraphics[width=0.33\textwidth]{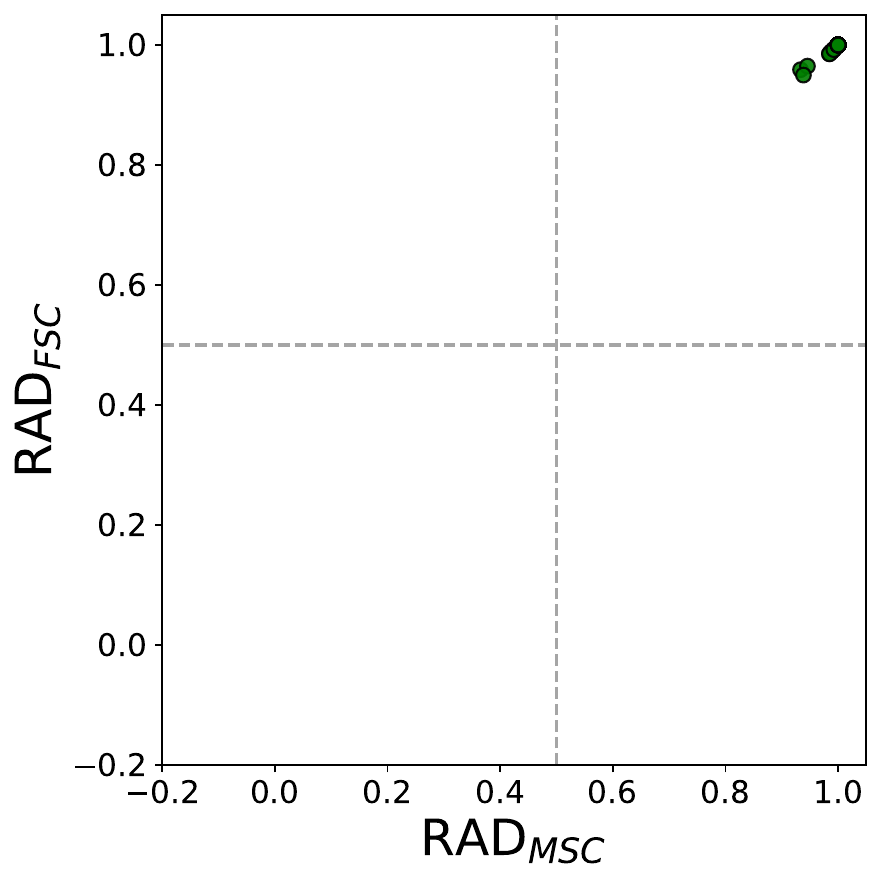}}
\subfloat[Medium overlap RAD Plot]{\includegraphics[width=0.33\textwidth]{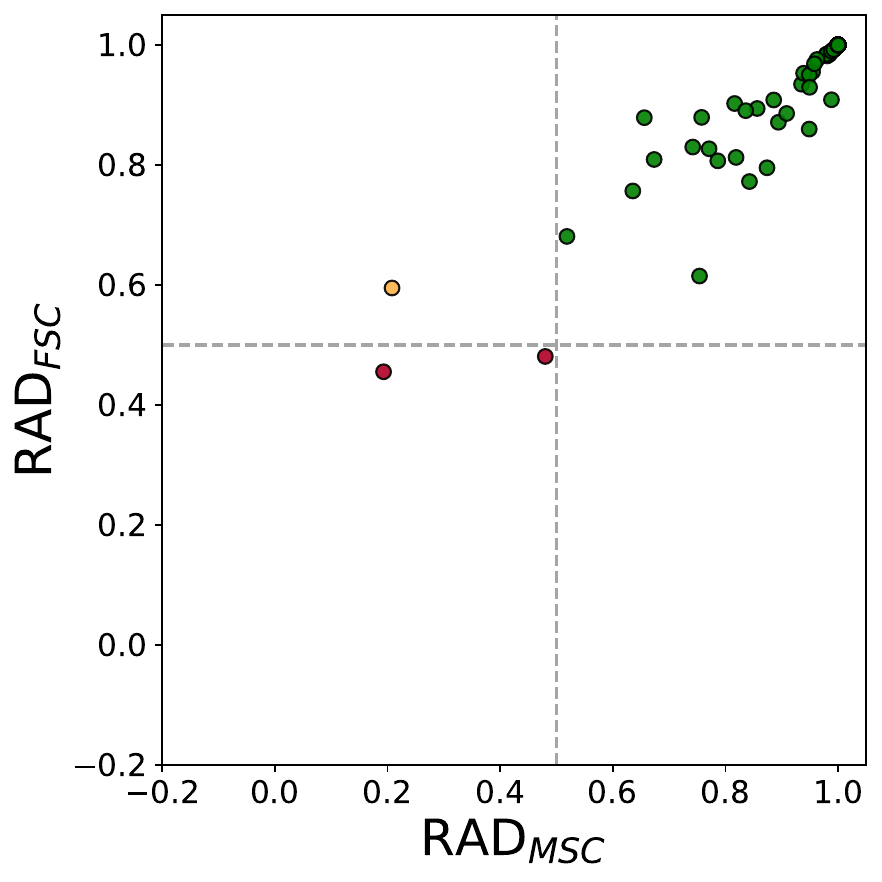}}
\subfloat[High overlap RAD Plot]{\includegraphics[width=0.33\textwidth]{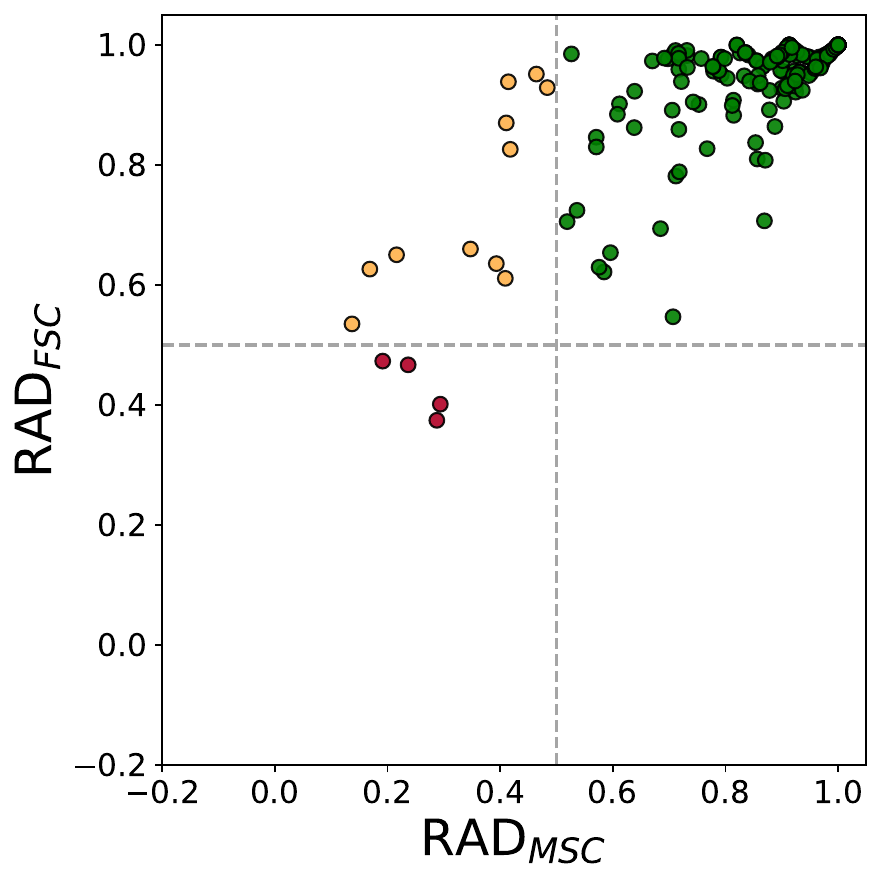}}

\caption{Left column: Decision boundary plots in 2-dimensions for the \textit{two circles} dataset. Right column: RAD Plot showing the distribution of RAD Score-Pair.}

\label{fig:synthetic_dataset_blobs_circles}
\end{figure}

\begin{figure}[!t]
\centering

\subfloat[Low overlap]{\includegraphics[width=0.33\textwidth]{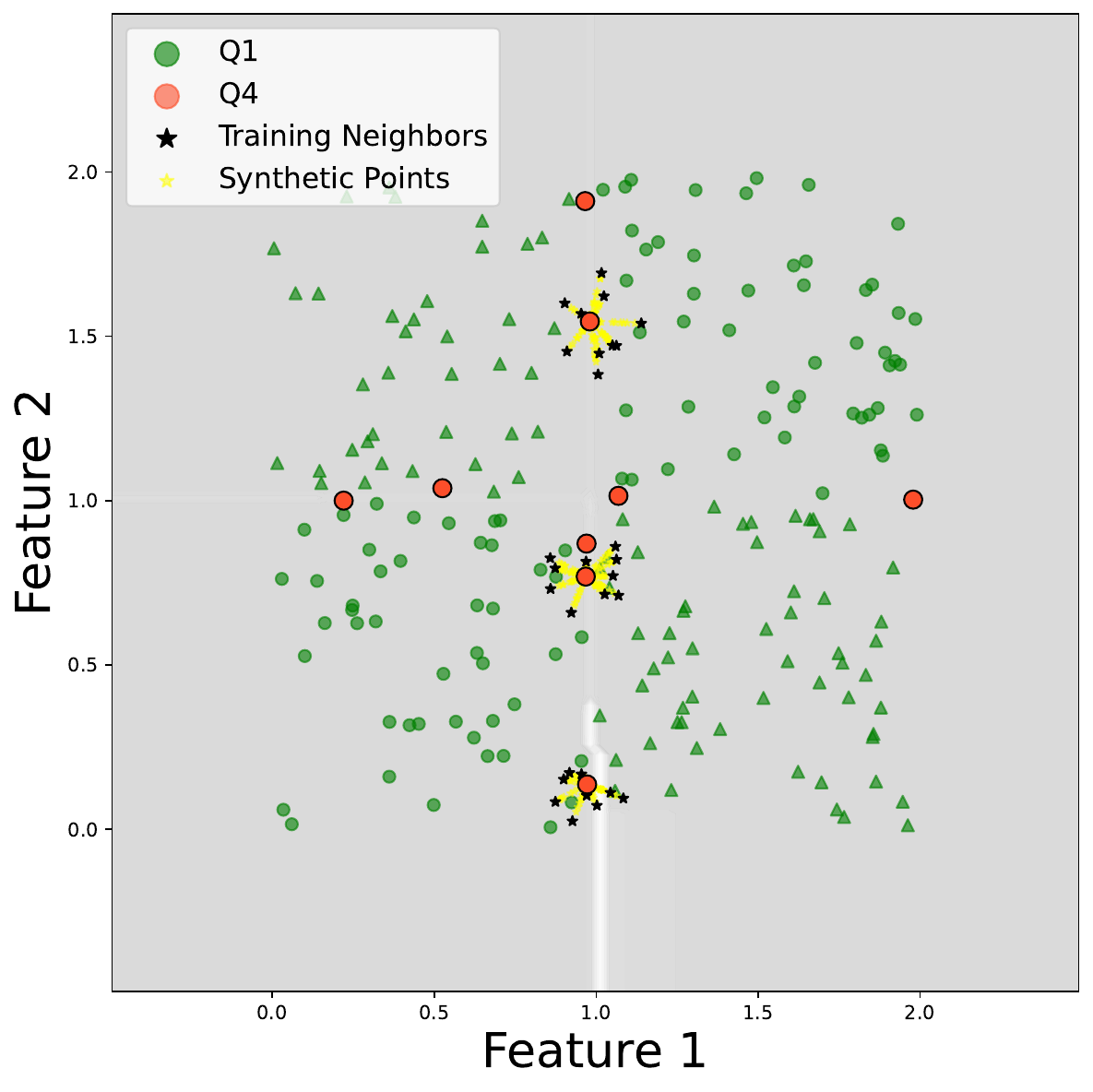}}
\subfloat[Medium overlap]{\includegraphics[width=0.33\textwidth]{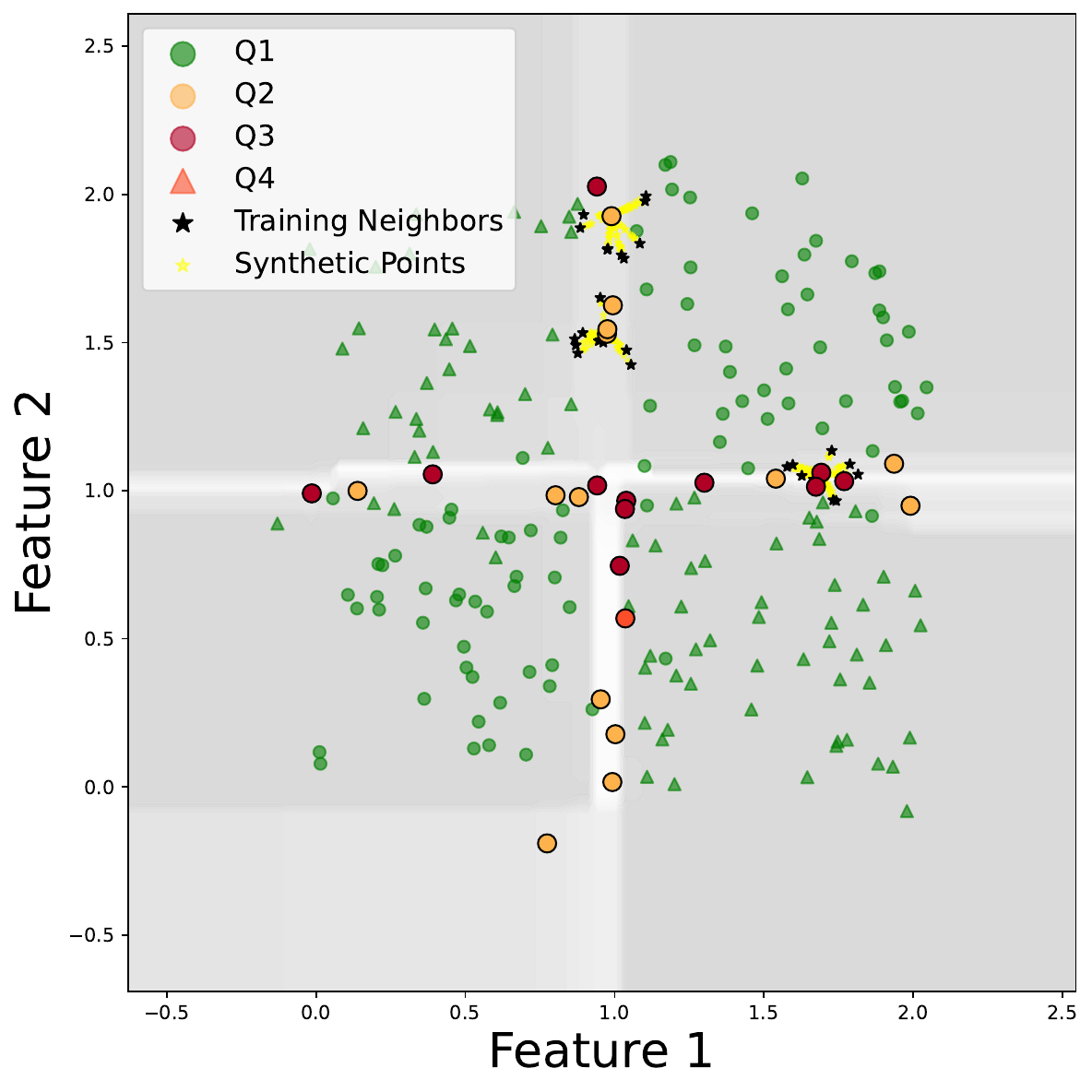}}
\subfloat[High overlap]{\includegraphics[width=0.33\textwidth]{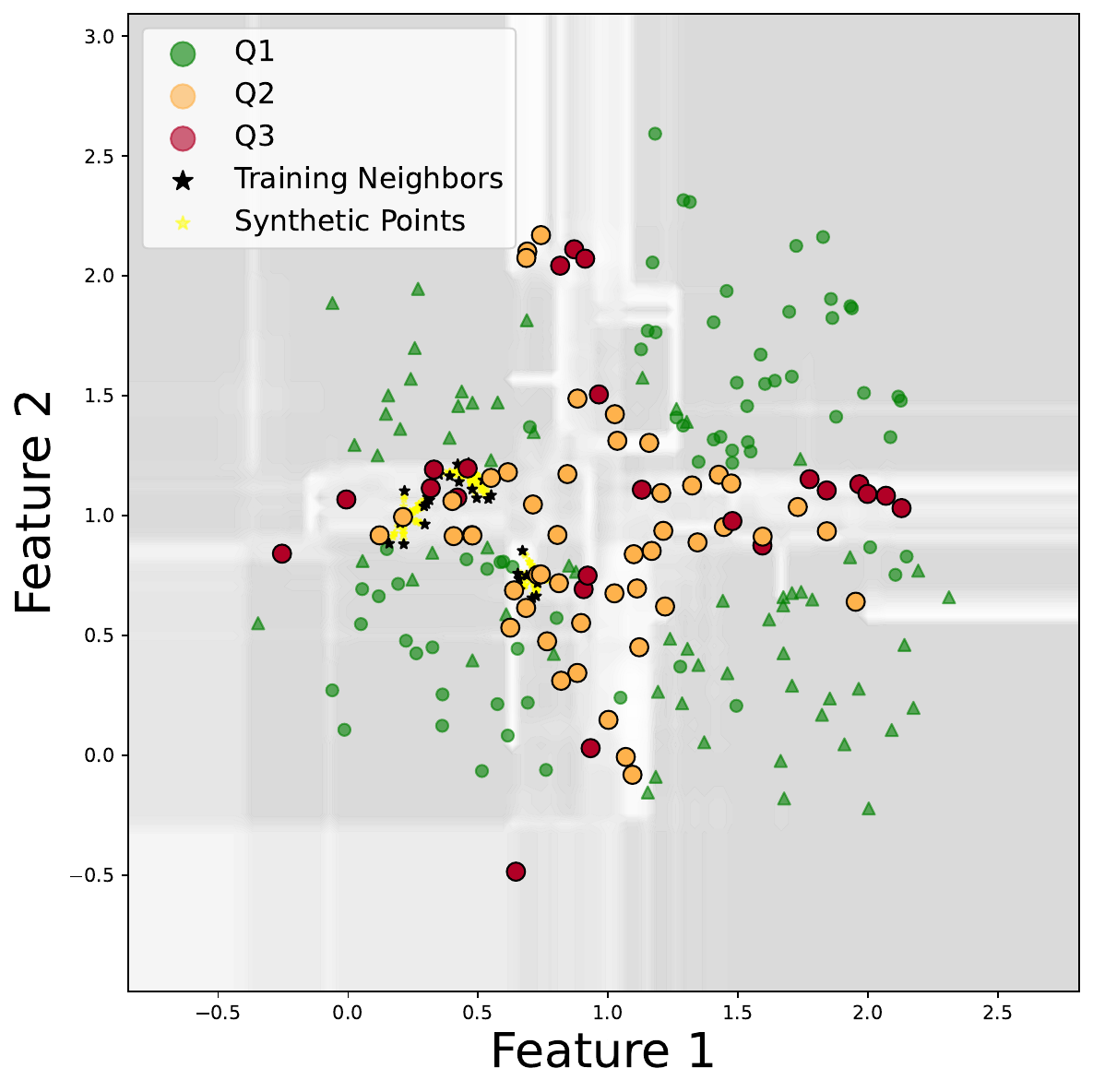}}

\subfloat[Low overlap RAD plot]{\includegraphics[width=0.33\textwidth]{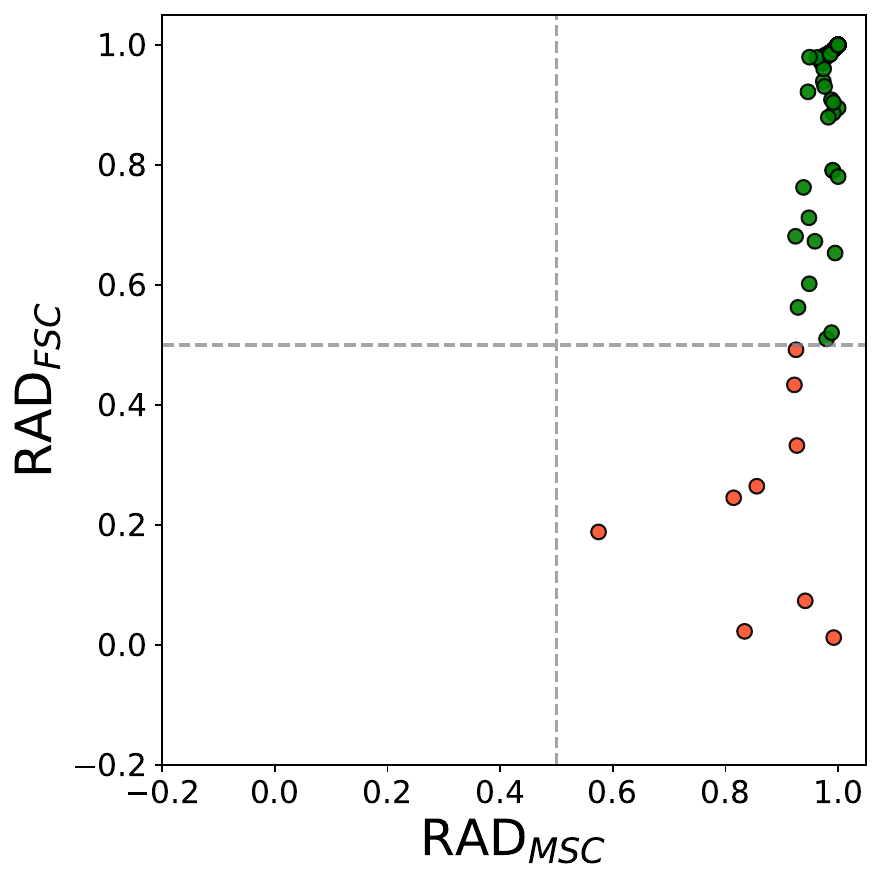}}
\subfloat[Medium overlap RAD plot]{\includegraphics[width=0.33\textwidth]{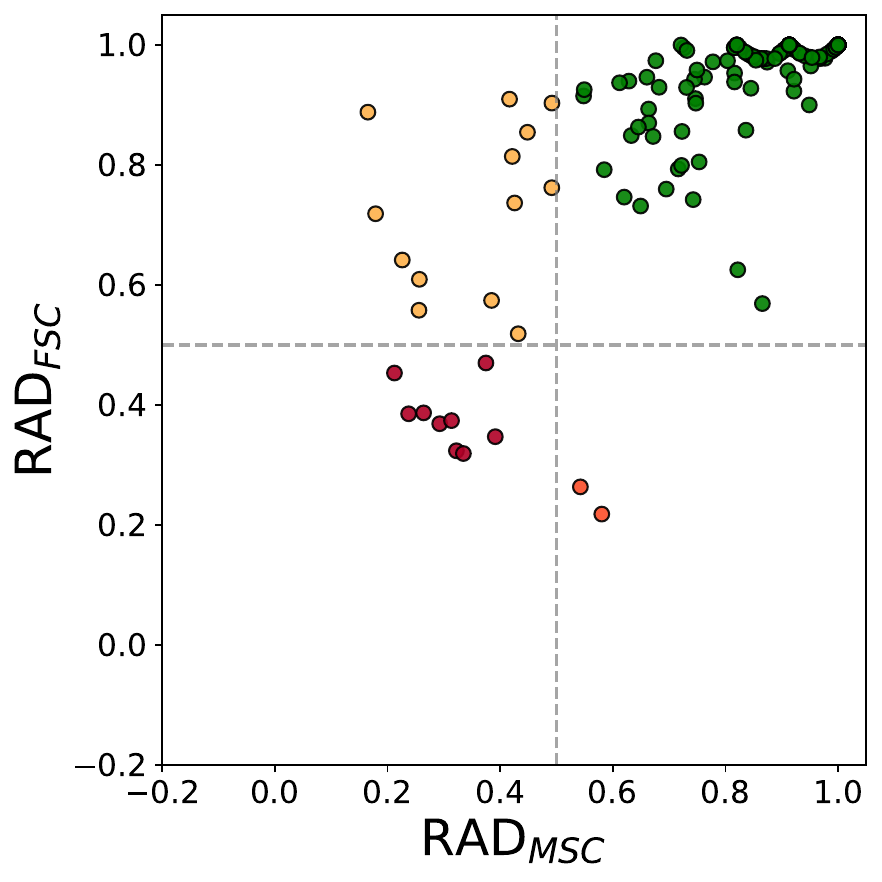}}
\subfloat[High overlap RAD plot]{\includegraphics[width=0.33\textwidth]{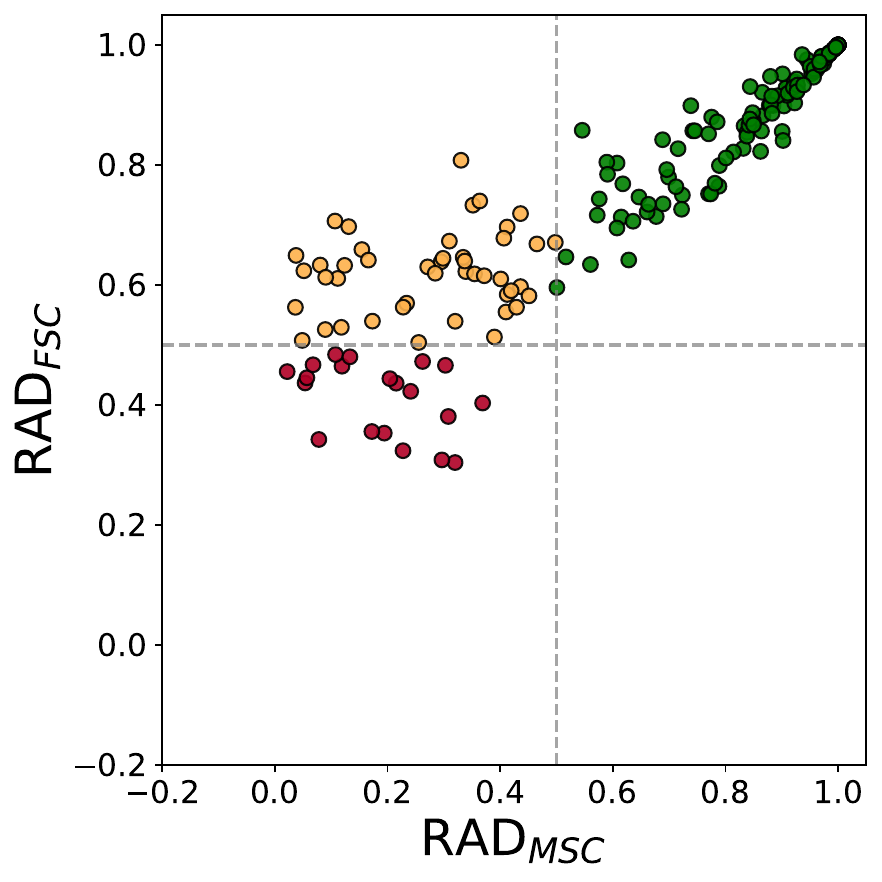}}

\caption{Left column: Decision boundary plots in 2-dimensions for the \textit{checkerboard} dataset. Right column: RAD Plot showing the distribution of RAD Score-Pair.}
\label{fig:synthetic_datasets_check}
\end{figure}

\subsection{Selecting $k$ for RAD Stage 2}\label{subsec:result_k_select}
The number of nearest neighbours $k$ in Eq.~\eqref{eq:smote_like} (Stage 2, Section \ref{subsec:stage2}) controls how local the perturbation sampling is. We choose a plausible default at which the quadrant assignments are stable. Choosing from $k \in \{2, 3, 4, 6, 8, 10, 13, 15, 20, 25, 30\}$, we recorded the smallest $k$ beyond which test points stopped switching quadrants. On $21$ of the $32$ datasets where any switching occurred, the mean stabilisation point was $k = 11.7$ (median $11.0$). The rest showed no switches across the range. We use $k = 10$ throughout, which is the closest round value and is for all datasets for our purpose.

\subsection{Synthetic Datasets}\label{subsec:result_syn}
Using the synthetic datasets, experiments were designed to test two hypotheses of our framework. First, increasing class overlap moves points away in the RAD Plot from the perfect agreement vertex $(1,1)$ in Q1 towards the ambiguous quadrants. Second, the direction in which the points move depends on the nature of the decision boundary. The results show that both hypotheses are true. Fig. \ref{fig:synthetic_dataset_blobs_circles} and Fig. \ref{fig:synthetic_datasets_check} illustrate two cases. The full set of results for the synthetic datasets is available in the Appendix~\ref{sec:appendix}.


Across all five synthetic datasets, as class overlap is increased from low to high, the RAD Plots show systematic movement of the points out of Q1 and into the ambiguous quadrants. At low overlap, the cloud is concentrated in Q1, at medium and high overlap, the mass spreads towards Q2, Q3, or Q4 depending on the nature of the decision boundary. This is the predicted behaviour, where RAD identifies more ambiguous points as the modelling task itself becomes truly ambiguous.

In Fig. \ref{fig:synthetic_dataset_blobs_circles}, for the concentric circles dataset with mid and high overlap, the ambiguous points cluster in Q2 in the RAD Plot, with some spread into Q3. This is the \textit{upward-skew} pattern mentioned in Section \ref{subsec:radplot}. Each individual model is internally stable in the neighbourhood of these points, but different models in the ambiguity-matrix have placed their boundaries through the overlapping region. The width of the ambiguous region in feature-space Fig.  \ref{fig:synthetic_dataset_blobs_circles} is wide enough that perturbations stay on one side of any given model's boundary, but the boundaries themselves disagree. The diagnosis is therefore model-space dominant.

In the case of the checkerboard dataset, in Fig. \ref{fig:synthetic_datasets_check}, in the low overlap case, the dataset is nearly perfectly split along the two features, but induced noise places a small number of points exactly on the partition lines. These points appear in Q4, where the models agree about where the boundary is, but the points sit on top of it, so local perturbations flip predictions within each model. This aligns with the interpretation of Q4 from Section \ref{sec:rad_interpret}. As overlap increases, individual boundaries diverge, and the Q4 band widens into a mix of Q3 and Q4, indicating model disagreement now compounds the local instability.

When boundary regions are wide, and models disagree about their placements, ambiguous datapoints populate both Q2 and Q3, a diagonal-spread signature. The aggregate pattern correctly identifies that both model-space and feature-space instabilities are important.

These results serve as a validation of the framework. This same logic extends to the real-world dataset, where the decision boundaries cannot be directly inspected.

\subsection{Real-world Datasets}\label{subsec:result_real}

With the synthetic experiments confirming that RAD Score-Pair and RAD Plot can correspond to interpretable decision boundaries, we now analyse the results of the real-world datasets. We present five binary datasets and two multi-class datasets to analyse them. The full set of results of 17 datasets are available in the Appendix~\ref{sec:appendix}.

\subsubsection{Binary Datasets}
\begin{figure}[htbp]
\centering
\subfloat[\textit{Gamma Telescope}\label{fig:binary_uci_gamma}]{\includegraphics[width=0.33\textwidth]{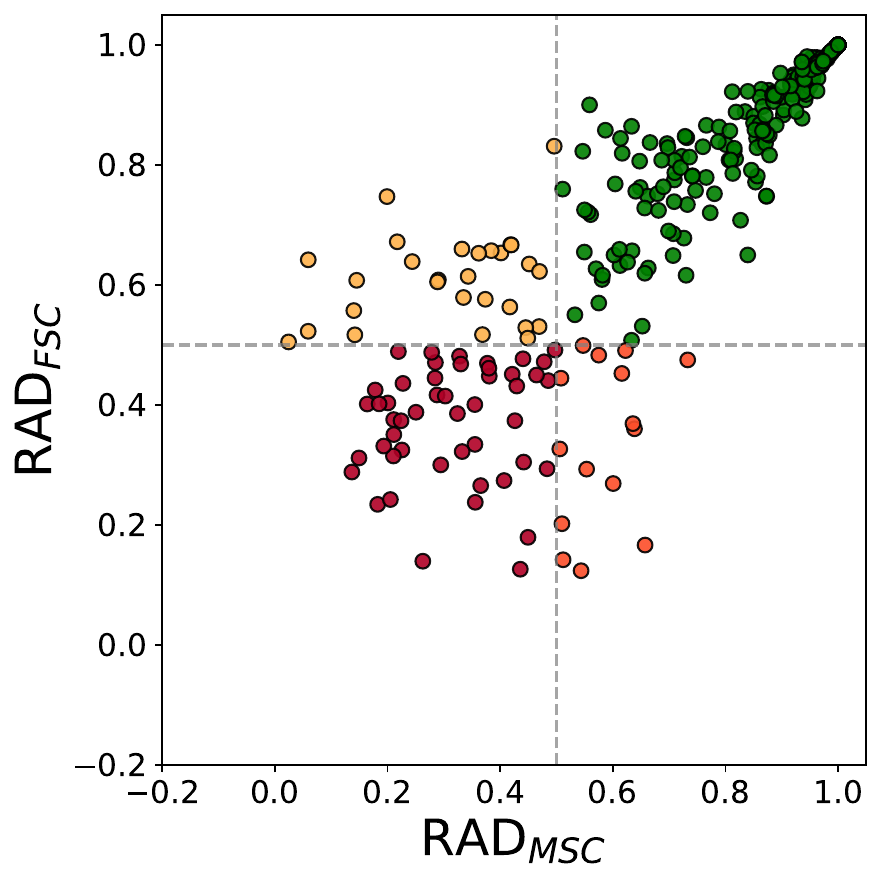}}
\subfloat[\textit{Banknote Authentication}\label{fig:binary_uci_bank}]{\includegraphics[width=0.33\textwidth]{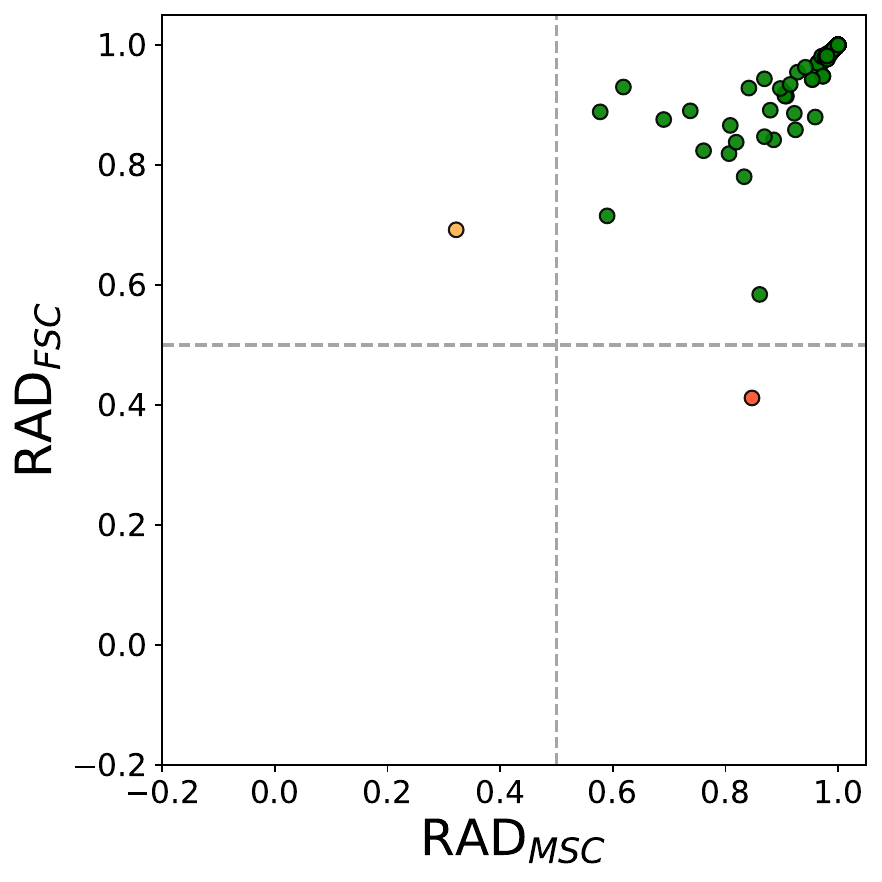}}
\subfloat[\textit{Heart Failure}\label{fig:binary_uci_heart}]{\includegraphics[width=0.33\textwidth]{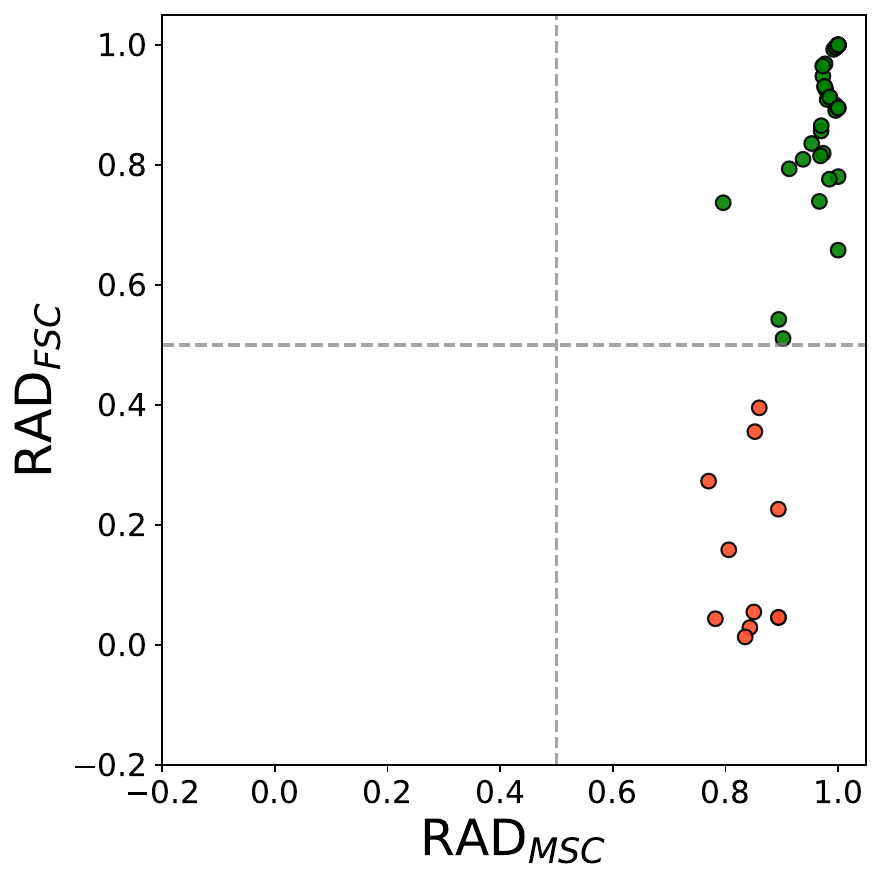}}

\subfloat[\textit{Breast Cancer (Wisconsin)}\label{fig:binary_uci_wisconsin}]{\includegraphics[width=0.33\textwidth]{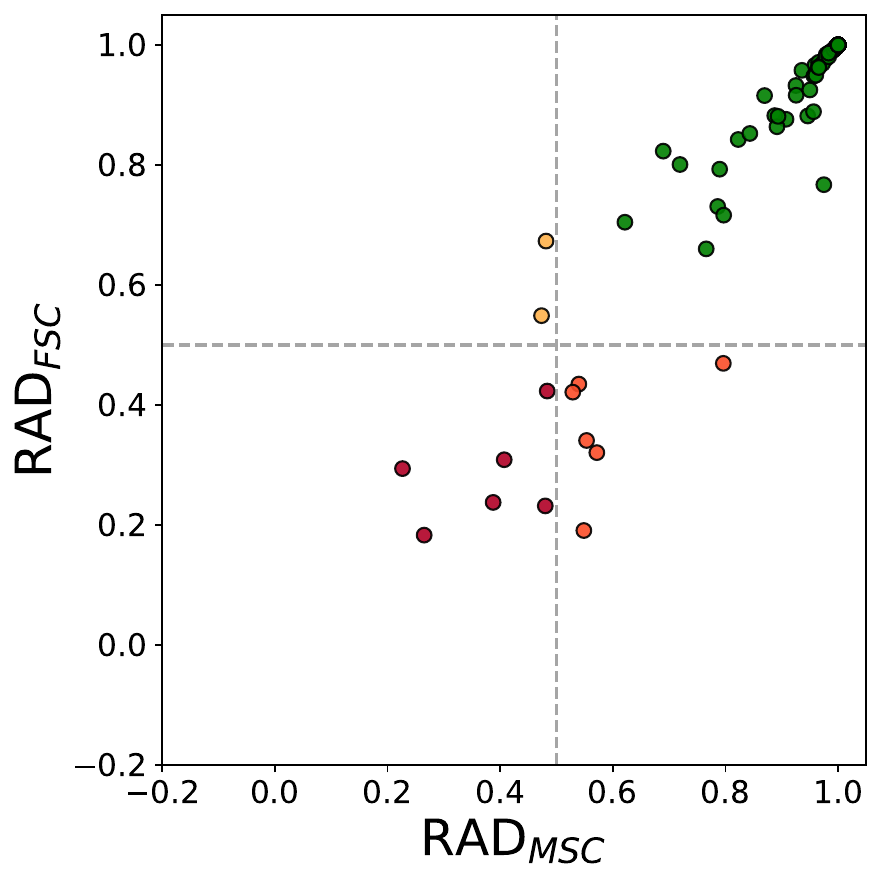}}
\subfloat[\textit{Rice}\label{fig:binary_uci_rice}]{\includegraphics[width=0.33\textwidth]{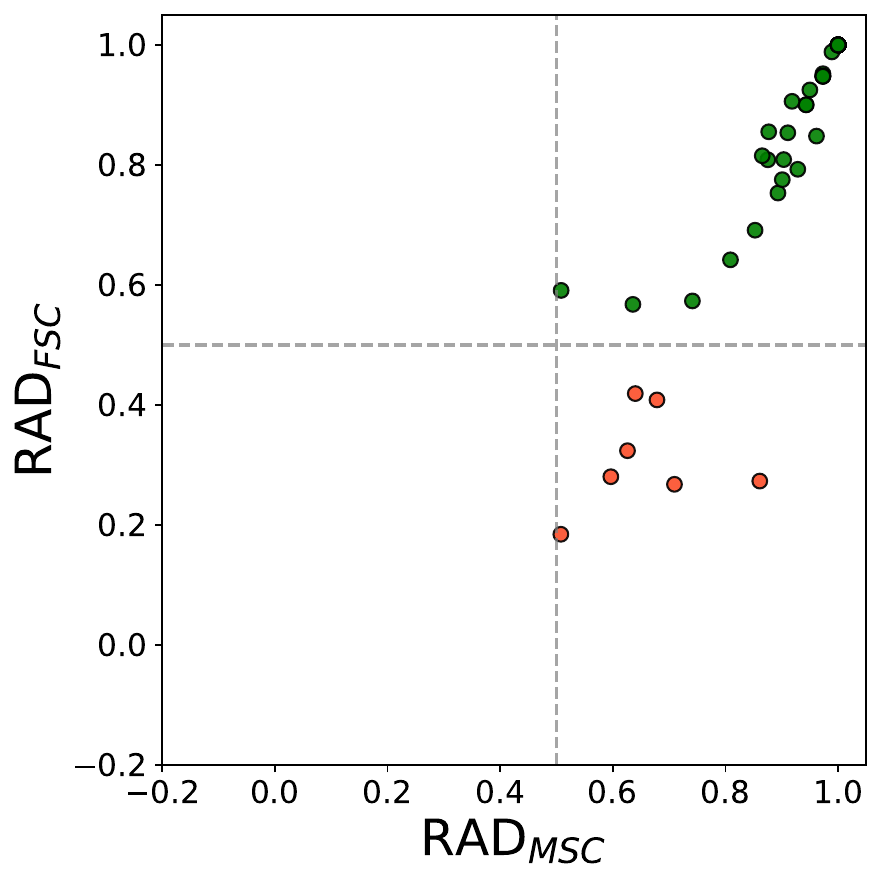}}

\caption{Real-world binary dataset RAD Plots.}
\label{fig:binary_uci}
\end{figure}

We show five binary datasets, each specifically known for possible class overlap. Fig.\ref{fig:binary_uci} shows the RAD Plot for each of them.

In Fig. \ref{fig:binary_uci_gamma} for the Gamma Telescope dataset, the task is to predict whether a captured light signal originates from a \textit{gamma} event (class 0) or a \textit{hadron} event (class 1). These two phenomena produce physically similar signals. Points are distributed across every quadrant, indicating that both sources of ambiguity are contributing to this dataset. Some samples sit on a contested boundary (Q4), some have model disagreement (Q2), and some are unstable in both spaces (Q3). A practitioner deploying such a model should inspect the predictions in Q2/Q3/Q4, and then can use the recommendations made in Section \ref{subsec:readingplot}.

In Fig. \ref{fig:binary_uci_bank} Banknote Authentication dataset is inherently well-separated, as the image-statistic features provide strong discrimination between genuine and forged notes. This separation is clearly reflected in the RAD Plot, where nearly all datapoints lie in Q1, with only two datapoints appearing in Q2 and Q4 (using a threshold of $0.5$). In practical deployments where even a small number of ambiguous cases may carry significant operational risk - for example, highly sophisticated counterfeits designed to closely resemble genuine notes - the threshold can be increased to flag a larger set of samples for manual review. In this setting, RAD functions less as a routine rejection mechanism and more as a calibrated early-warning or inspection prioritization tool.

For the \textit{Heart Failure} dataset, the model is trained to predict whether the patients' clinical risk factors will lead to heart failure or not. In Fig. \ref{fig:binary_uci_heart}, all ambiguous datapoints fall in Q4, where models agree on a coherent decision boundary, but a subset of patients sit on top of it. This is precisely the boundary-on-point case in Section \ref{subsec:radplot}. Clinically, these can be genuine borderline cases, or lack of patient cases in the region, patients whose risk profile places them at the decision threshold rather than evidence that of a model issue. The appropriate action is dedicated human review by an expert clinician, not model retraining. Similarly, Fig. \ref{fig:binary_uci_wisconsin} and \ref{fig:binary_uci_rice} can be interpreted.

Together, these three datasets exemplify the three non-trivial ambiguity sources separately (Heart Failure, Rice: feature-space only, Banknote: minor/well-separated; Gamma Telescope: all sources), demonstrating that RAD Plot's quadrant structure surfaces interpretable distinctions even in high-dimensional data where the boundary cannot be visualised.

\subsubsection{Multi-Class Datasets}
RAD is able to provide a richer analysis when applied to multi-class datasets, as a one-vs-all RAD plot can be generated, allowing us to inspect which classes are the most confusable. We use the \textit{Wine Quality} dataset, which scores wines from $3$ to $9$ from their chemical properties. It is expected that the scores in the middle ($5$, $6$, $7$) to overlap, and the extreme scores (3, 9) to be well-separated, since intermediate quality is a more subjective opinion than a "clearly bad" or a "clearly excellent".

\begin{figure}[htbp]
\centering

\subfloat[\textit{3/10}]{\includegraphics[width=0.25\textwidth]{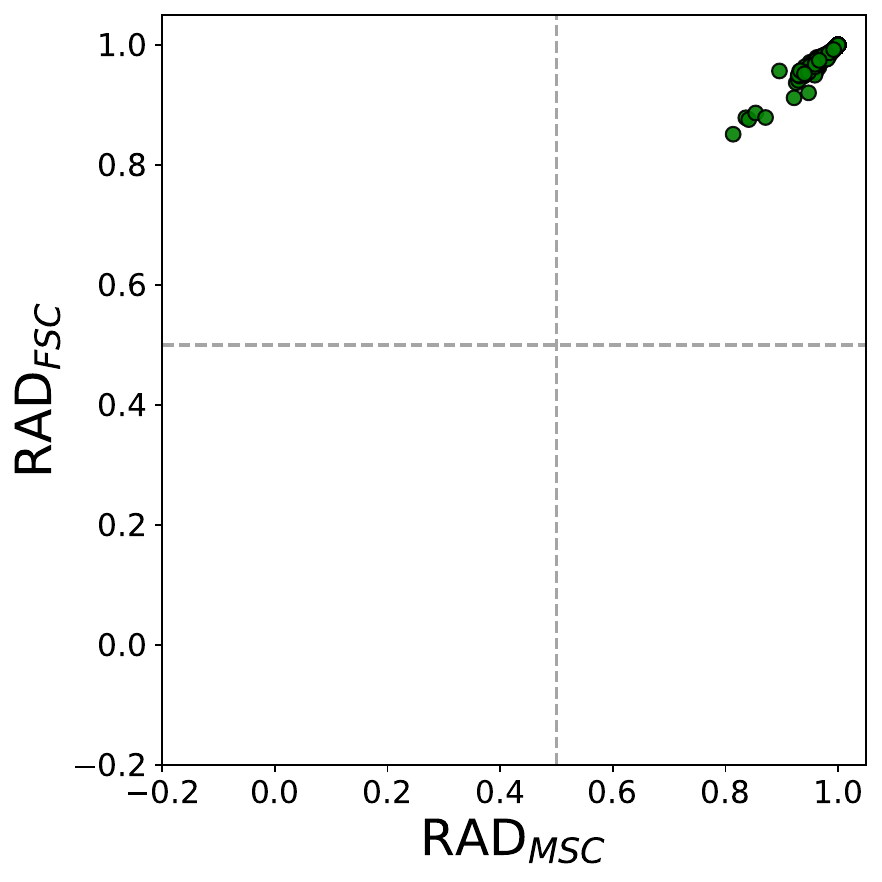}}
\subfloat[\textit{4/10}]{\includegraphics[width=0.25\textwidth]{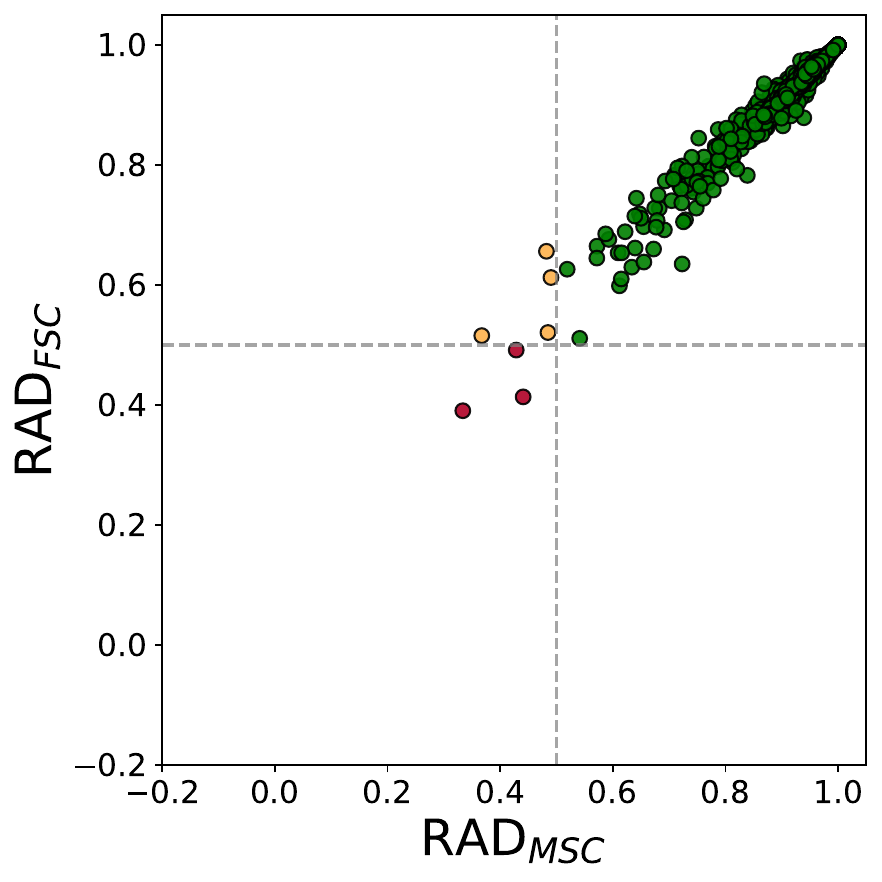}}
\subfloat[\textit{5/10}]{\includegraphics[width=0.25\textwidth]{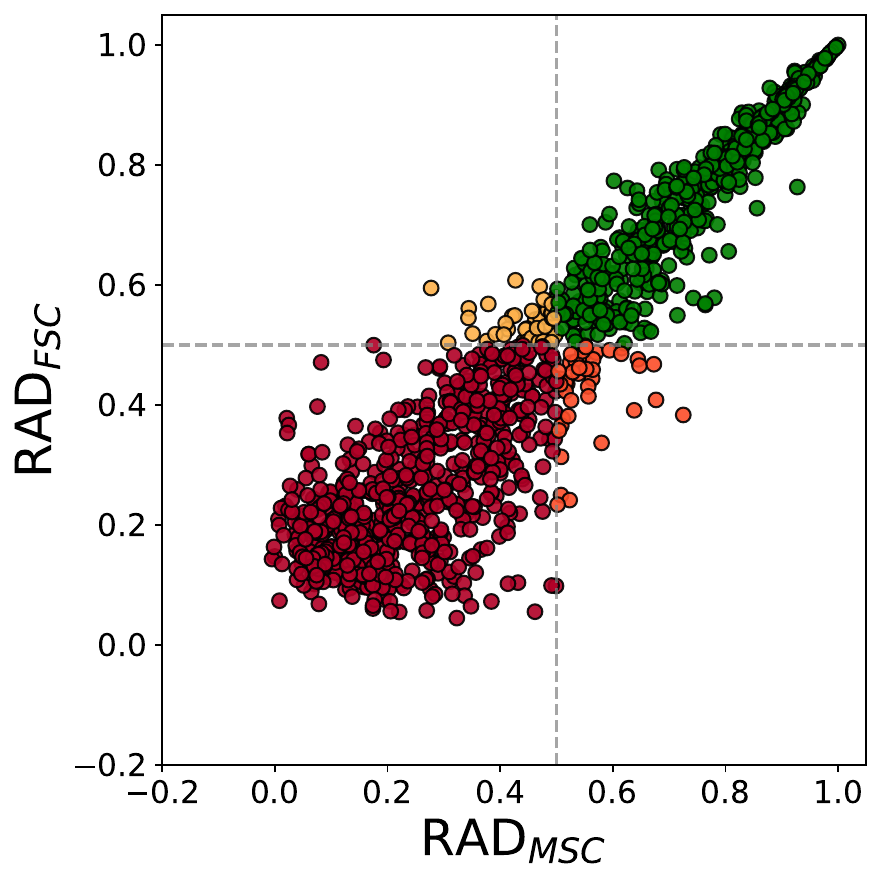}}
\subfloat[\textit{6/10}]{\includegraphics[width=0.25\textwidth]{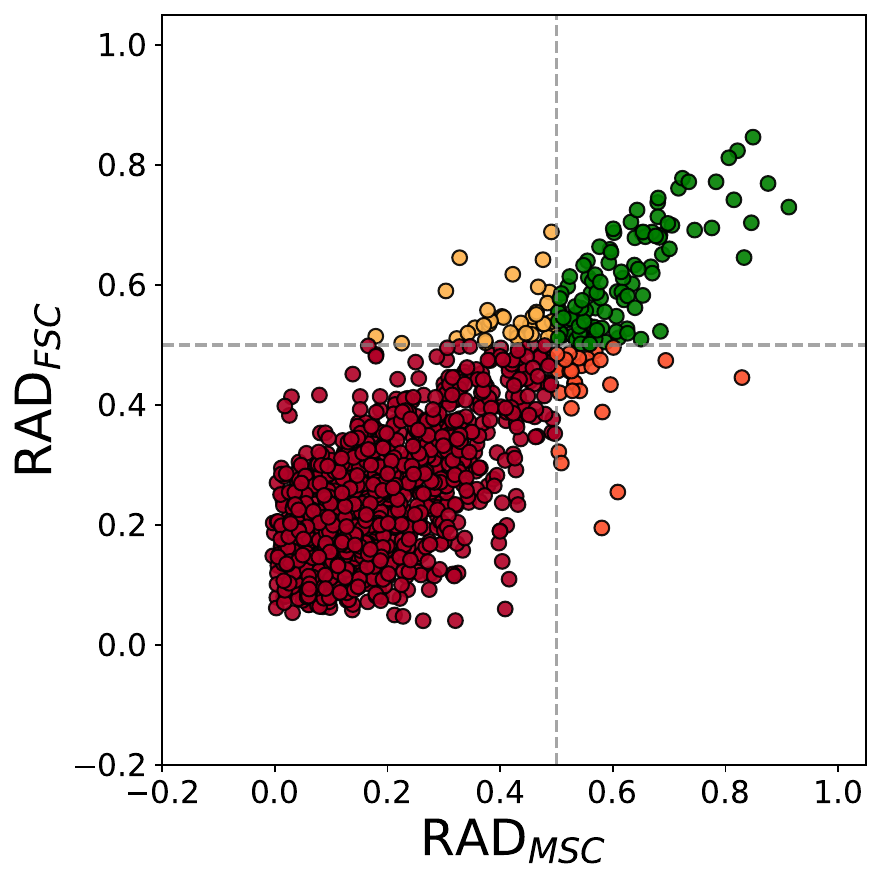}}

\subfloat[\textit{7/10}]{\includegraphics[width=0.25\textwidth]{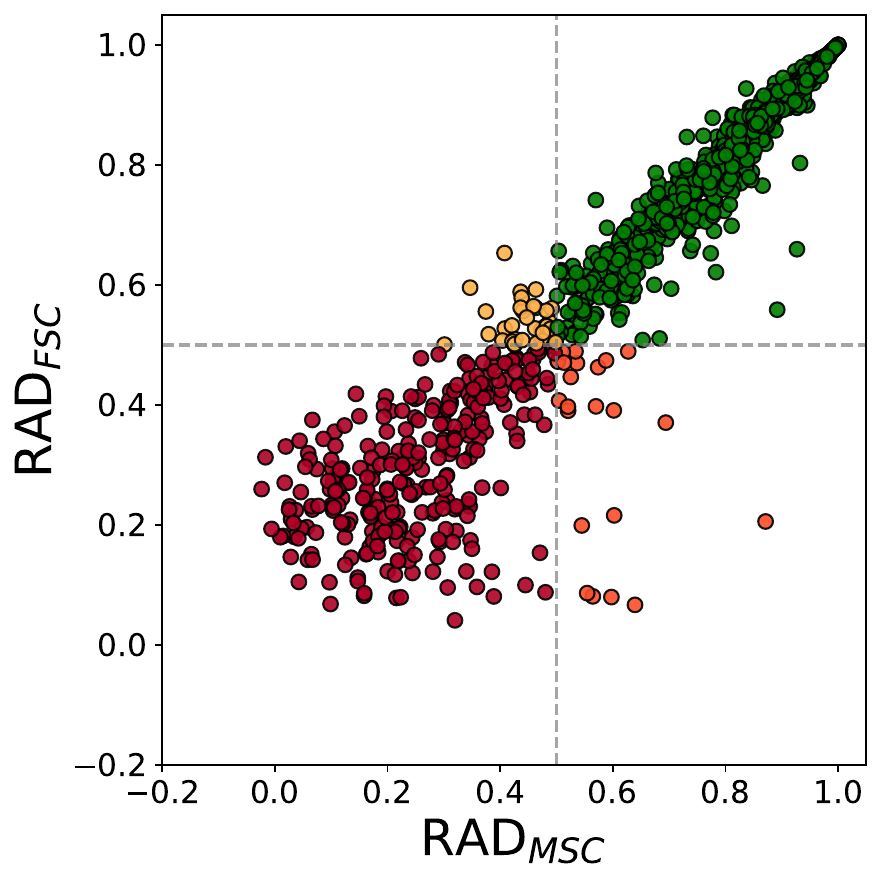}}
\subfloat[\textit{8/10}]{\includegraphics[width=0.25\textwidth]{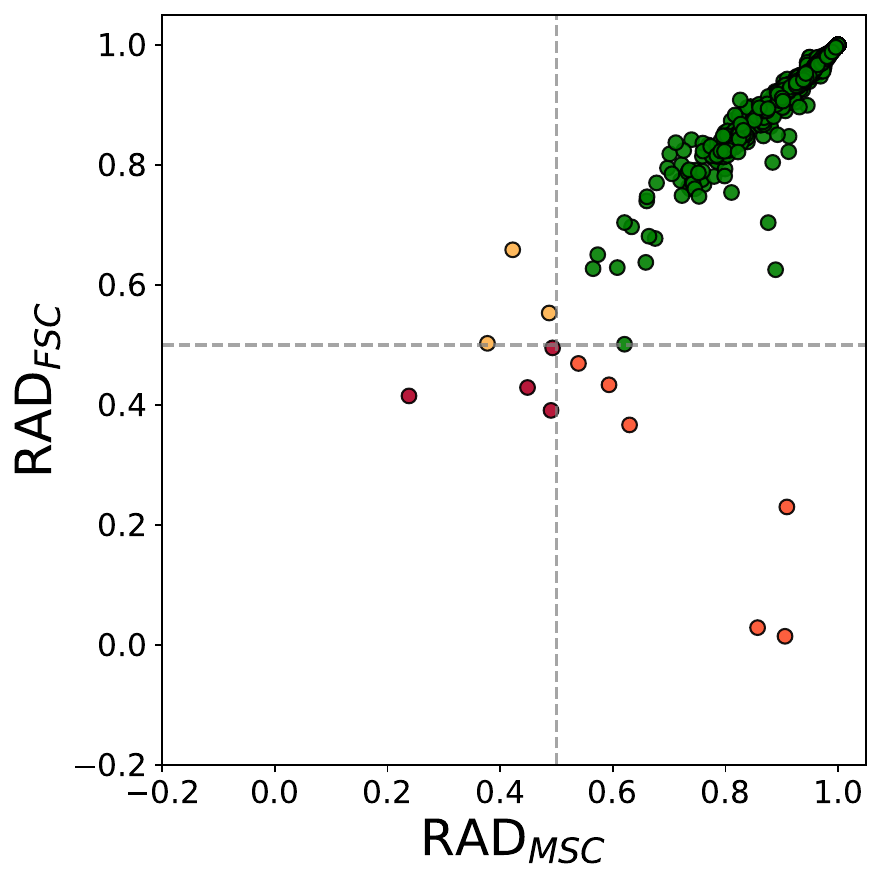}}
\subfloat[\textit{9/10}]{\includegraphics[width=0.25\textwidth]{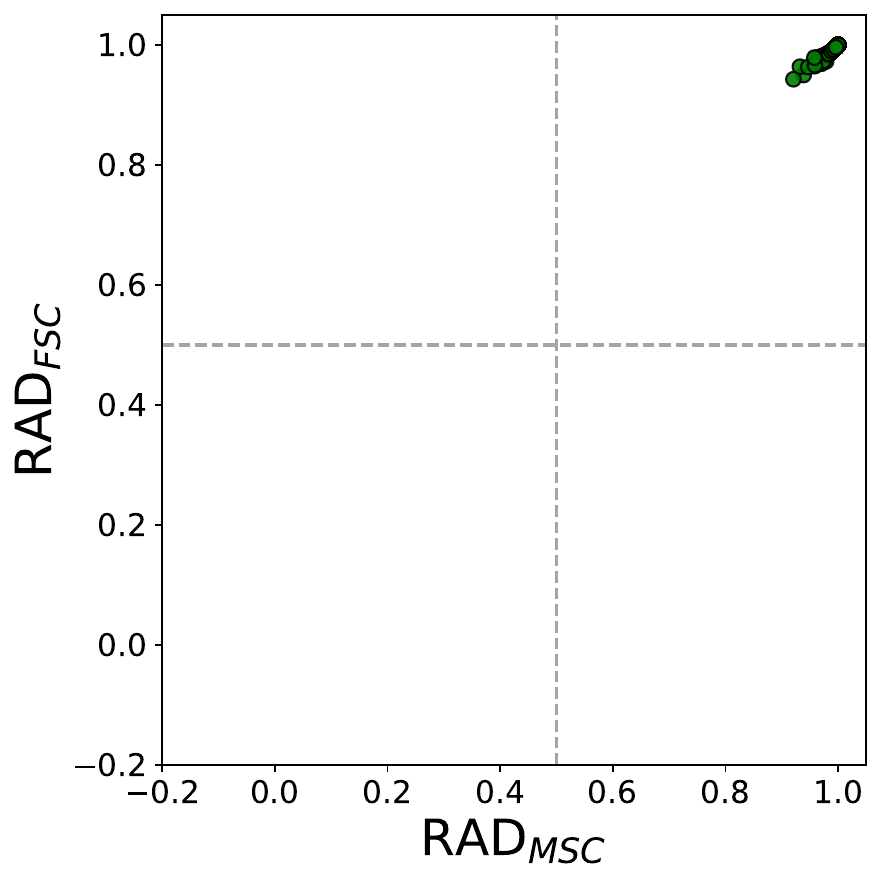}}

\caption{Multi-class dataset \textit{Wine Quality} class-wise RAD Plots. Wine quality scores in the middle are the most confusable classes.}
\label{fig:uci_multiclass}
\end{figure}

\begin{figure}[htbp]
\centering

\subfloat[\textit{Class 0}]{\includegraphics[width=0.2\textwidth]{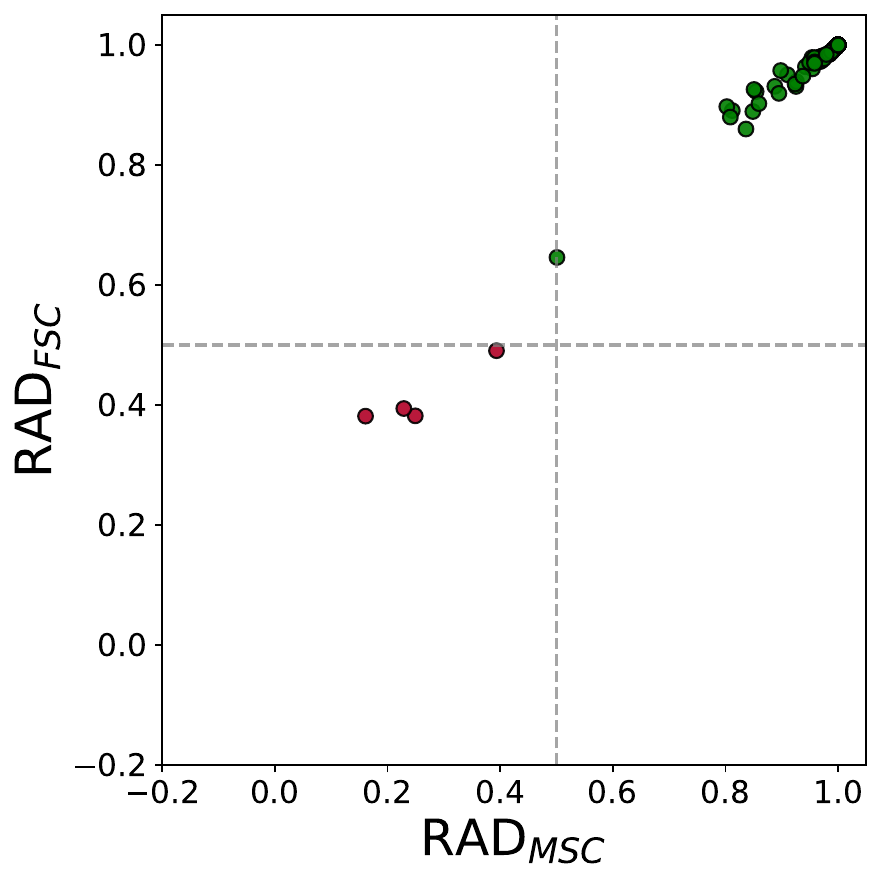}}
\subfloat[\textit{Class 1}]{\includegraphics[width=0.2\textwidth]{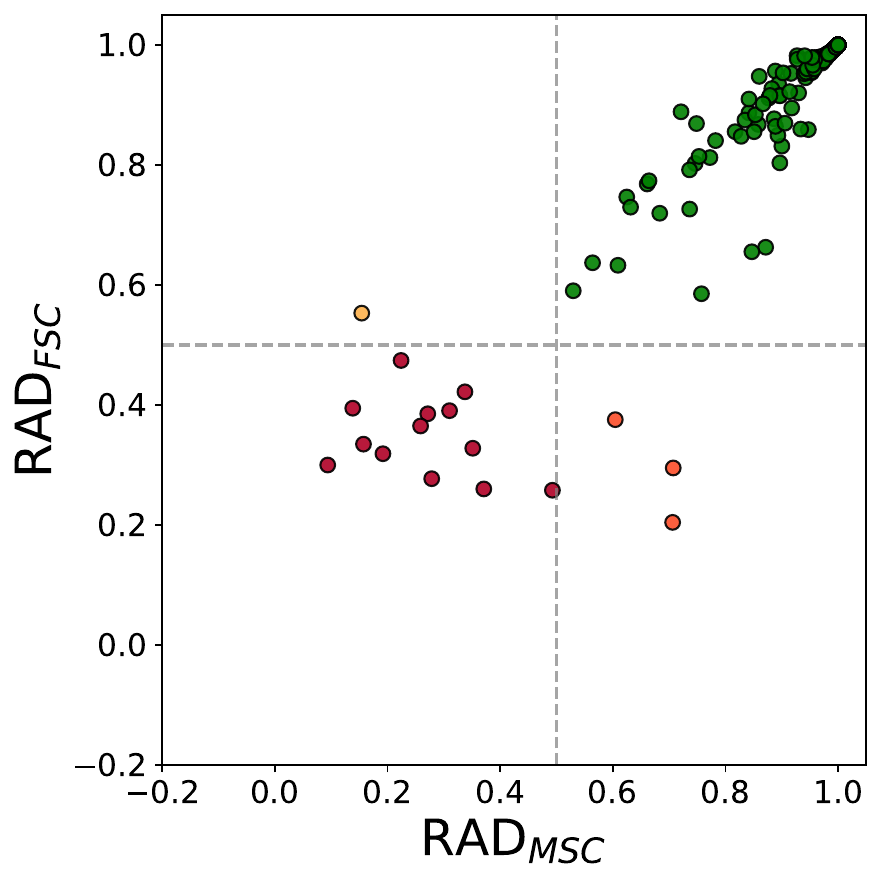}}
\subfloat[\textit{Class 2}]{\includegraphics[width=0.2\textwidth]{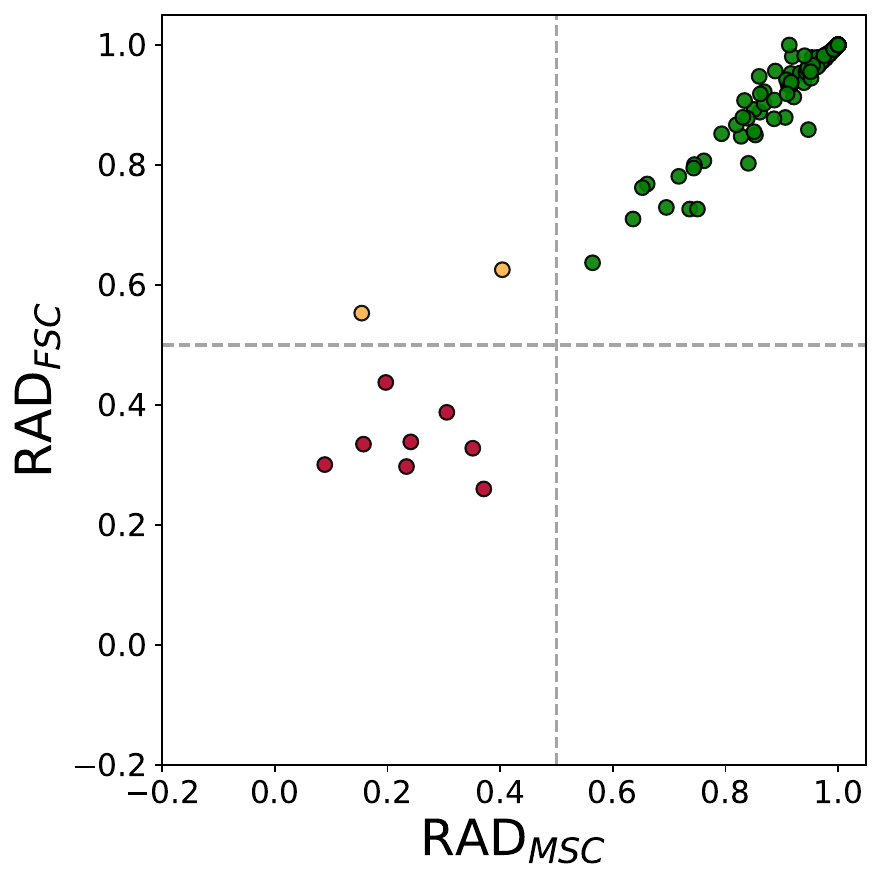}}
\subfloat[\textit{Class 3}]{\includegraphics[width=0.2\textwidth]{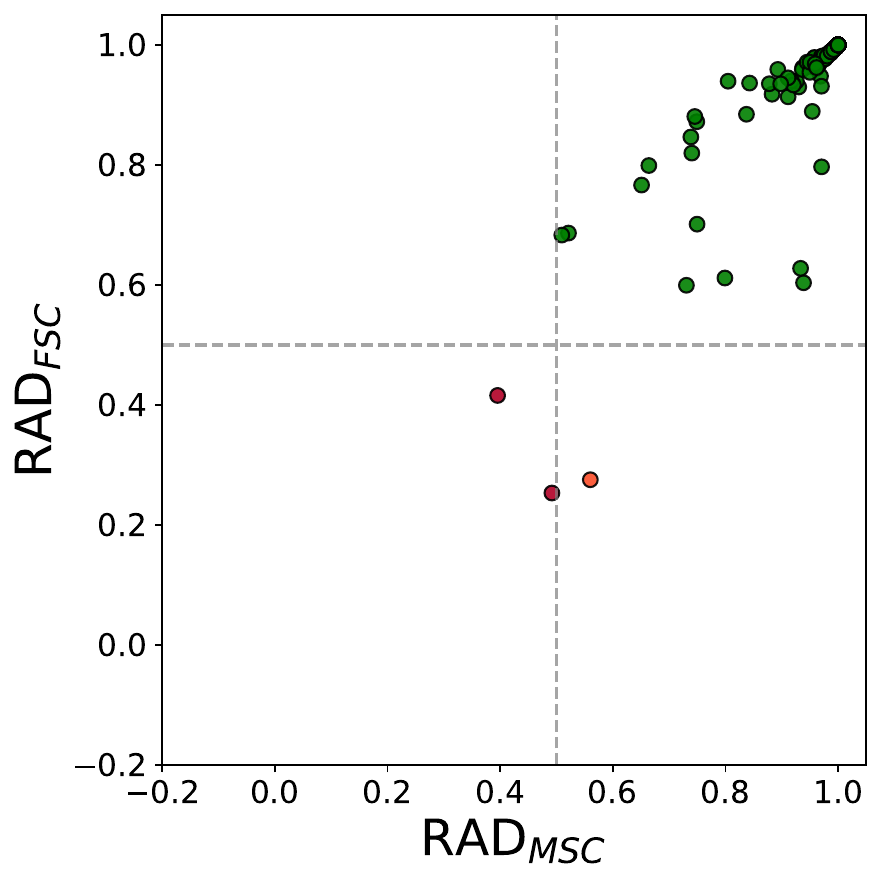}}
\subfloat[\textit{Class 4}]{\includegraphics[width=0.2\textwidth]{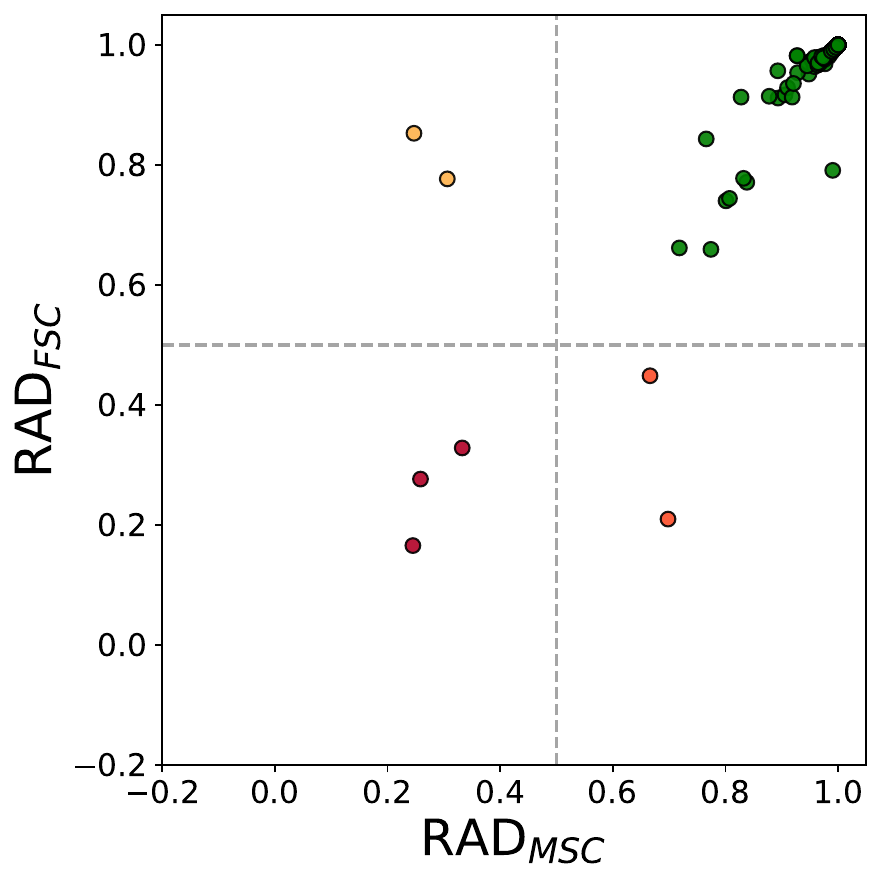}}
\\
\subfloat[\textit{Class 5}]{\includegraphics[width=0.2\textwidth]{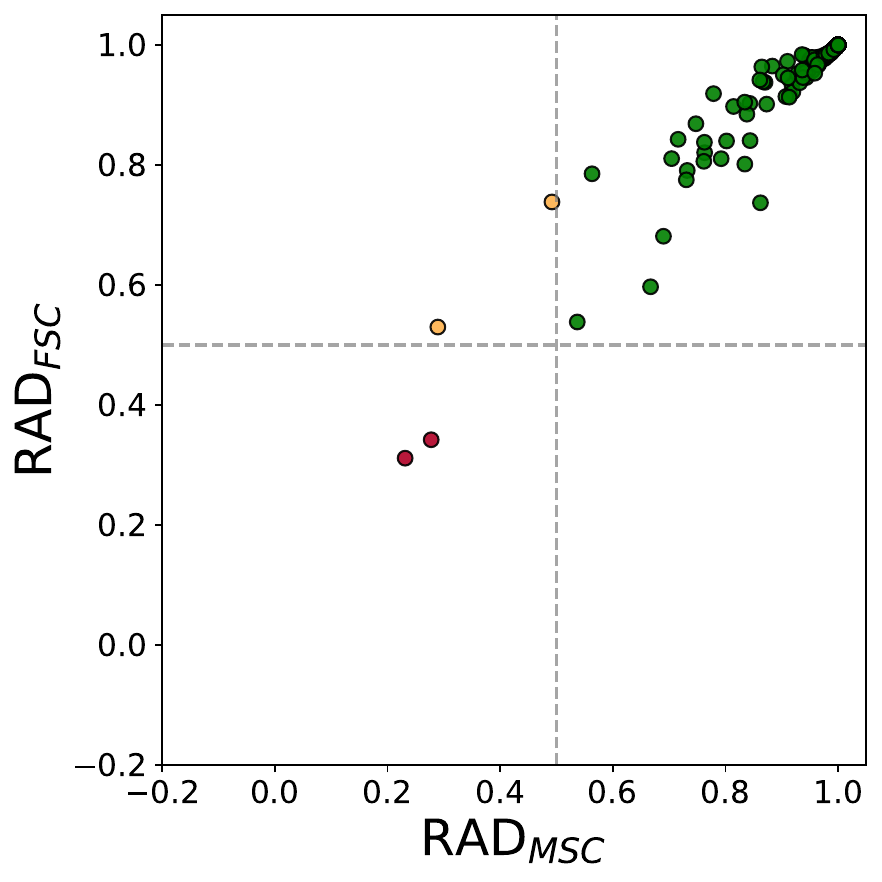}}
\subfloat[\textit{Class 6}]{\includegraphics[width=0.2\textwidth]{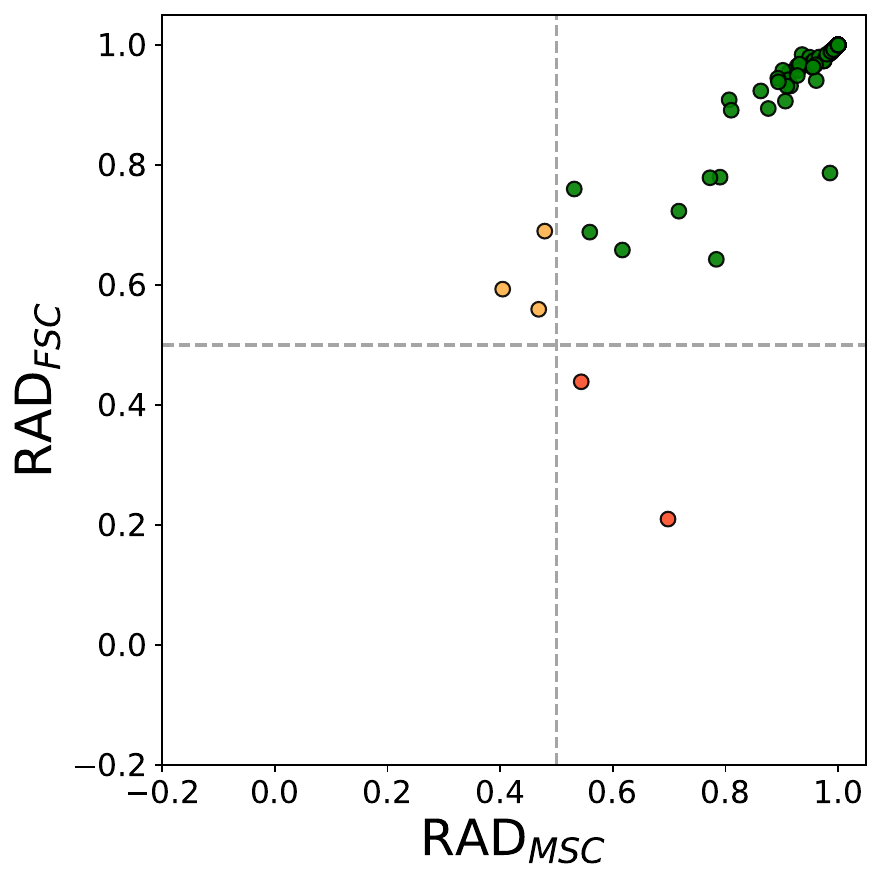}}
\subfloat[\textit{Class 7}]{\includegraphics[width=0.2\textwidth]{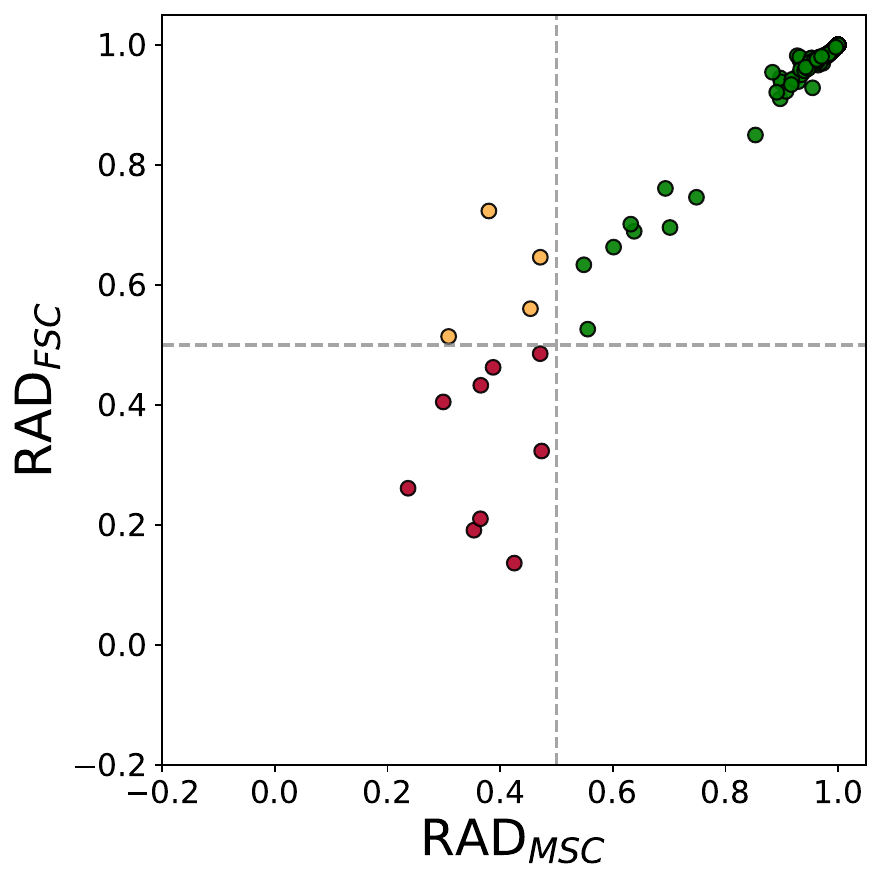}}
\subfloat[\textit{Class 8}]{\includegraphics[width=0.2\textwidth]{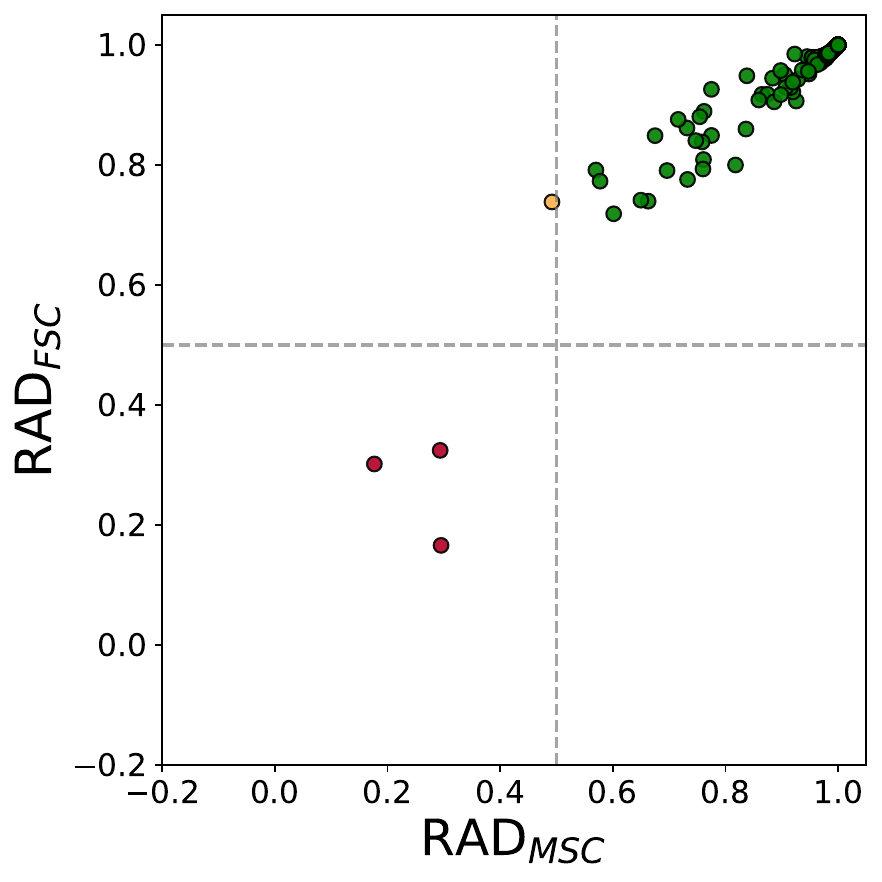}}
\subfloat[\textit{Class 9}]{\includegraphics[width=0.2\textwidth]{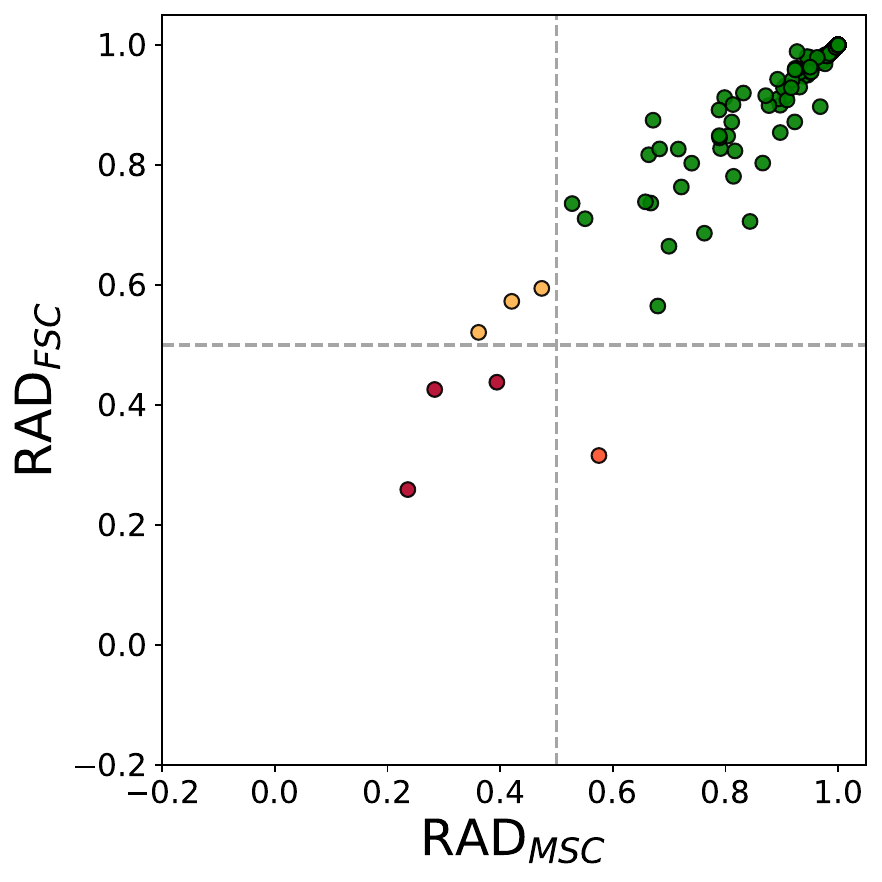}}

\caption{Multi-class dataset \textit{Handwritten Digit Recognition} class-wise RAD Plots. Spread patterns show, digits 1, 2 and 7 are the most confusable classes.}
\label{fig:hand_digits}
\end{figure}

The per-class RAD Plots (Fig. \ref{fig:uci_multiclass}) precisely show this phenomenon. Score classes $3$ and $9$ contain no ambiguous datapoints, and RAD shows them as cleanly separable. Scores $5$, $6$, and $7$ show substantial mass in Q3, indicating both model disagreement and local instability are present. This is the \textit{diagonal-spread} signature from Section \ref{subsec:readingplot}, which indicates that there might be label-noise or truly ambiguous class definitions. Both interpretations can apply here. Human raters scoring wines on a 10-point scale produce labels that are intrinsically noisy in the middle range, and the class definitions themselves (difference between $6$ and $7$) are not crisply defined. This indicates that the task is genuinely ambiguous at the middle of the quality scale, and that a coarser class taxonomy (e.g., low/medium/high) could be considered. Similarly, in Fig. \ref{fig:hand_digits} the spread patterns show in \textit{Handwritten Digit Recognition} dataset, the digit $1$, $2$ and $7$ are the most ambiguous ones.

This per-class view is one of RAD's very useful diagnostic outputs for multi-class problems, which a single accuracy number could not have uncovered.

\subsection{RAD Score-Pair Component Analysis}
We investigate if RAD Score-Pair's components $\mathrm{RAD}_{\mathrm{MSC}}$ and $\mathrm{RAD}_{\mathrm{FSC}}$, have independent informative or redundant signals. We test this directly across all test-point across all datasets. When every test datapoint is pooled over all datasets, $\mathrm{RAD}_{\mathrm{MSC}}$ and $\mathrm{RAD}_{\mathrm{FSC}}$ are moderately correlated ($\rho = 0.61$), but this is actually due to Q1 has most of the datapoints, where the equivalent models and neighbourhood mostly agree. $94.7\%$ of synthetic and $83.4\%$ of real-world dataset points has $\mathrm{RAD}_{\mathrm{MSC}} \approx \mathrm{RAD}_{\mathrm{FSC}} \approx 1$ (most datapoints are not ambiguous). Restricting to only the \emph{ambiguous} quadrants, Q2/Q3/Q4, indicates a Spearman $\rho = -0.23$ on the synthetic ambiguous subset and $\rho = +0.17$ on the real-world ambiguous subset (to be interpreted as a descriptive indication, as a small number of points are ambiguous). Among the points where the framework is useful for identifying ambiguity, $\mathrm{RAD}_{\mathrm{MSC}}$ and $\mathrm{RAD}_{\mathrm{FSC}}$ are essentially uncorrelated. This demonstrates that the components of the RAD Score-Pair carry independent information.

\subsection{Abstention Results}\label{subsec:result_abstain}

\begin{table}[t]
\centering
\caption{Summary of the abstention experiment, compared against RAD-Pareto. The W/L/T row reports the number of datasets on which RAD-Pareto wins/loses/ties with the method in the corresponding column. The Avg.\ rank row reports the mean rank across methods on that dataset group (lower is better).}
\label{tab:wlt_summary}
\small
\setlength{\tabcolsep}{4.5pt}
\renewcommand{\arraystretch}{1.15}
\begin{tabular}{@{}l c | c c c | c c@{}}
\toprule
 & RAD-Pareto & Random & Entropy & SC
 & $\mathrm{RAD}_{\mathrm{MSC}}$ & $\mathrm{RAD}_{\mathrm{FSC}}$ \\
\midrule
\multicolumn{7}{@{}l}{\textit{Synthetic datasets ($N=15$)}}\\
\hline
W/L/T vs Pareto & ---       & 11/0/4 & 11/0/4 & 7/3/5  & 7/3/5  & 5/4/6  \\
Average rank        & \textbf{2.47} & 5.27       & 4.47       & 3.23       & 2.87       & 2.70       \\
\midrule
\multicolumn{7}{@{}l}{\textit{Real-world datasets ($N=17$)}}\\
\hline
W/L/T vs Pareto & ---       & 17/0/0 & 17/0/0 & 11/6/0 & 11/6/0 & 12/4/1 \\
Average rank        & \textbf{1.97} & 5.24       & 5.06       & 3.24       & 2.71       & 2.79       \\
\bottomrule
\end{tabular}
\end{table}

We evaluate the abstention experiment mentioned in Section \ref{sec:experiments} on the synthetic and Real-world datasets separately. The complete per-dataset AURC table in Appendix~\ref{sec:appendix}. Here we summarise the results focusing on demonstrating the feasibility of using RAD in a downstream task.

\begin{figure}[htbp]
  \centering
  \includegraphics[width=0.75\textwidth]{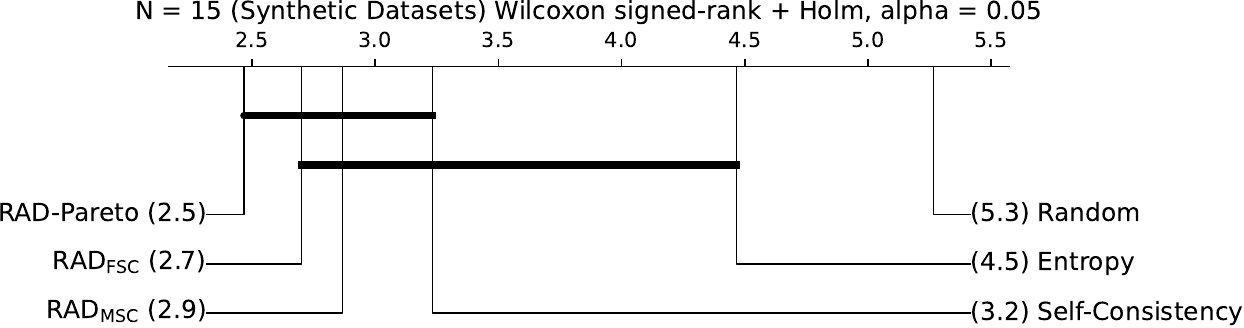}\\
  \vspace{1cm}
  \includegraphics[width=0.75\textwidth]{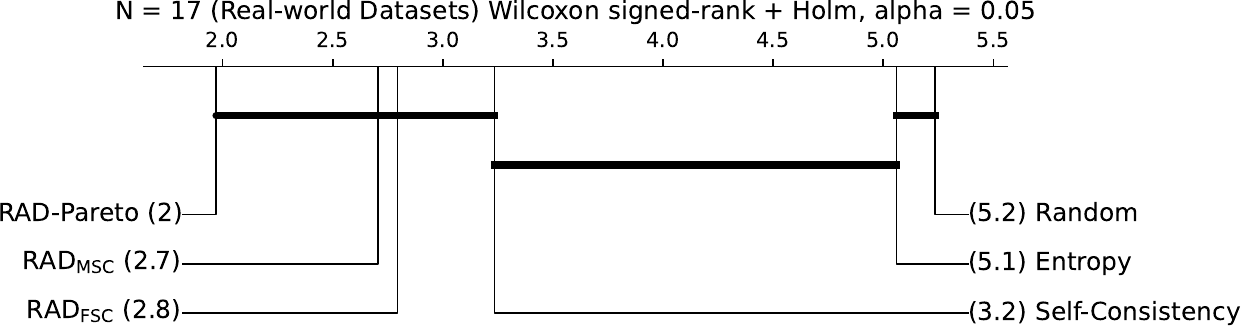}
  \caption{Critical-difference diagrams summarising rejection AURC across the Synthetic datasets ($N=15$, top) and Real-world datasests ($N=17$, bottom). Methods are arranged by mean rank (lower is better). A bar connecting a group of methods indicates that they are not significantly different by pairwise Wilcoxon signed-rank test with Holm correction at $\alpha = 0.05$.}
  \label{fig:cdplots}
\end{figure}

Figure \ref{fig:cdplots} summarises pairwise comparisons across all methods using the Wilcoxon signed-rank test with Holm correction \cite{demvsar2006statistical}. RAD-Pareto achieves the lowest mean rank in both types of datasets ($2.47$ in synthetic, $1.97$ in UCI). It is statistically indistinguishable from the other RAD components and Self-Consistency, but significantly better than Entropy and Random. This result demonstrates that the RAD Score-Pair, ranked through RAD Pareto-Rank, produces a rejection score that performs at least as well as standard model-disagreement baselines while introducing the interpretability of the quadrant diagnosis.

\section{RAD Use-cases}\label{sec:deployment}
In Section \ref{subsec:result_abstain}, we demonstrate a downstream abstention task. The RAD Plot also can be used in several deployment time uses. Some possible use cases are outlined below.

\begin{flushleft}
    \textbf{Distinguishing data drift from model staleness.} In production, new incoming datapoints can be scored against the deployed models and placed on the RAD Plot alongside the original test/train distribution. A systematic shift toward Q3/Q4 while the reference remains in Q1 indicate a possible \emph{data-drift} \cite{gama2014survey}, incoming data has moved into regions where deployed models are unstable. Points clustering in Q2/Q3 with degrading labelled-sample accuracy indicates a possible \emph{concept-drift} \cite{gama2014survey}. Combined with rolling predictive performance, the RAD Plot makes a stronger drift diagnostic than individual signals.
\end{flushleft}

\begin{flushleft}
    \textbf{Quality control for retrained models.} When a deployed model is retrained, the RAD Plot can serve as an acceptance test. A practitioner can define a threshold, for example, "at most 5\% of a held-out reference set may fall in Q3", and reject retrained models that satisfy accuracy targets but violate this threshold. This treats predictive ambiguity as an important deployment criterion along with accuracy and calibration, particularly relevant in high-stakes domains where regressions in predictive consistency are costly even when the accuracy is preserved. 
\end{flushleft}

\begin{flushleft}
    \textbf{Insurance policy retention.} An insurance company can score their customers through RAD and identify the ambiguous points, which are the customers who are unsure about if they will renew or not. Ranking them and contacting them for a better offer to retain them. Such a ranked approach can be effective as it can maximise the retention rate under a budget.
\end{flushleft}

\section{Conclusion}\label{sec:conclusion}
We defined \emph{robust-ambiguity} as predictive inconsistency across both model-space and feature-space simultaneously. Prior work on predictive multiplicity quantifies model-space disagreement at a single point~\cite{cooper2024arbitrariness,marx2020predictive, hsu2022rashomon}, and prior work on local robustness quantifies feature-space stability of a single model~\cite{leino2021relaxing, zhong2021understanding}. Neither captures the interaction. A datapoint can be predictively consistent at a point yet flip under a perturbation that any reasonable practitioner would consider admissible. Conversely, models may agree at one point but disagree across an entire neighbourhood. These are distinct failure modes, and reliability assessment that addresses only one is incomplete.

The RAD framework operationalises this through the RAD Score-Pair $[\mathrm{RAD}_{\mathrm{MSC}}, \mathrm{RAD}_{\mathrm{FSC}}]$ and the RAD Plot, which together reveal not only \emph{whether} a prediction is ambiguous but \emph{why}: model disagreement versus boundary proximity versus both. Our experiments validate that RAD identifies ambiguous datapoints across synthetic and real-world datasets. Ranking these datapoints via the RAD-Pareto-Rank, which utilises the Pareto frontier, yields a downstream rejection score competitive with established uncertainty baselines.

A key property of RAD is that its scores are always relative to the deployed model class/method, that is, they identify which predictions cannot be made reliably by the models being used. The same dataset analysed with decision trees may flag different samples than with neural networks.

A more fundamental empirical finding underwrites this result: predictions in the ambiguous quadrants Q2, Q3, and Q4 are near chance-level accurate, while Q1 predictions average 0.91. This is not a property of poorly fitted models, that is, the same models that achieve high accuracy in Q1 produce near-random predictions in the ambiguous quadrants. It is the framework working as intended: RAD identifies, before any prediction is made, which datapoints the deployed model class cannot decide reliably.

RAD requires $n \times p$ predictions per test sample, which can be expensive at scale. We focused on tree-based methods here to demonstrate the framework, but extending to deep networks, making RAD work on predicted scores instead of class labels, and to regression is natural future work. The SMOTE-style perturbation strategy requires some domain adaptation for structured input types (audio, text), but the framework itself is type-agnostic.


\bibliographystyle{plainnat}
\bibliography{references}

\newpage
\appendix

\section{Appendix}\label{sec:appendix}
\subsection{Experiment Set Up}

\begin{table}[h!]
\centering
\footnotesize
\caption{Synthetic and real datasets with geometric construction details}
\renewcommand{\arraystretch}{1.25}
\resizebox{\textwidth}{!}{
\begin{tabular}{l|l|l|p{5.6cm}|l}
\toprule
\textbf{Dataset} & \textbf{Lib} & \textbf{Method} & \textbf{Geometry / Construction} & \textbf{Params} \\
\midrule
Blobs
& sklearn
& Gaussian mixture
& Two isotropic Gaussian clusters centered at $(-5,-5)$ and $(5,5)$ forming linearly separable regions with tunable overlap via variance
& $\sigma = 2\cdot overlap$ \\

CustomBlobs
& numpy
& Radial Gaussians
& Four Gaussian clusters placed on a circle around the origin. Cluster centers are computed using polar angles (0°, 120°, 240°) with distance inversely proportional to overlap, controlling inter-class mixing
& spread, overlap$_{ij}$ \\

Spirals
& numpy
& Parametric curves
& Two interleaving logarithmic-like spirals generated from polar coordinates where radius increases with angle; second spiral is mirrored and perturbed, creating tightly intertwined nonlinear manifolds
& noise, overlap \\

Checkerboard
& numpy
& Grid partitioning
& Uniform sampling in a bounded square domain partitioned into a grid; class labels alternate based on parity of integer cell indices $(\lfloor x \rfloor + \lfloor y \rfloor)\bmod 2$
& $k$ squares, noise \\

Moons
& sklearn
& Manifold learning toy data
& Two interleaving half-circle manifolds in 2D Euclidean space forming a nonlinear classification boundary with crescent-shaped clusters
& noise \\

Circles
& sklearn
& Concentric manifolds
& Two concentric circular decision regions where inner circle is scaled version of outer circle, producing nonlinear radial separability
& noise, factor \\

Sine
& numpy
& Functional data shift
& Two sinusoidal curves defined as $y=\sin(x)+\epsilon$ and a vertically shifted copy $y=\sin(x)+1.5+\epsilon$, forming partially overlapping functional bands
& noise \\

\midrule

COMPAS
& Hugging Face dataset
& Real-world tabular
& Structured dataset of criminal recidivism; no geometric structure, but mixed categorical features mapped into Euclidean space via label encoding
& categorical encoding \\

UCI
& ucimlrepo
& Real-world tabular
& Arbitrary feature space depending on dataset ID; no inherent geometry, but transformed into numeric feature space for learning tasks
& dataset ID \\

\bottomrule

\end{tabular}}

\end{table}

\begin{table}[h!]
\centering
\footnotesize
\caption{Hyperparameter tuning setup using 5-fold cross-validation for Decision Tree models}
\renewcommand{\arraystretch}{1.3}
\begin{tabular}{l|l|l|l}
\toprule
\textbf{Model} & \textbf{Hyperparameter} & \textbf{Search Space} & \textbf{Validation Strategy} \\
\midrule

Decision Tree
& max\_depth
& Integers $[2, 100]$
& 5-fold cross-validation (accuracy) \\

Tuning Objective
& Metric
& Accuracy
& Mean CV accuracy across folds \\

Model Selection
& Criterion
& Best mean CV score
& Selected hyperparameter = argmax score \\

\bottomrule

\end{tabular}

\end{table}

\subsection{Sensitivity Analysis}

\begin{figure}[!t]
    \centering

    \subfloat[Real Datasets (17)]{
    \includegraphics[width=0.45\linewidth]{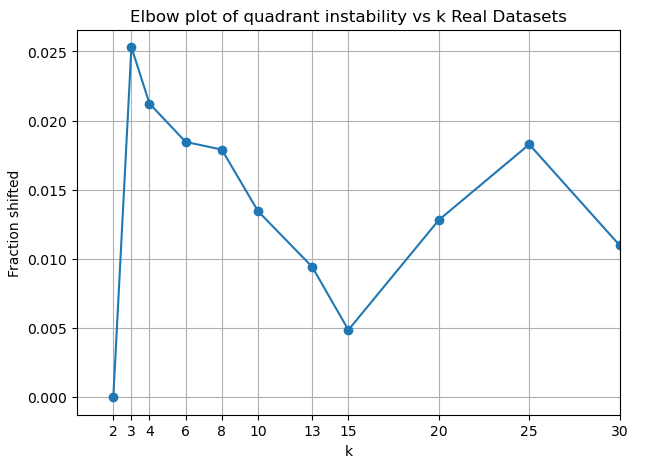}}
    \subfloat[Synthetic Datasets (15)]{
    \includegraphics[width=0.45\linewidth]{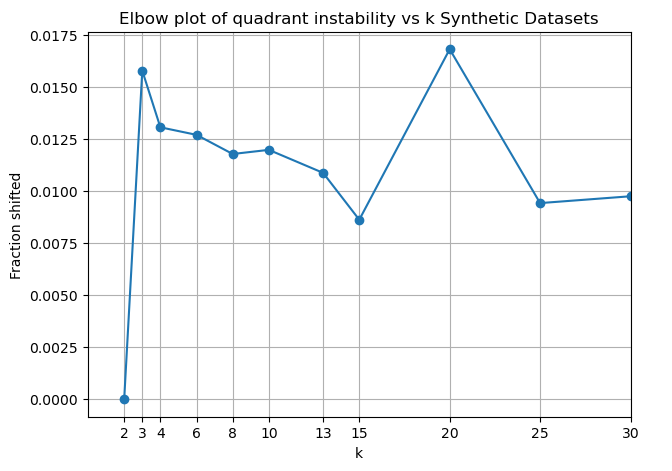}}
    \caption{Average fraction of samples that shift quadrants at different K-values across all datasets}
    \label{fig:elbow_K}
\end{figure}

\begin{flushleft}
    \textbf{RAD Stage 2 Setup}. In stage 2 of RAD, a choice needs to be made for the value of $k$ nearest neighbors to be considered when generating local synthetic samples. This choice can be made by understanding how much of an impact it has on the RAD scores first. Each instance has a RAD score pair that positions it into one of the 4 quadrants of the RAD plot. The RAD plot is generally divided into quadrants using the 0.5 threshold along each axis. The quadrant an instance belongs to, therefore, directly depends on the choice of this threshold. Considering the fixed threshold of 0.5, we computed the value of $k$ for which a data point moves from Q1 to any other quadrant. The motivation for checking this, is that if the quadrant shifts from Q1 (robust) to any other quadrant, a robust instance becomes ambiguous. The value of $k$ literally means the size of the neighborhood around which synthetic data is created, and so it is important to make sure this remains low enough to ensure the neighborhood is local and not global, and also high enough to ensure enough diverse (from across classes) data is collected. This is demonstrated in Fig. \ref{fig:elbow_K}, where the value of 10 is around which the quadrant shifts stabilise.


\end{flushleft}


\subsection{Experiment Results}
\begin{figure}[htbp]
\centering
\subfloat[Low overlap]
{\includegraphics[width=0.33\textwidth]{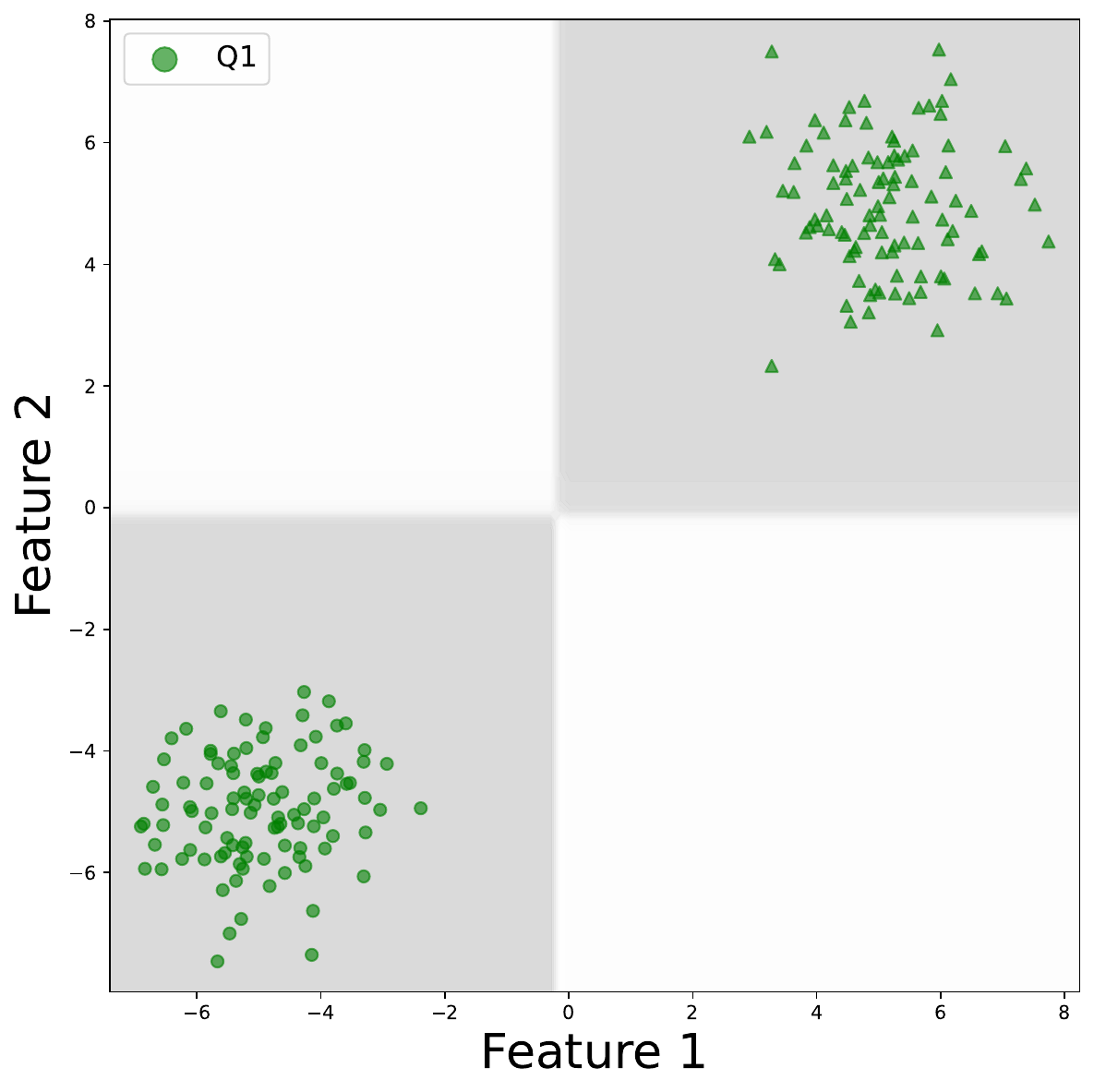}}
\hfill
\subfloat[Medium overlap]
{\includegraphics[width=0.33\textwidth]{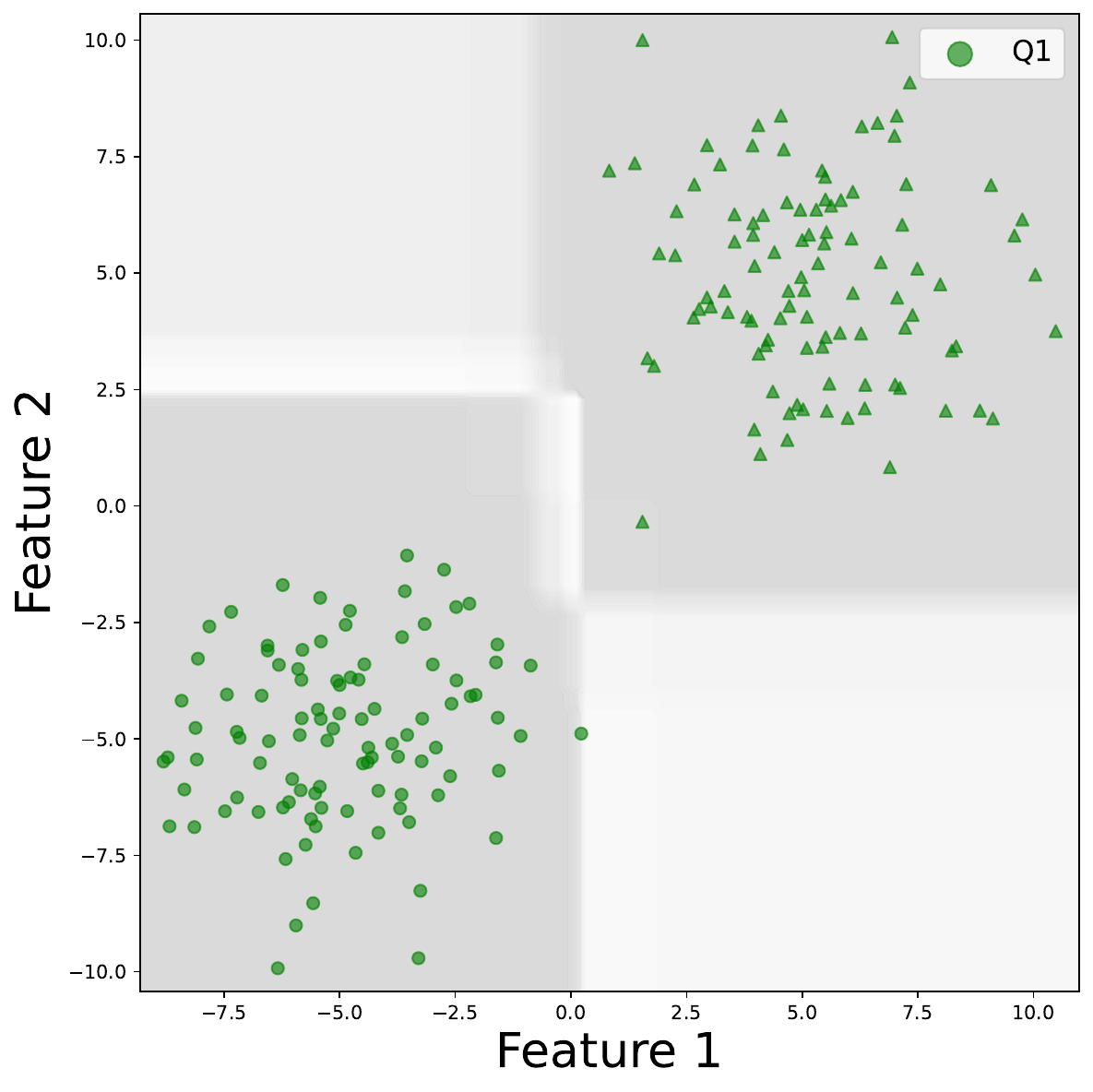}}
\hfill
\subfloat[High overlap]
{\includegraphics[width=0.33\textwidth]{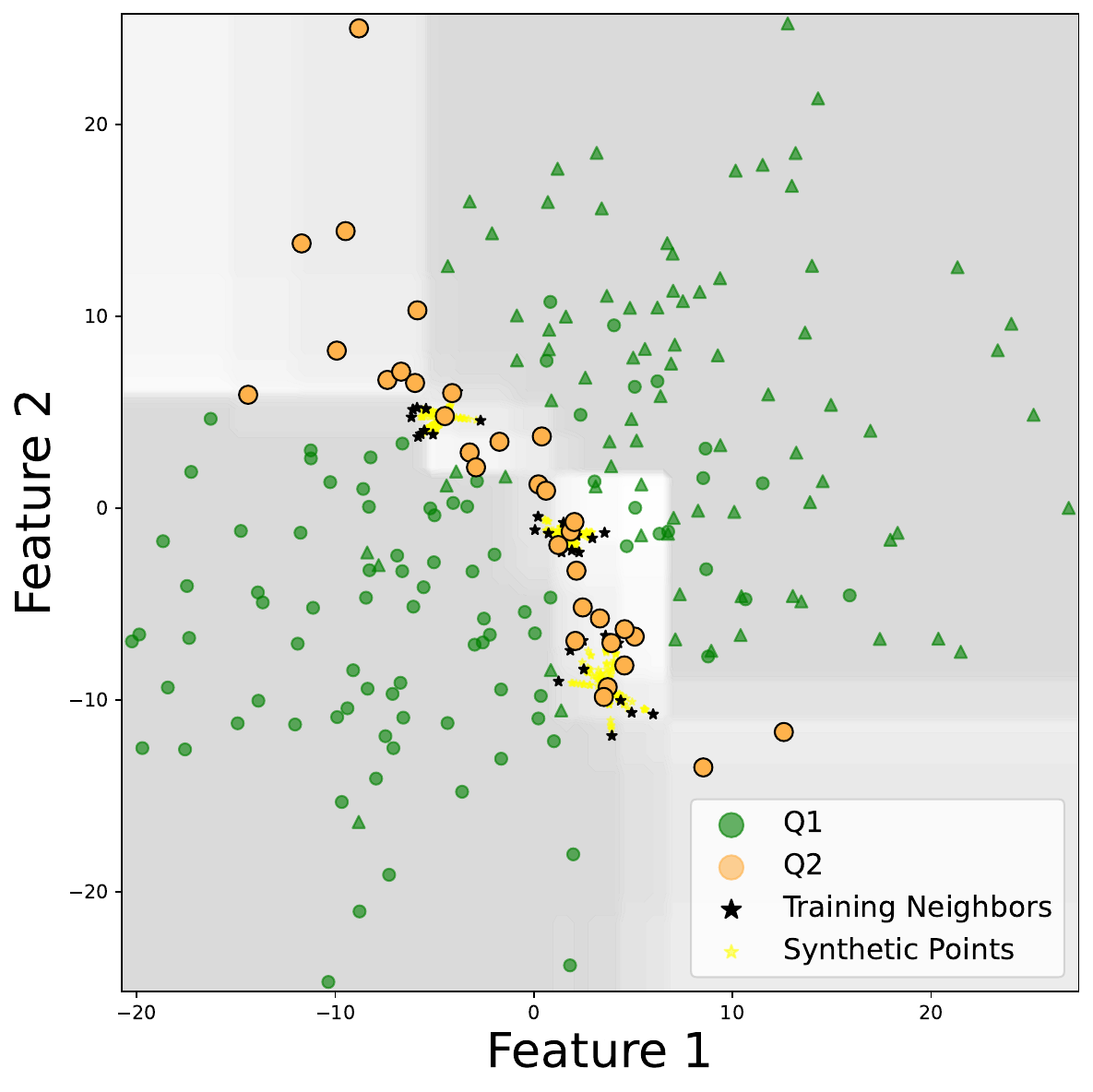}}

\subfloat[Low overlap RAD plot]{\includegraphics[width=0.33\textwidth]{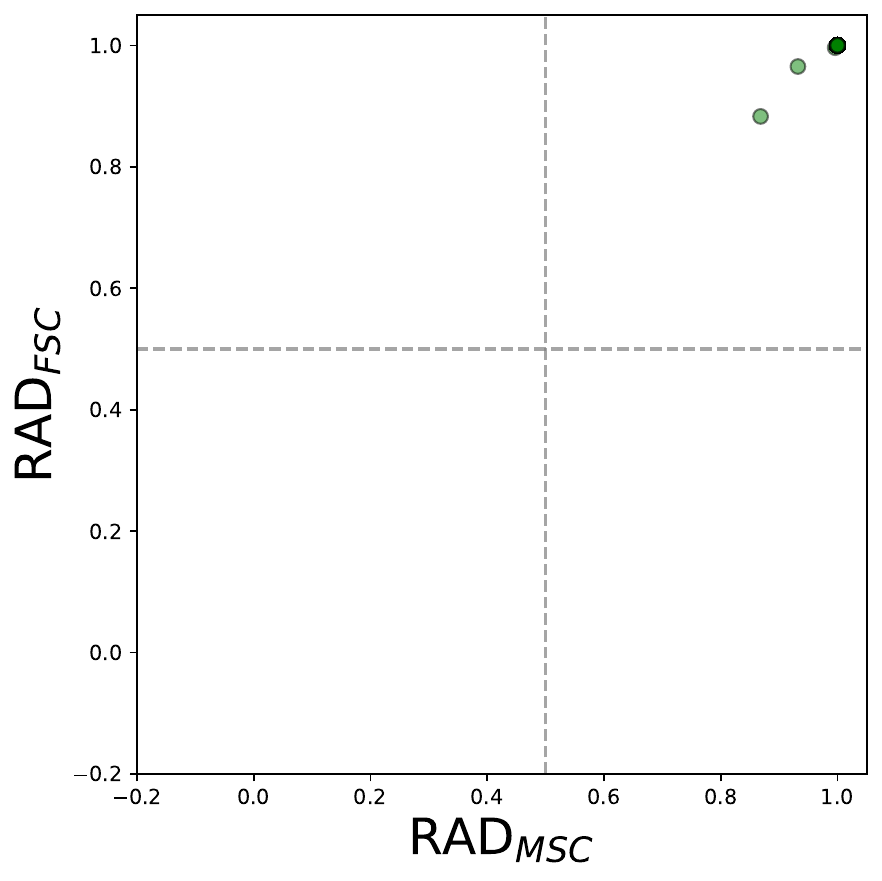}}
\hfill
\subfloat[Medium overlap RAD plot]{\includegraphics[width=0.33\textwidth]{gfx/appendix/Blobs_d1_DT_Class0_RAD_scatter.pdf}}
\hfill
\subfloat[High overlap RAD plot]{\includegraphics[width=0.33\textwidth]{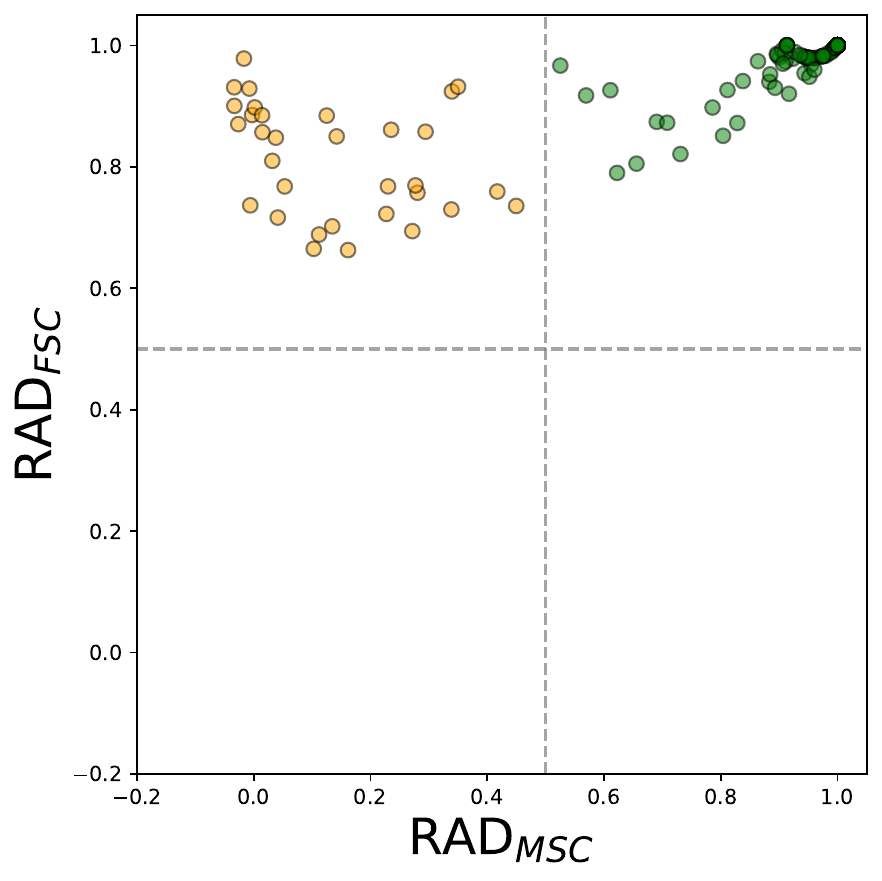}}

\caption{Blobs Dataset. Top Row: Decision boundary plots in 2-dimensions, Bottom Row: RAD Plot in 2-dimensions ($RAD_{MSC}$ and $RAD_{FSC}$).
}
\label{fig:blobs}
\end{figure}

\begin{figure}[htbp]
\centering
\subfloat[Low overlap]
{\includegraphics[width=0.33\textwidth]{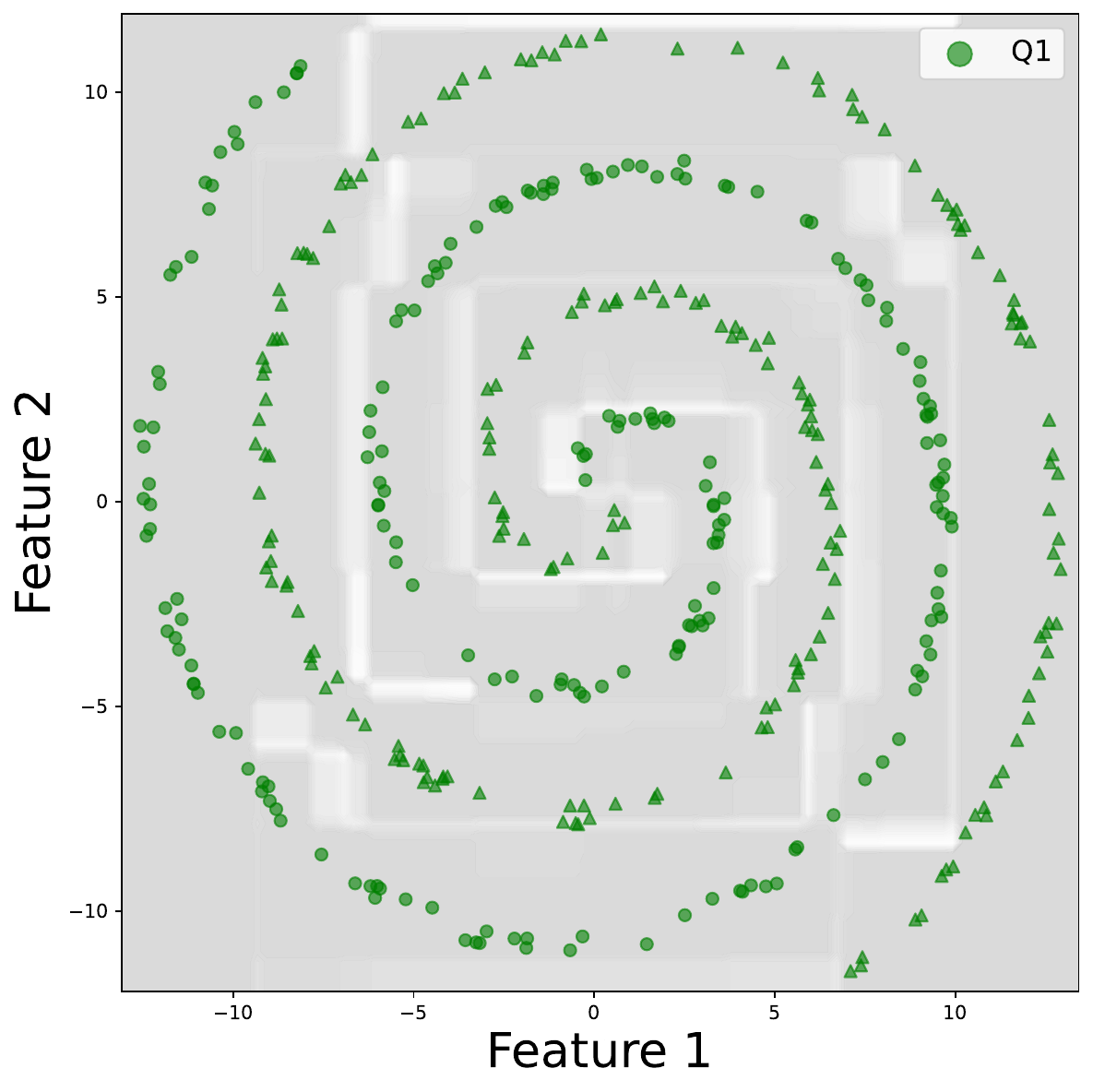}}
\hfill
\subfloat[Medium overlap]
{\includegraphics[width=0.33\textwidth]{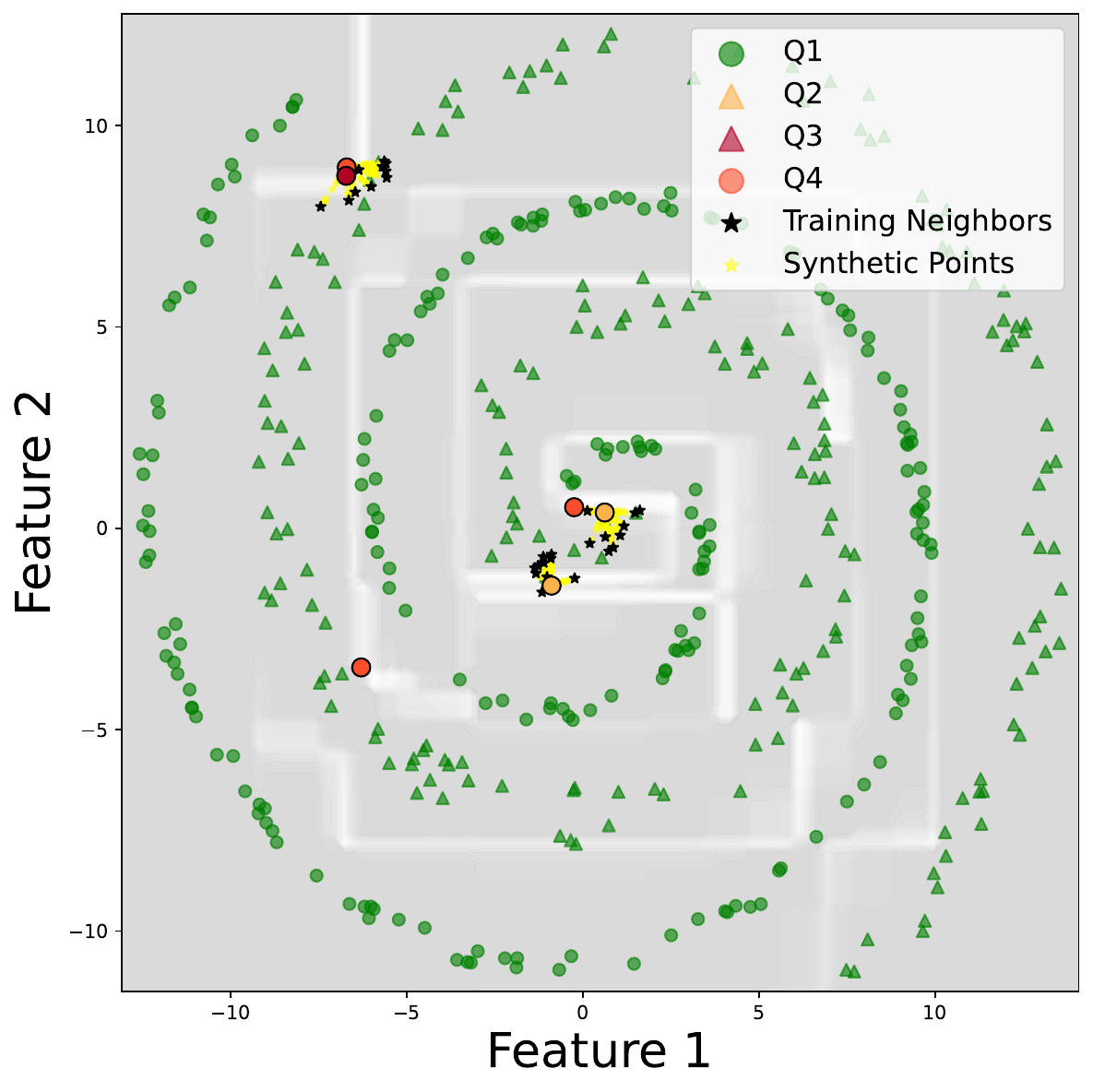}}
\hfill
\subfloat[High overlap]
{\includegraphics[width=0.33\textwidth]{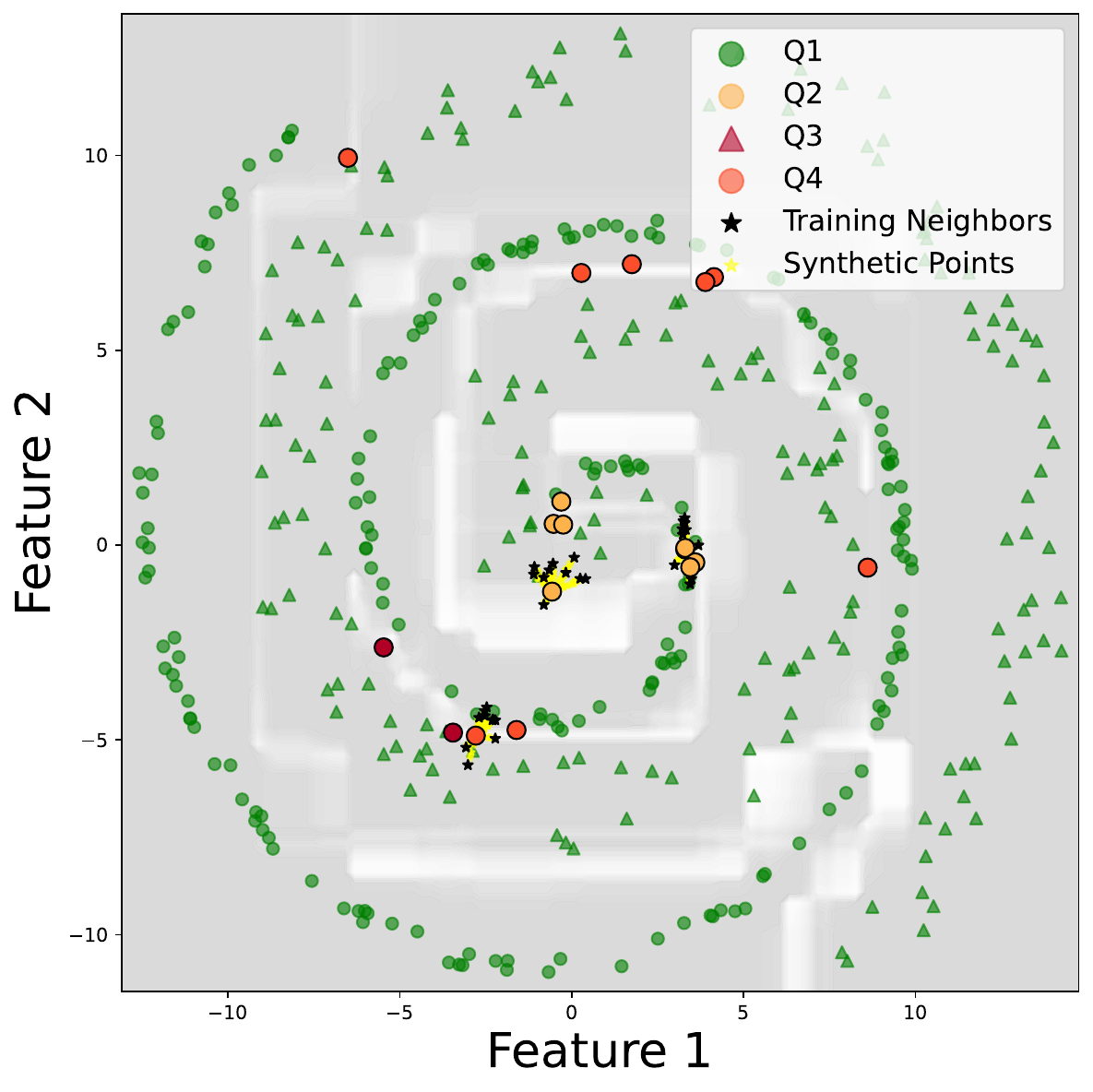}}

\subfloat[Low overlap RAD plot]{\includegraphics[width=0.33\textwidth]{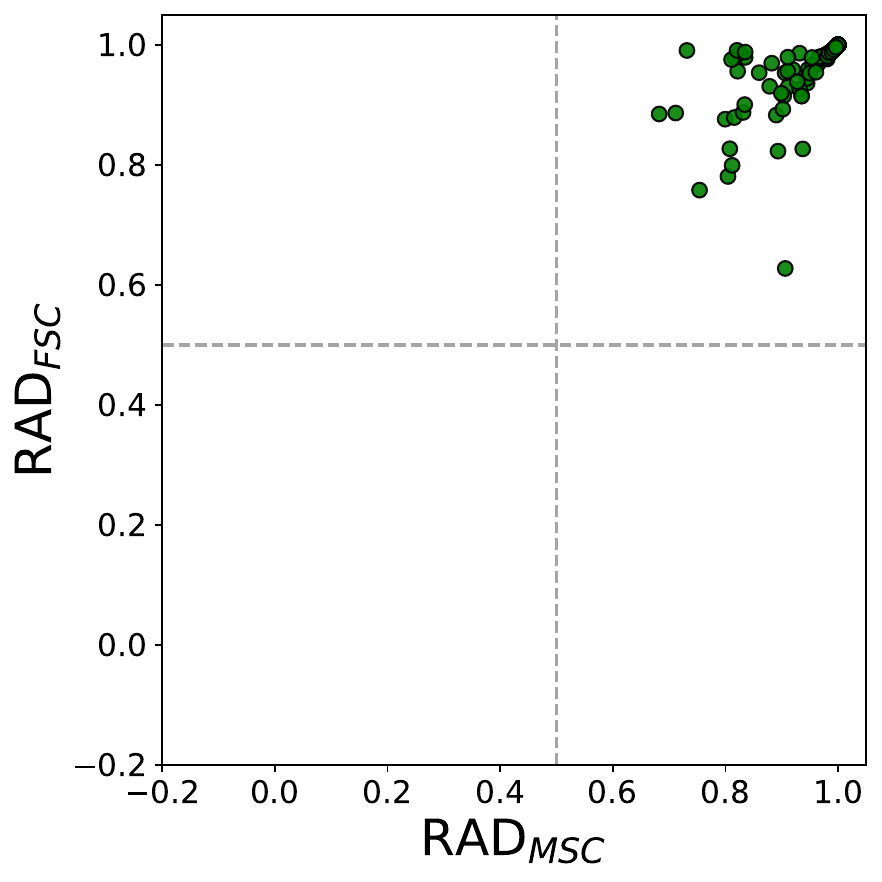}}
\hfill
\subfloat[Medium overlap RAD plot]{\includegraphics[width=0.33\textwidth]{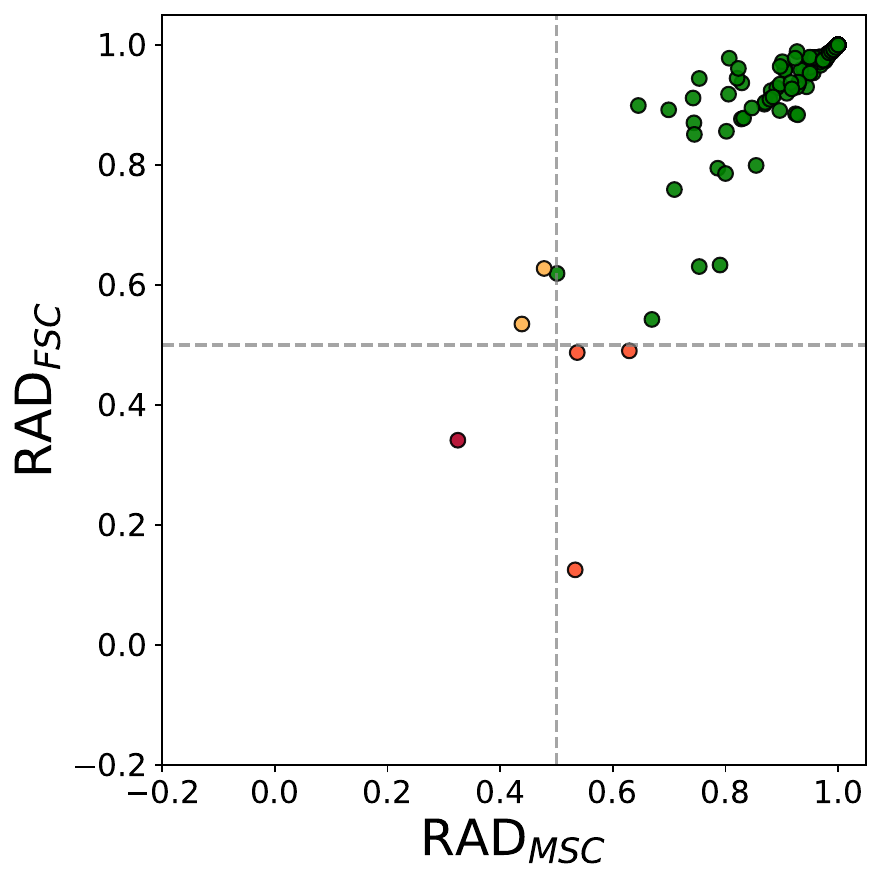}}
\hfill
\subfloat[High overlap RAD plot]{\includegraphics[width=0.33\textwidth]{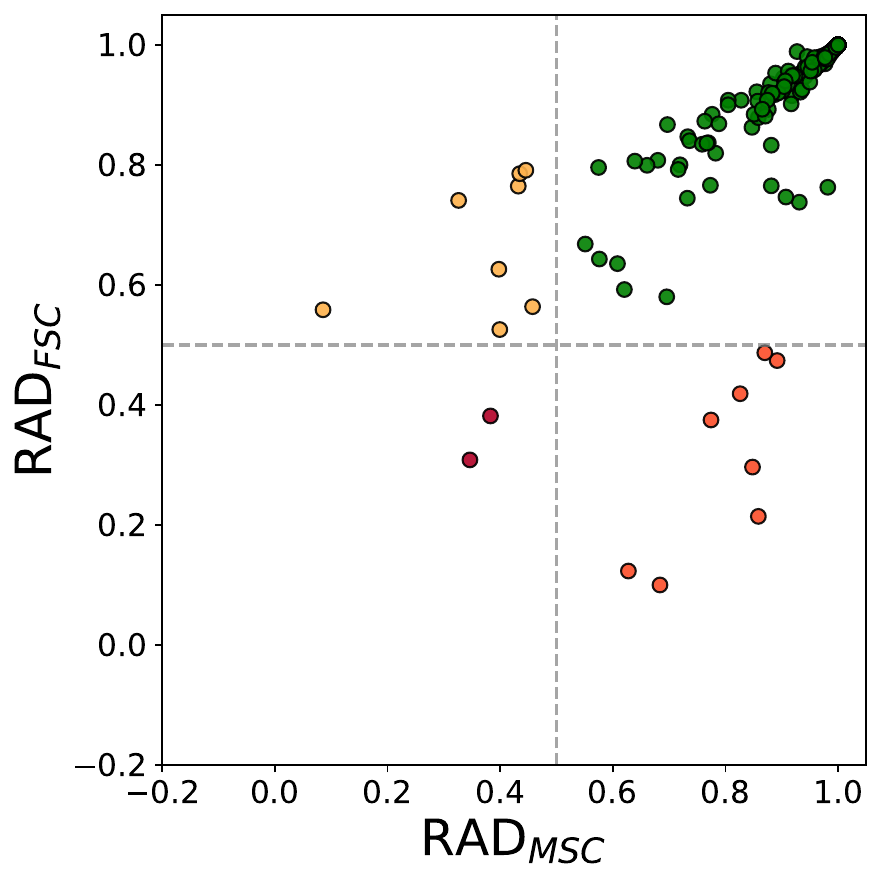}}

\caption{Spirals Dataset. Top Row: Decision boundary plots in 2-dimensions, Bottom Row: RAD Plot in 2-dimensions ($RAD_{MSC}$ and $RAD_{FSC}$).
}
\label{fig:spirals}
\end{figure}

\begin{figure}[htbp]
\centering
\subfloat[Low overlap]
{\includegraphics[width=0.33\textwidth]{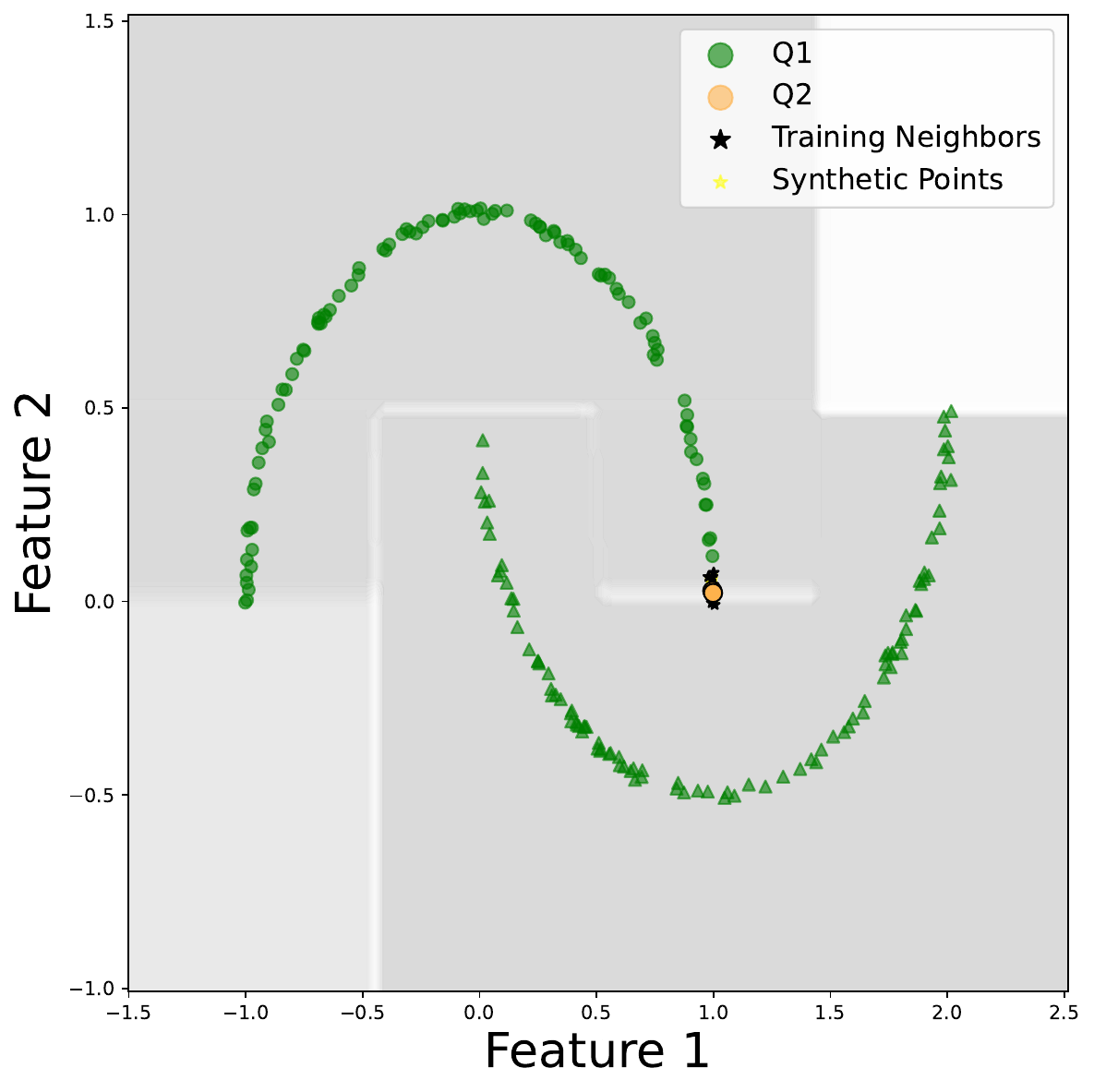}}
\hfill
\subfloat[Medium overlap]
{\includegraphics[width=0.33\textwidth]{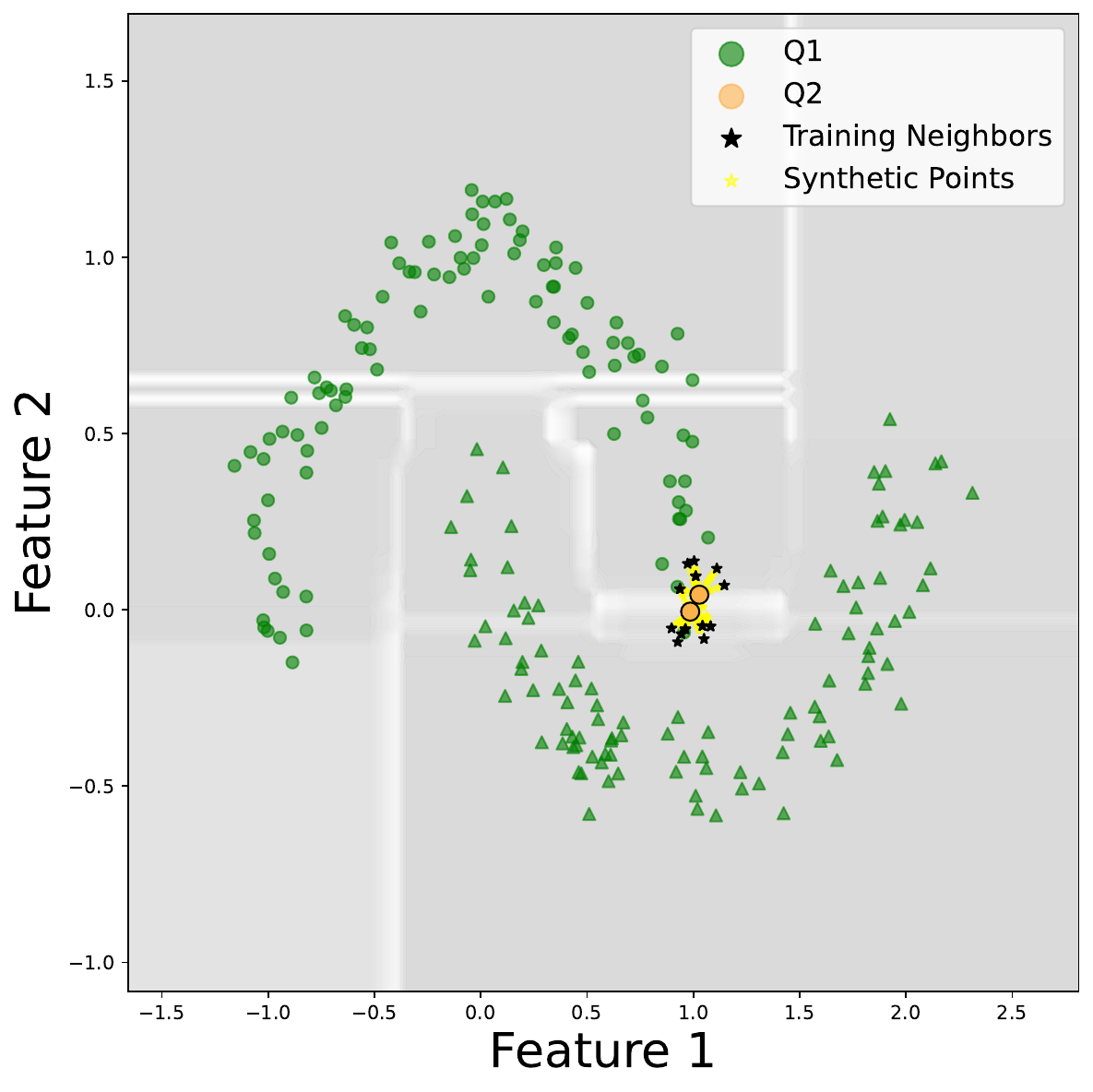}}
\hfill
\subfloat[High overlap]
{\includegraphics[width=0.33\textwidth]{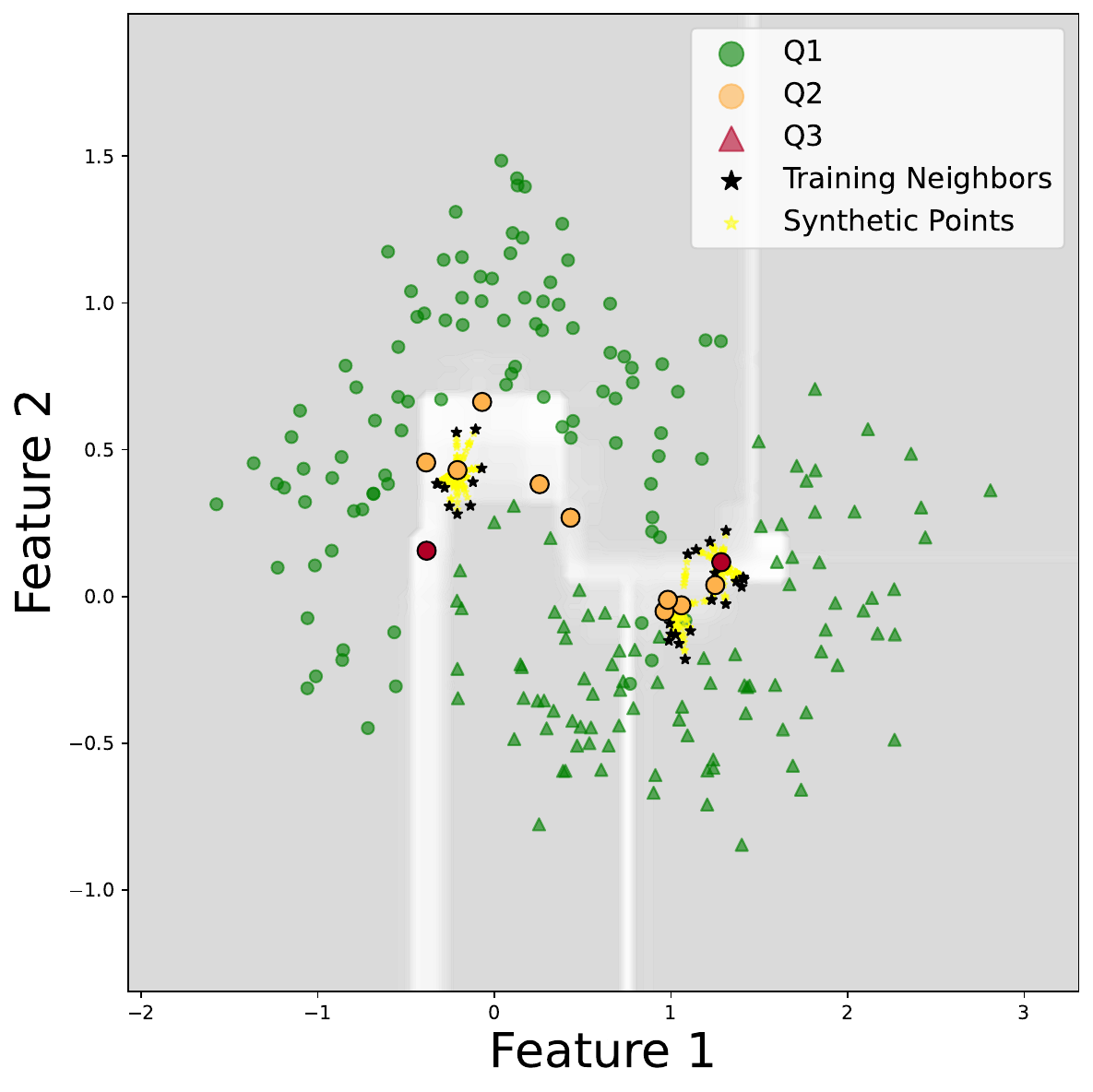}}

\subfloat[Low overlap RAD plot]{\includegraphics[width=0.33\textwidth]{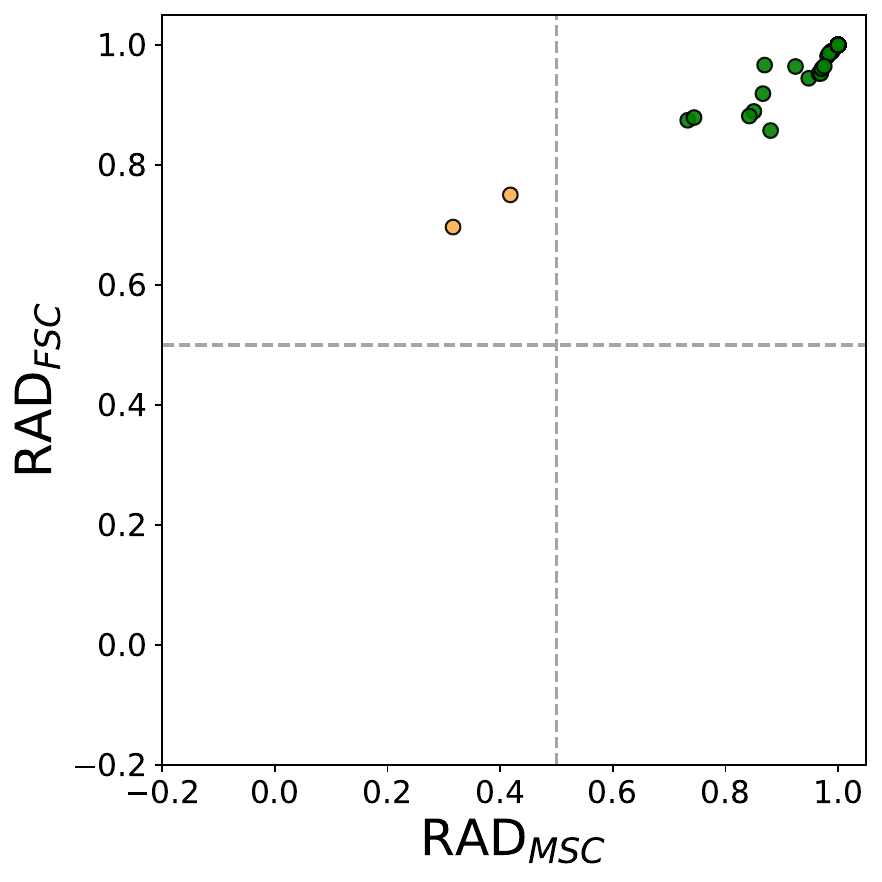}}
\hfill
\subfloat[Medium overlap RAD plot]{\includegraphics[width=0.33\textwidth]{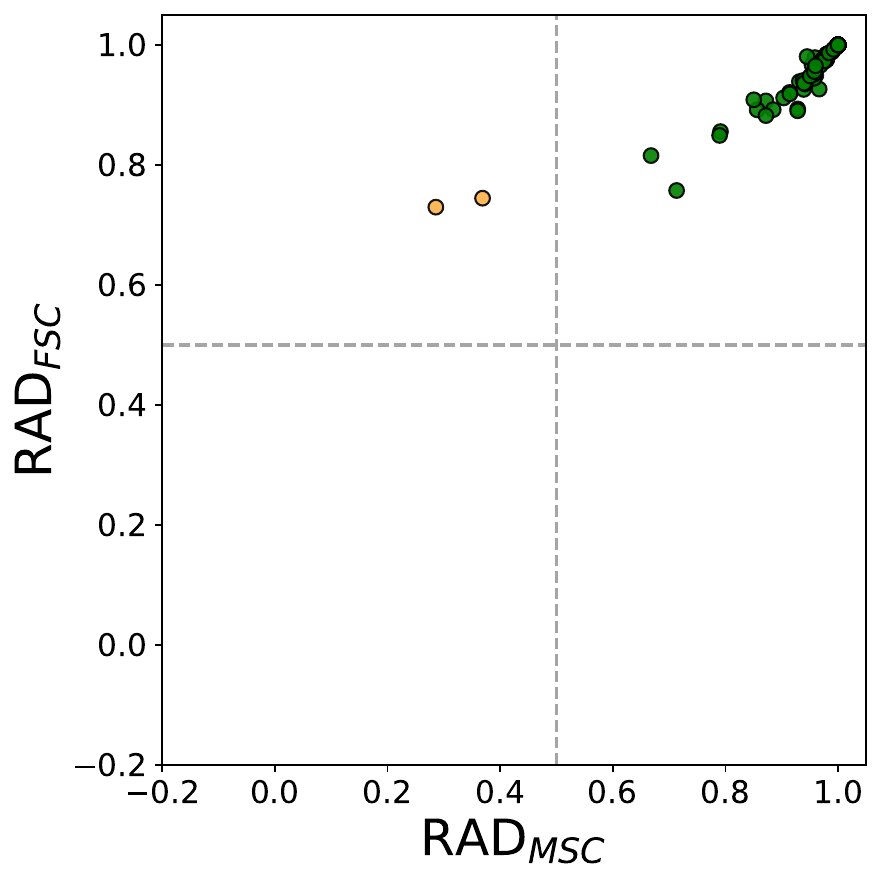}}
\hfill
\subfloat[High overlap RAD plot]{\includegraphics[width=0.33\textwidth]{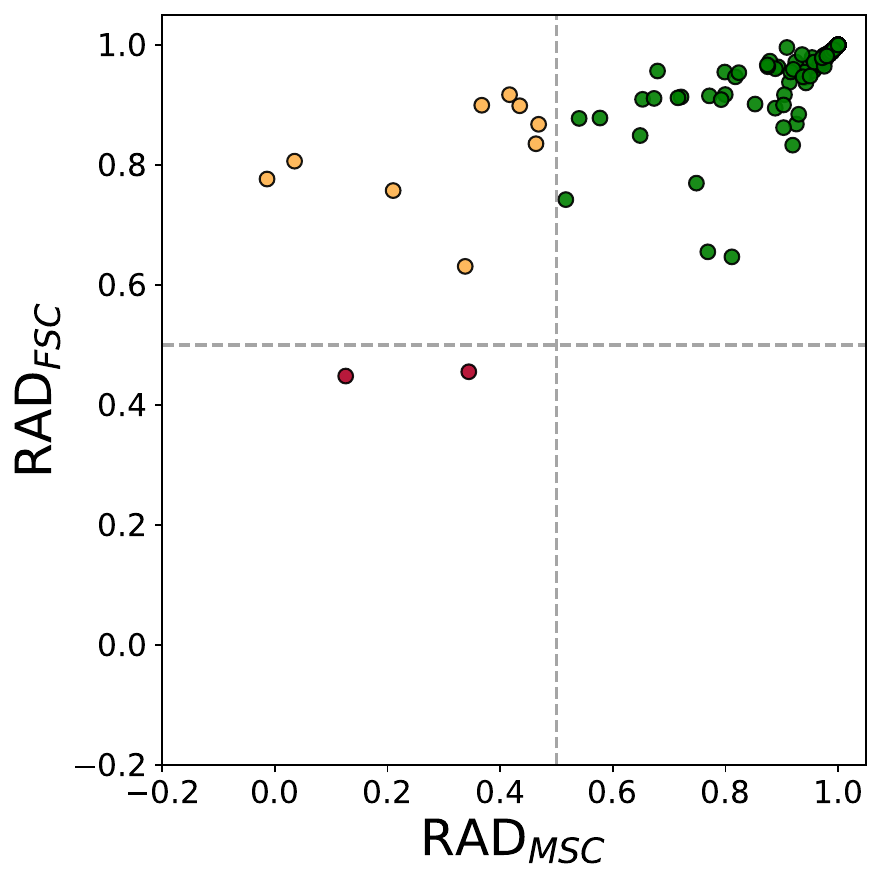}}

\caption{Moons Dataset. Top Row: Decision boundary plots in 2-dimensions, Bottom Row: RAD Plot.
}
\label{fig:moons}
\end{figure}


\begin{figure}[htbp]
\subfloat[\textit{Rice}]{\includegraphics[width=0.25\textwidth]{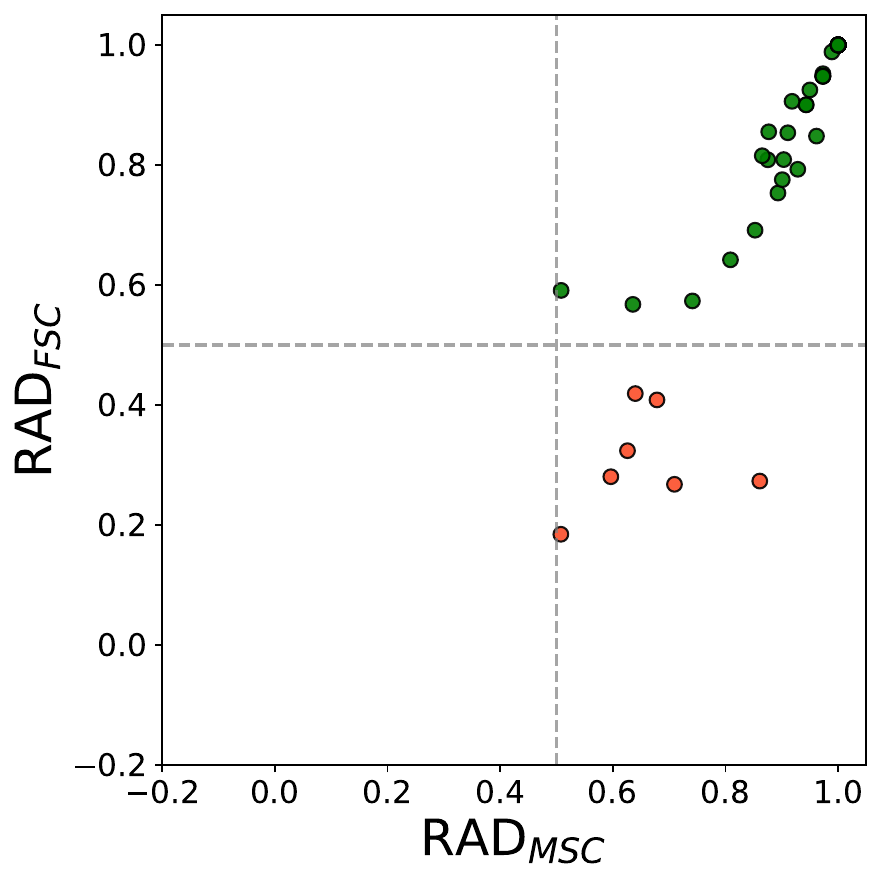}}
\subfloat[\textit{Default of Credit Card Clients}]{\includegraphics[width=0.25\textwidth]{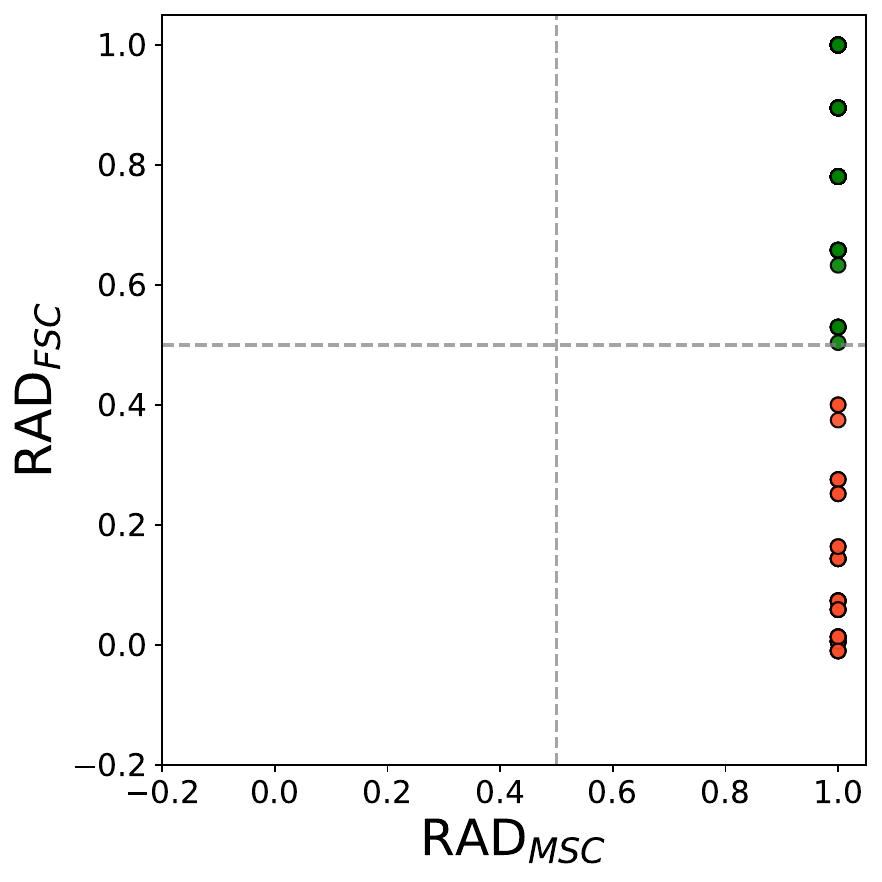}}
\subfloat[\textit{Mammographic Mass}]{\includegraphics[width=0.25\textwidth]{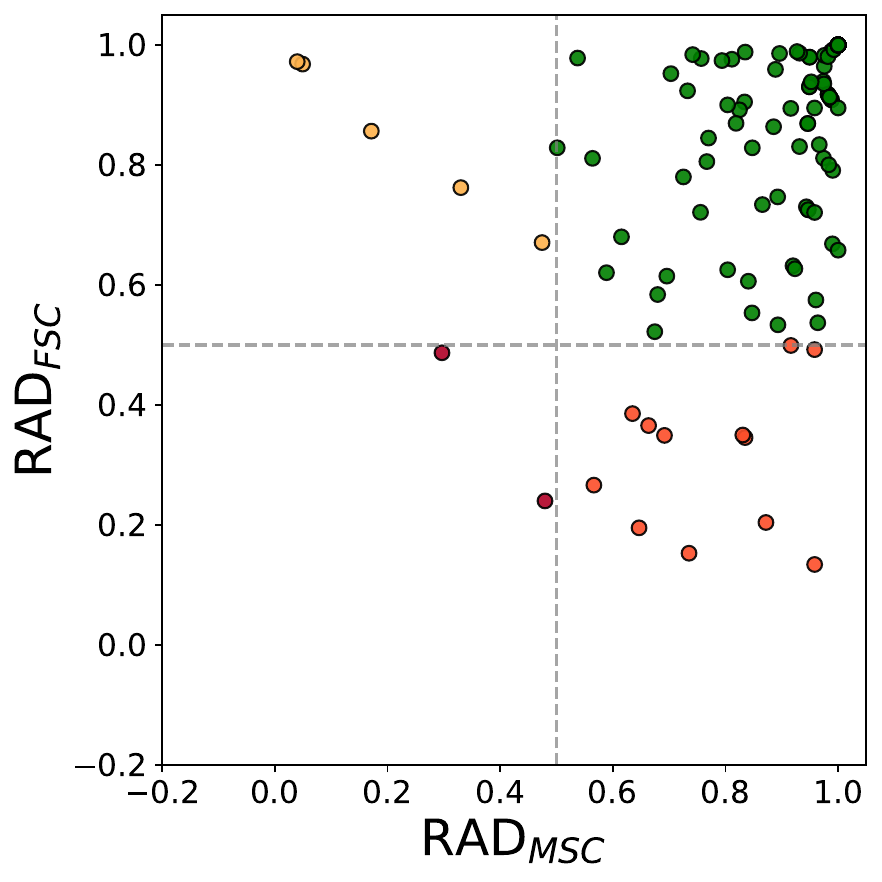}}
\subfloat[\textit{Breast Cancer Wisconsin}]
{\includegraphics[width=0.25\textwidth]{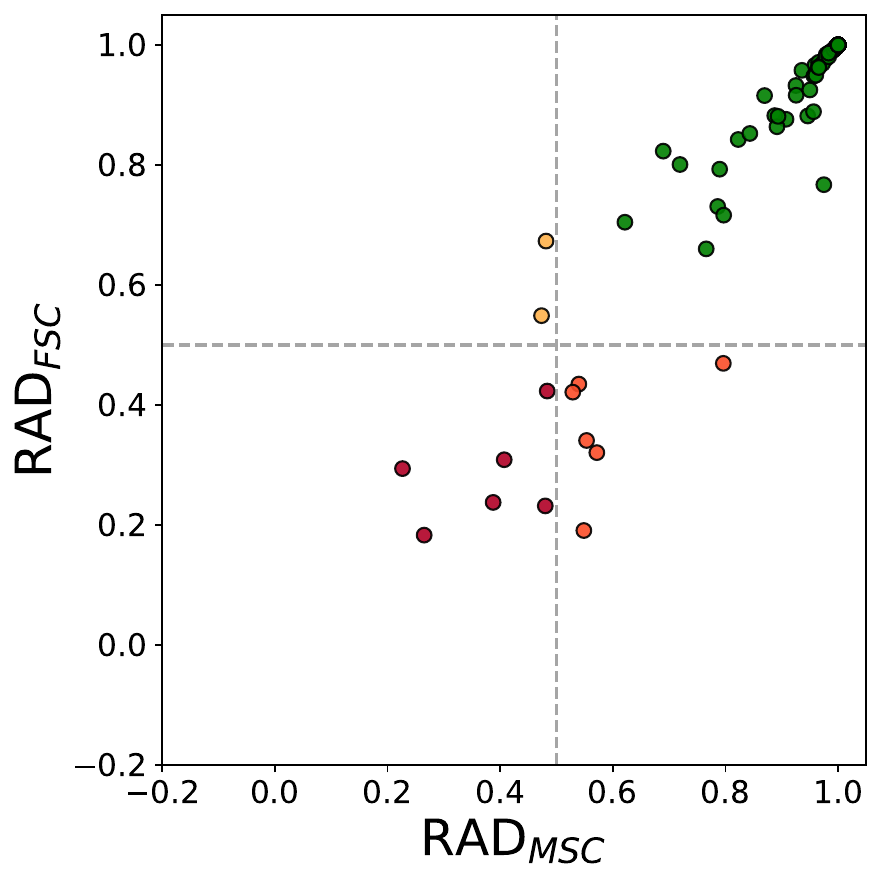}}

\caption{UCI Binary Datasets RAD Plots.
}
\label{fig:binary_uci_appendix}
\end{figure}

\begin{figure}[htbp]
\centering
\subfloat[\textit{Class 0}]{\includegraphics[width=0.17\textwidth]{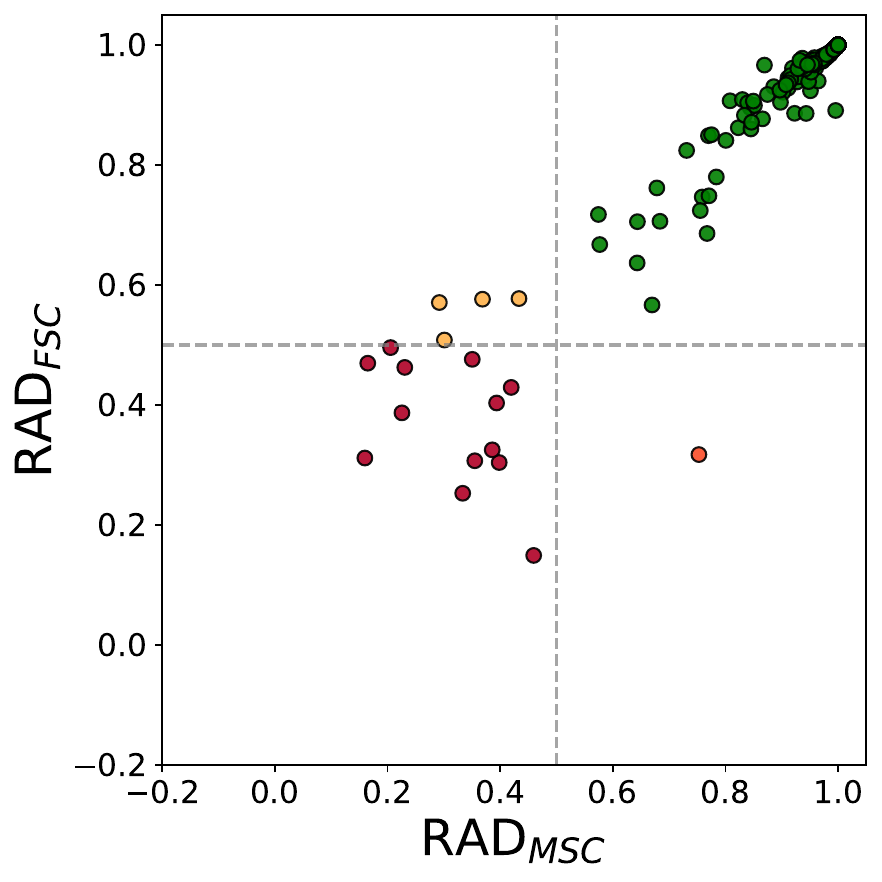}}
\subfloat[\textit{Class 1}]{\includegraphics[width=0.17\textwidth]{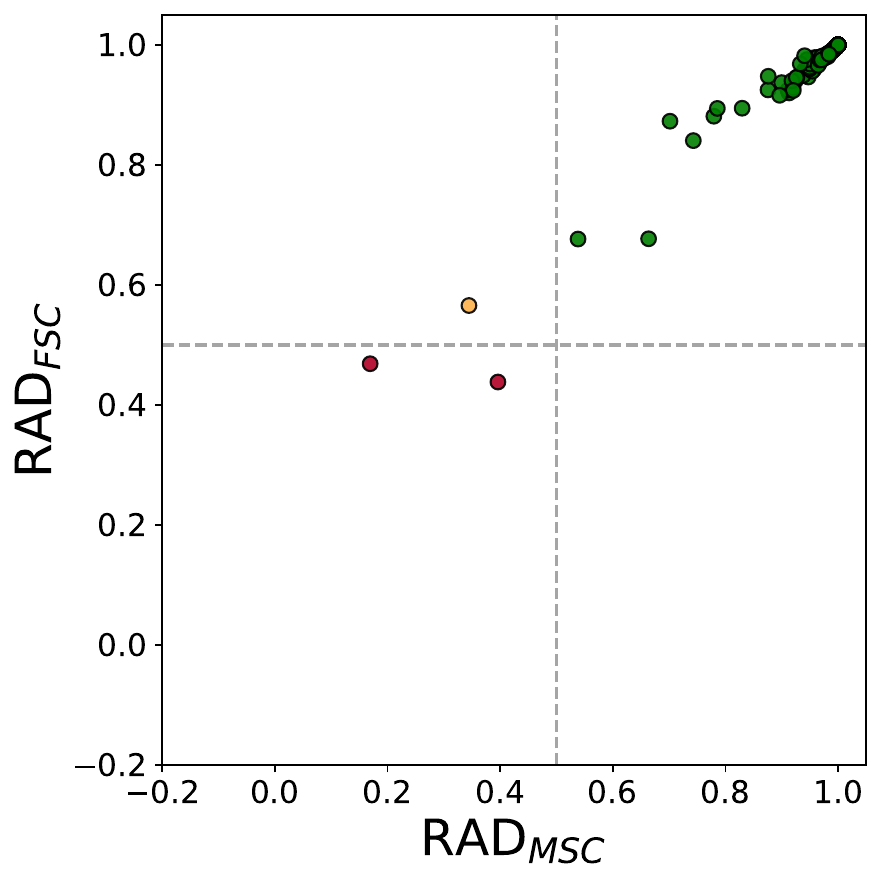}}
\subfloat[\textit{Class 2}]{\includegraphics[width=0.17\textwidth]{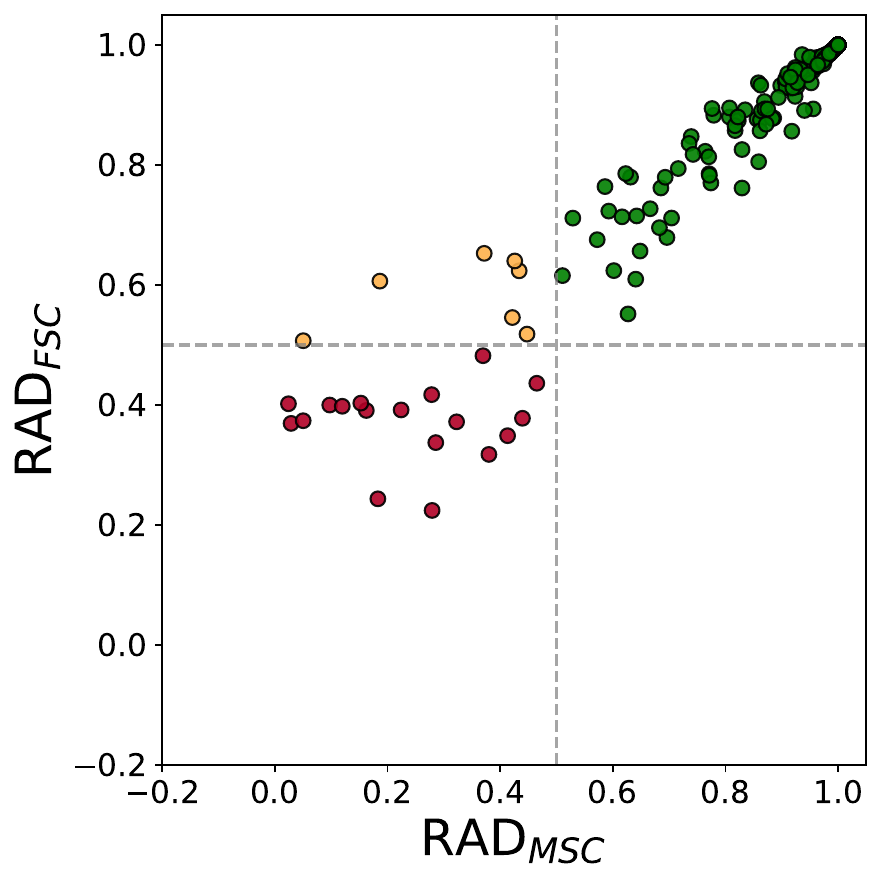}}
\subfloat[\textit{Class 3}]{\includegraphics[width=0.17\textwidth]{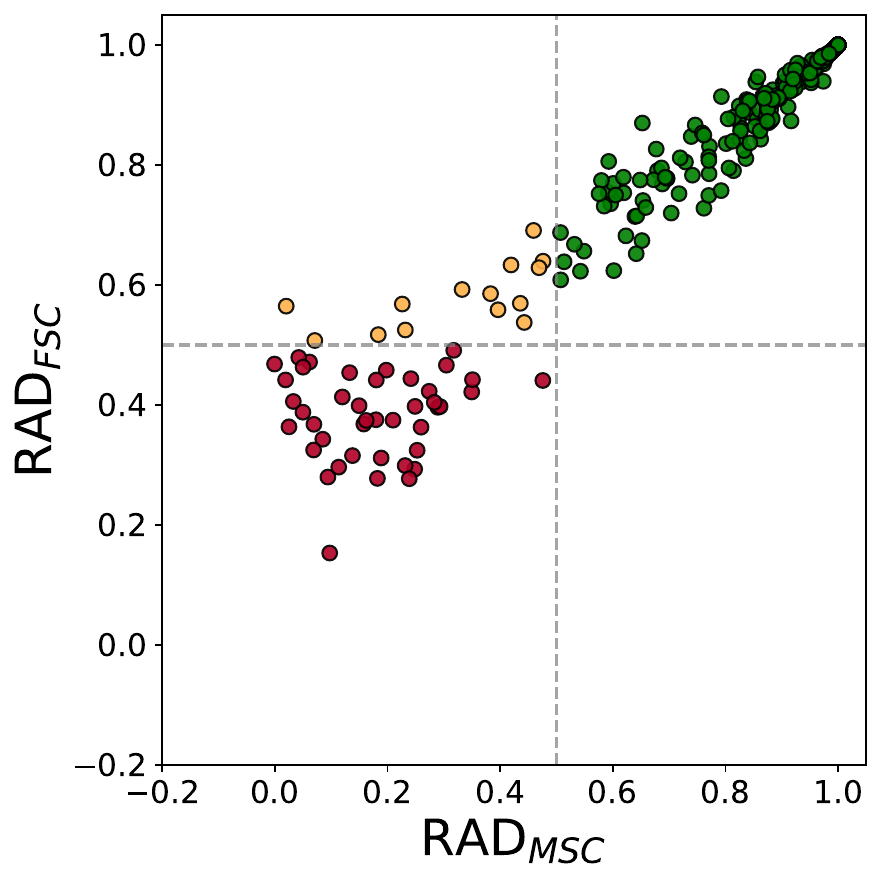}}
\subfloat[\textit{Class 4}]{\includegraphics[width=0.17\textwidth]{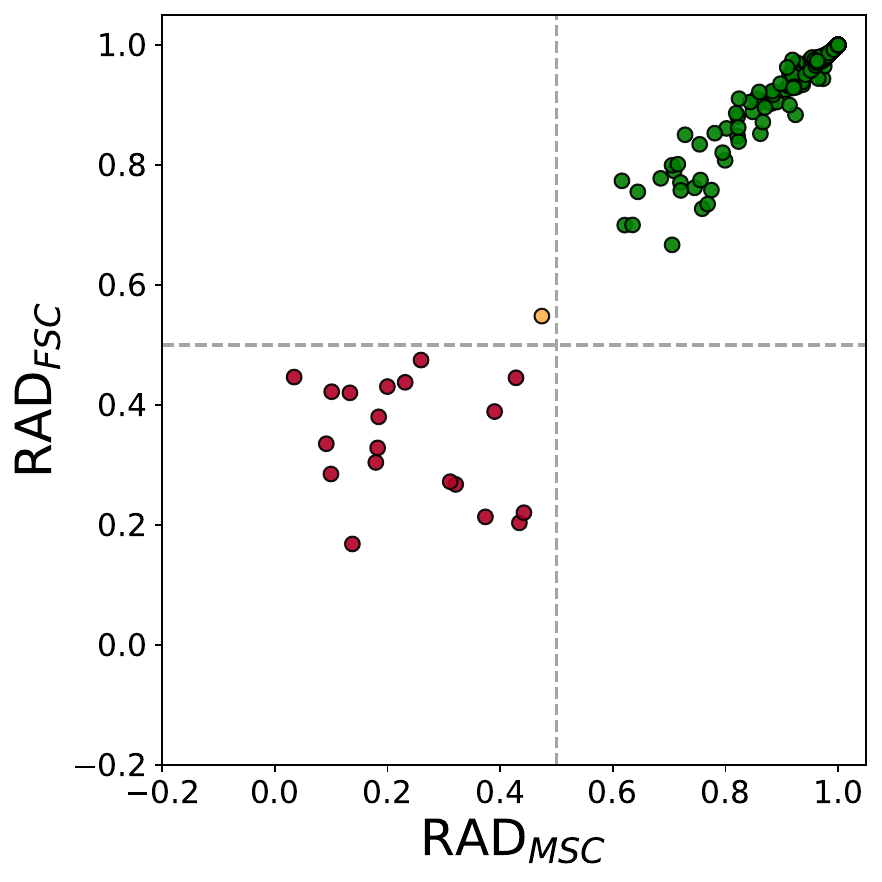}}
\subfloat[\textit{Class 5}]{\includegraphics[width=0.17\textwidth]{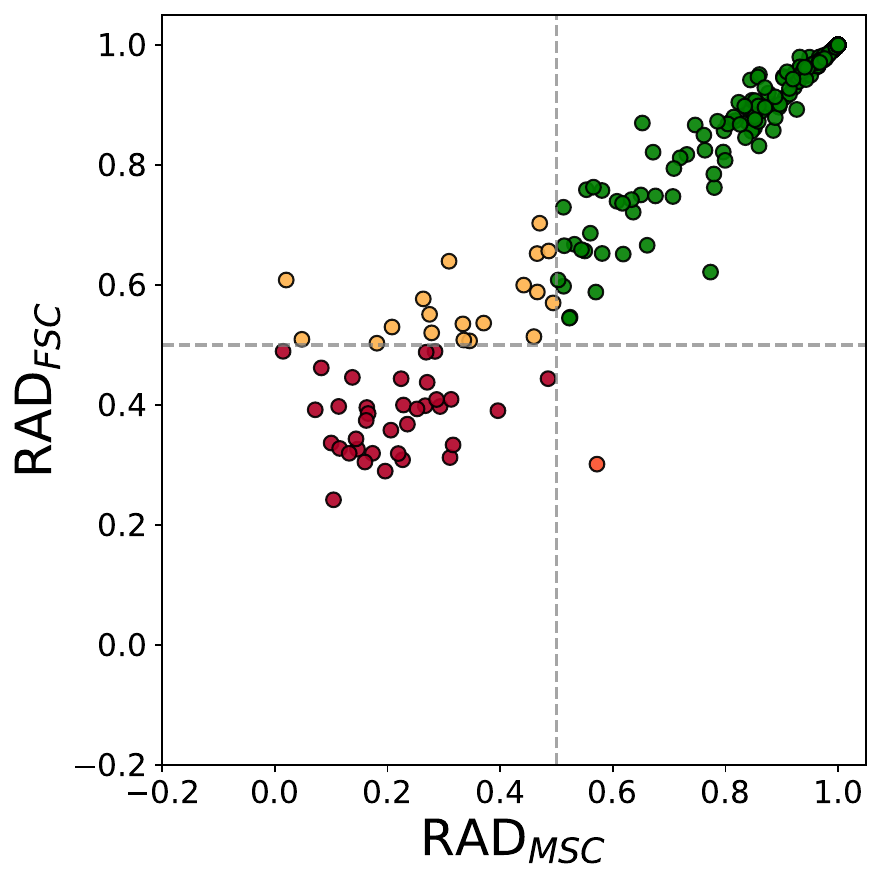}}

\caption{UCI Multi-class dataset \textit{Statlog Landsat} RAD Plots.}
\label{fig:statlog}
\end{figure}

\begin{figure}[htbp]
\centering
\subfloat[\textit{Class 0}]{\includegraphics[width=0.14\textwidth]{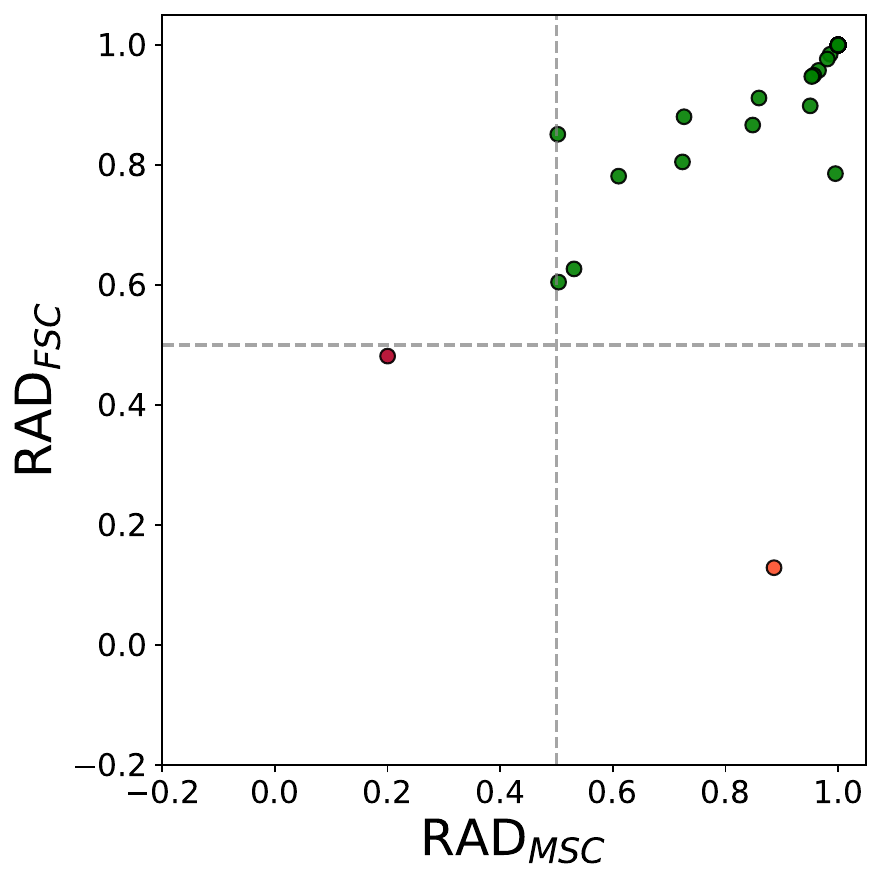}}
\subfloat[\textit{Class 1}]{\includegraphics[width=0.14\textwidth]{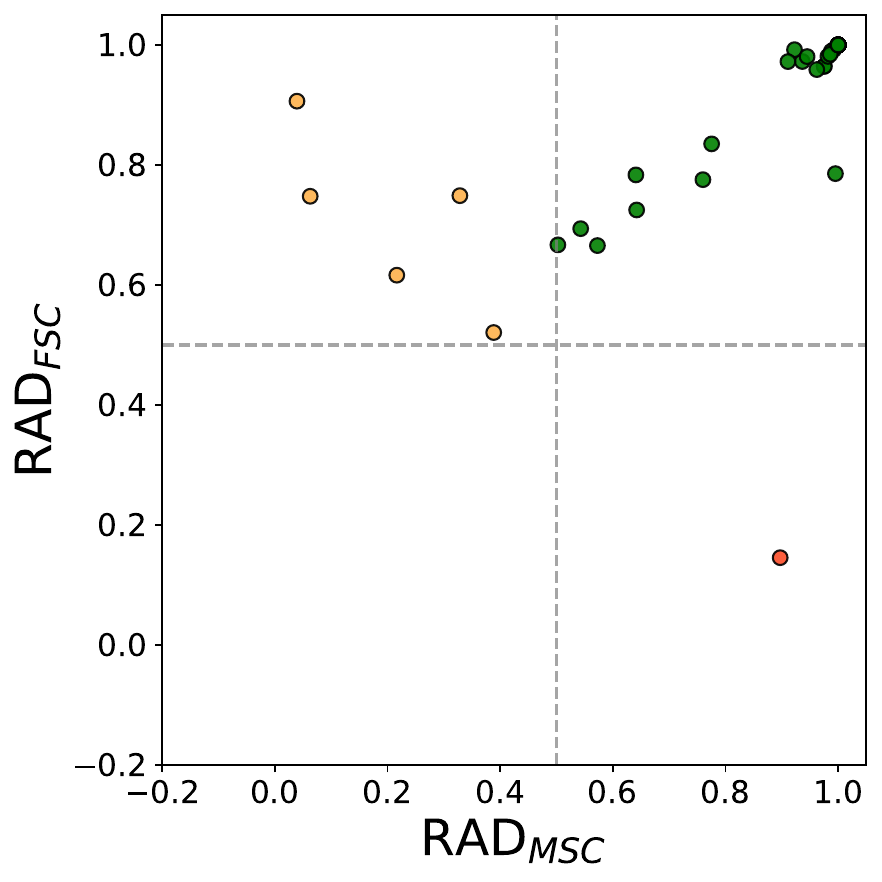}}
\subfloat[\textit{Class 2}]{\includegraphics[width=0.14\textwidth]{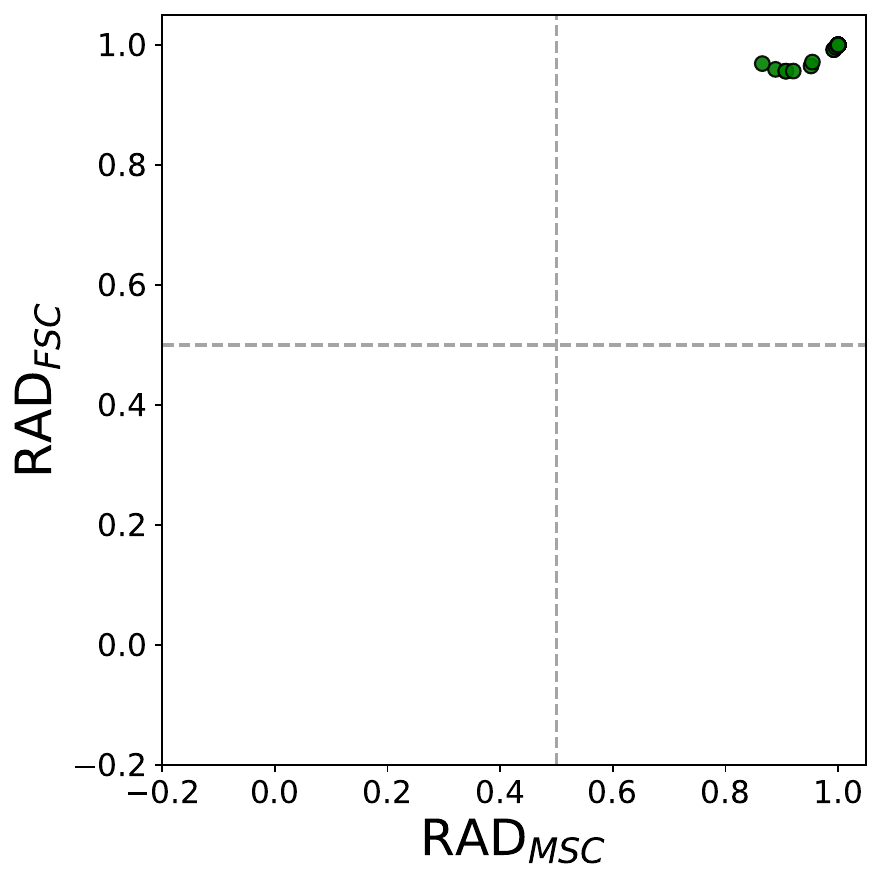}}
\subfloat[\textit{Class 3}]{\includegraphics[width=0.14\textwidth]{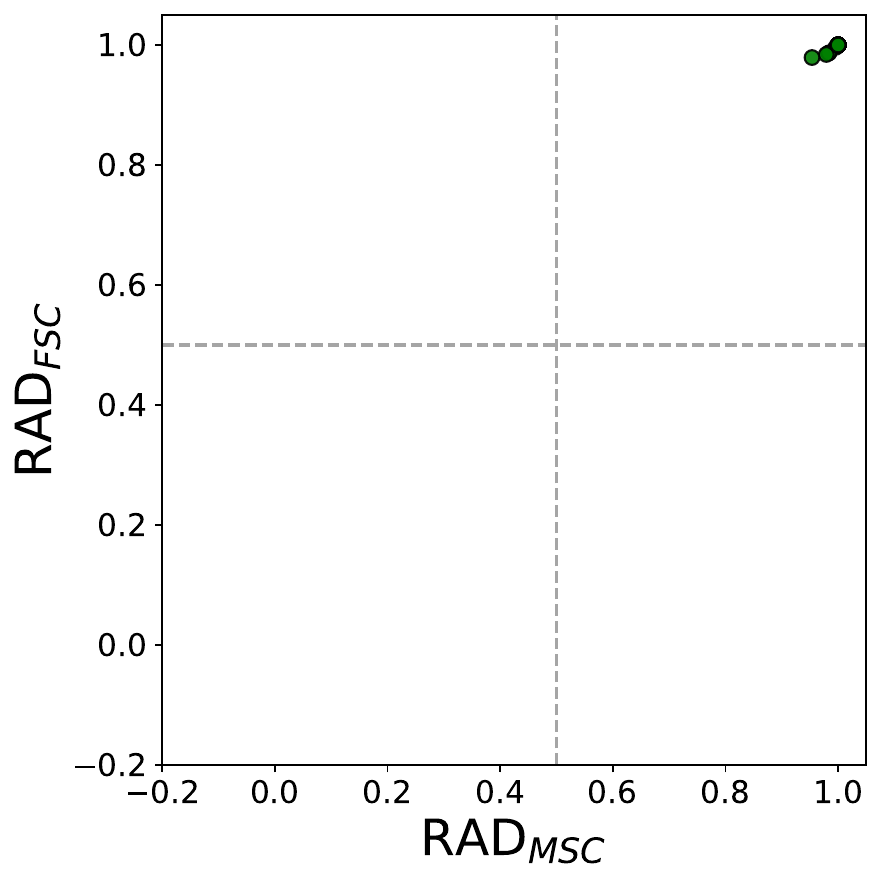}}
\subfloat[\textit{Class 4}]{\includegraphics[width=0.14\textwidth]{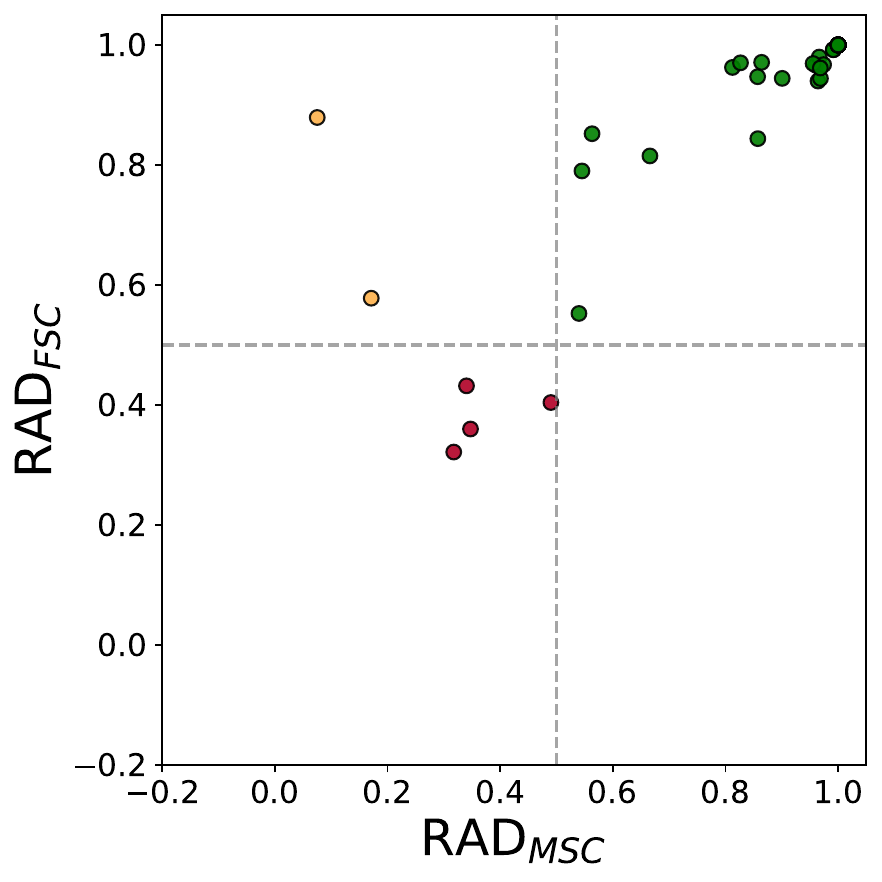}}
\subfloat[\textit{Class 5}]{\includegraphics[width=0.14\textwidth]{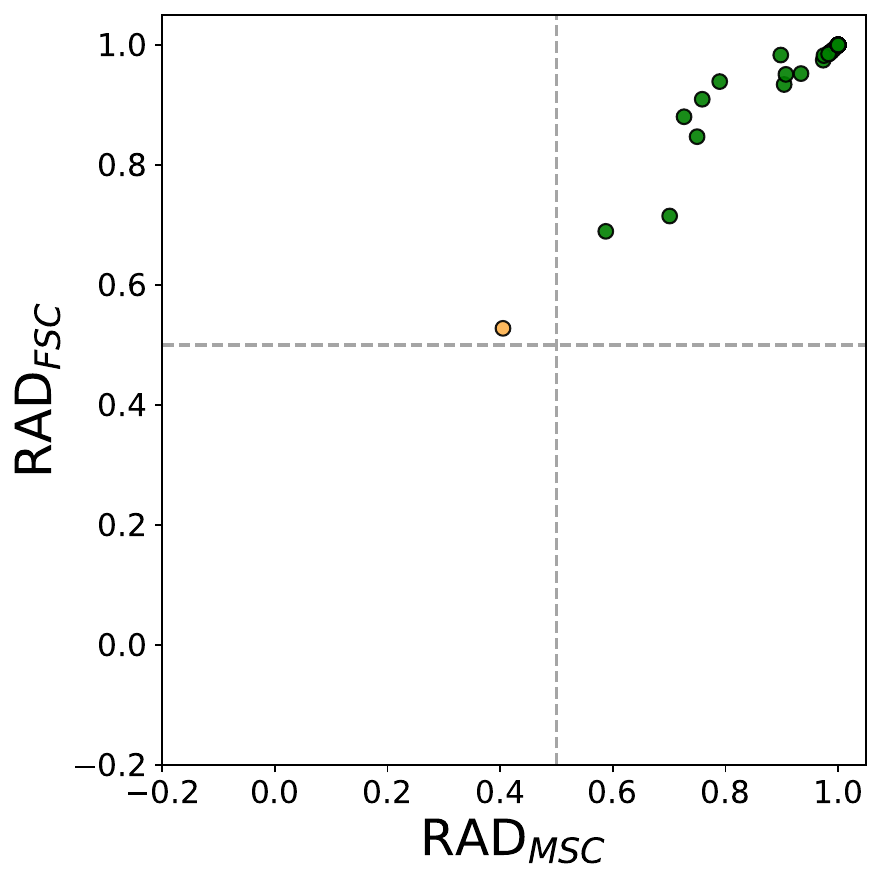}}
\subfloat[\textit{Class 6}]{\includegraphics[width=0.14\textwidth]{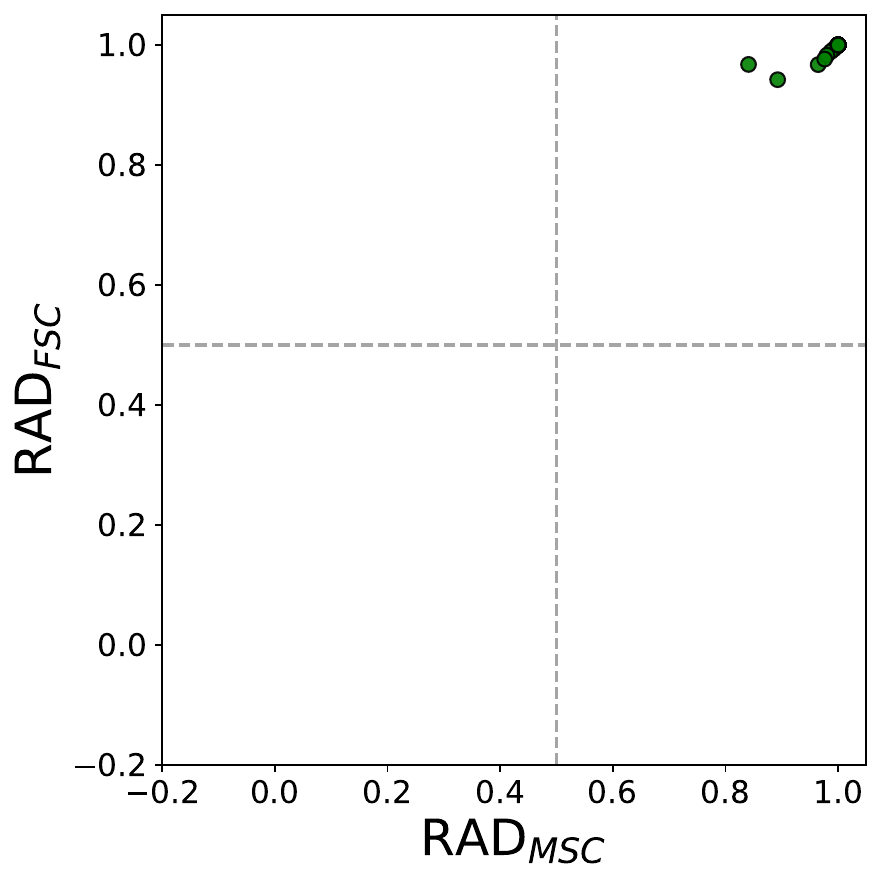}}

\caption{UCI Multi-class dataset \textit{E.coli} RAD Plots.
}
\label{fig:ecoli}
\end{figure}

\begin{figure}[htbp]
\centering
\subfloat[\textit{Class 0}]{\includegraphics[width=0.2\textwidth]{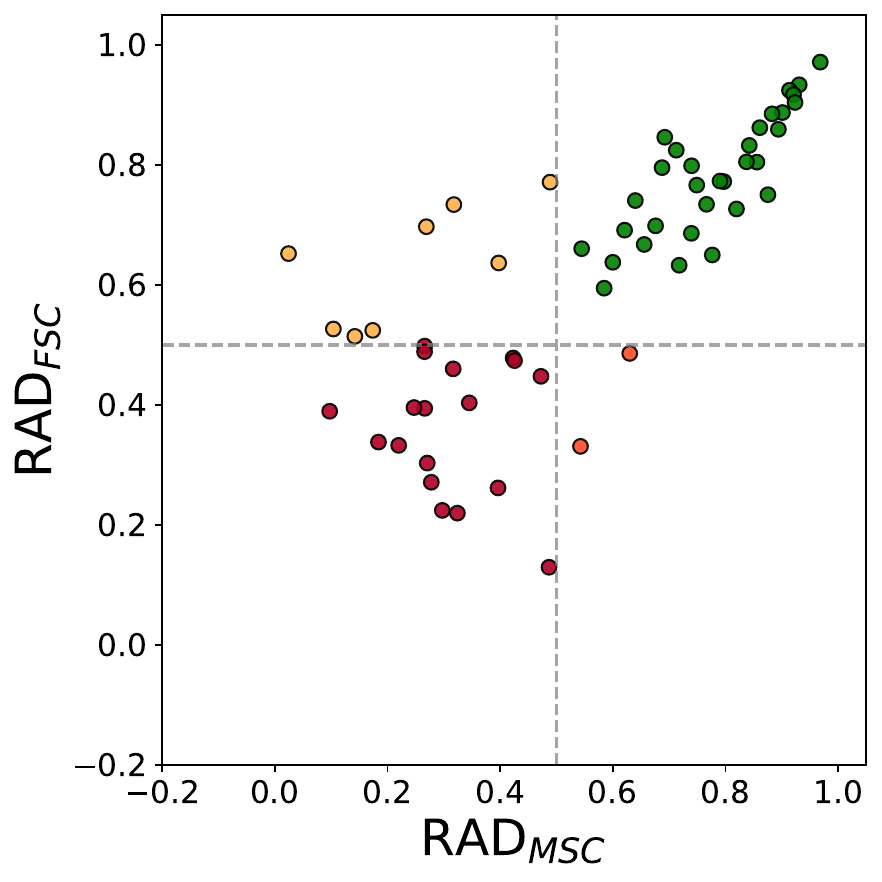}}
\subfloat[\textit{Class 1}]{\includegraphics[width=0.2\textwidth]{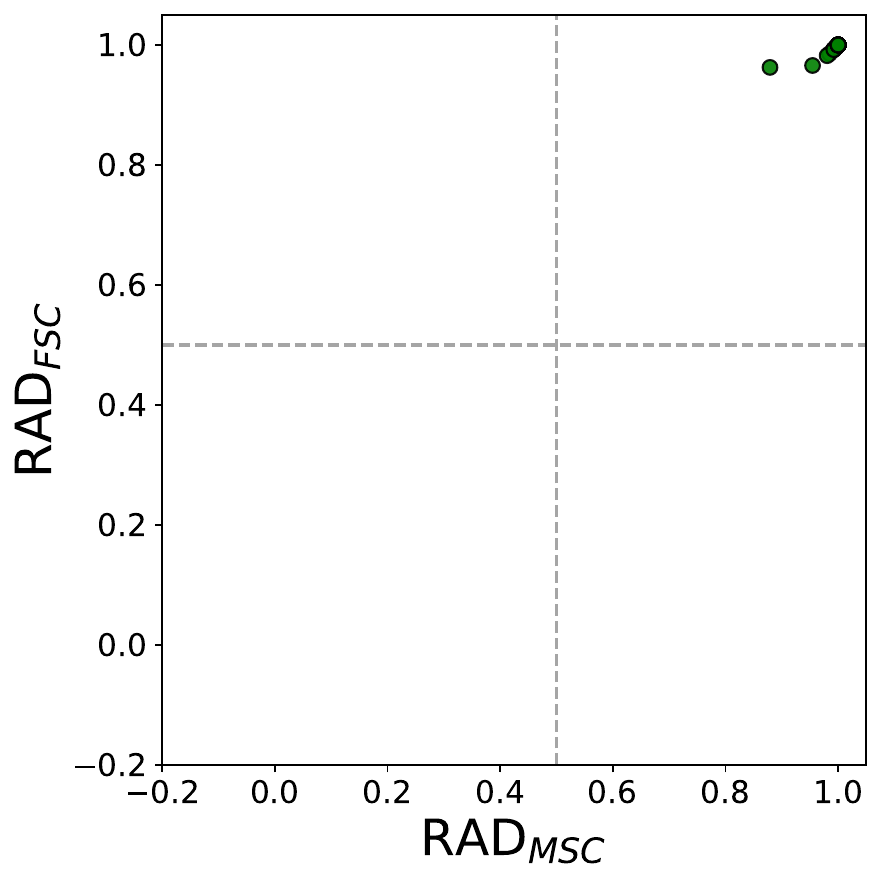}}
\subfloat[\textit{Class 2}]{\includegraphics[width=0.2\textwidth]{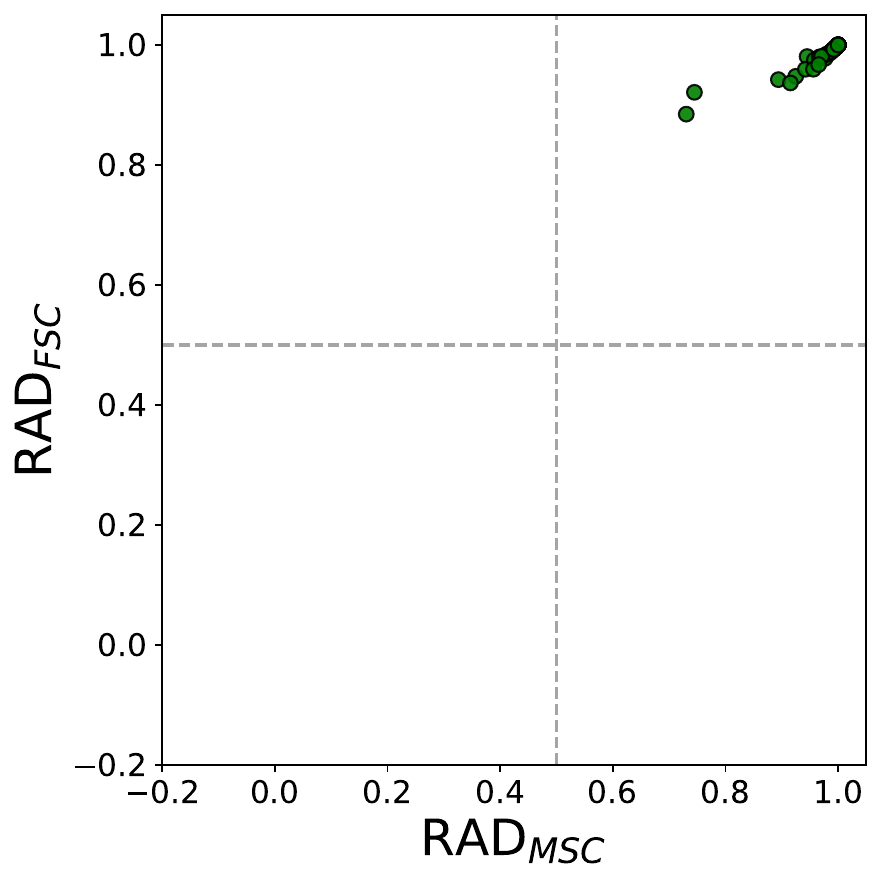}}
\subfloat[\textit{Class 3}]{\includegraphics[width=0.2\textwidth]{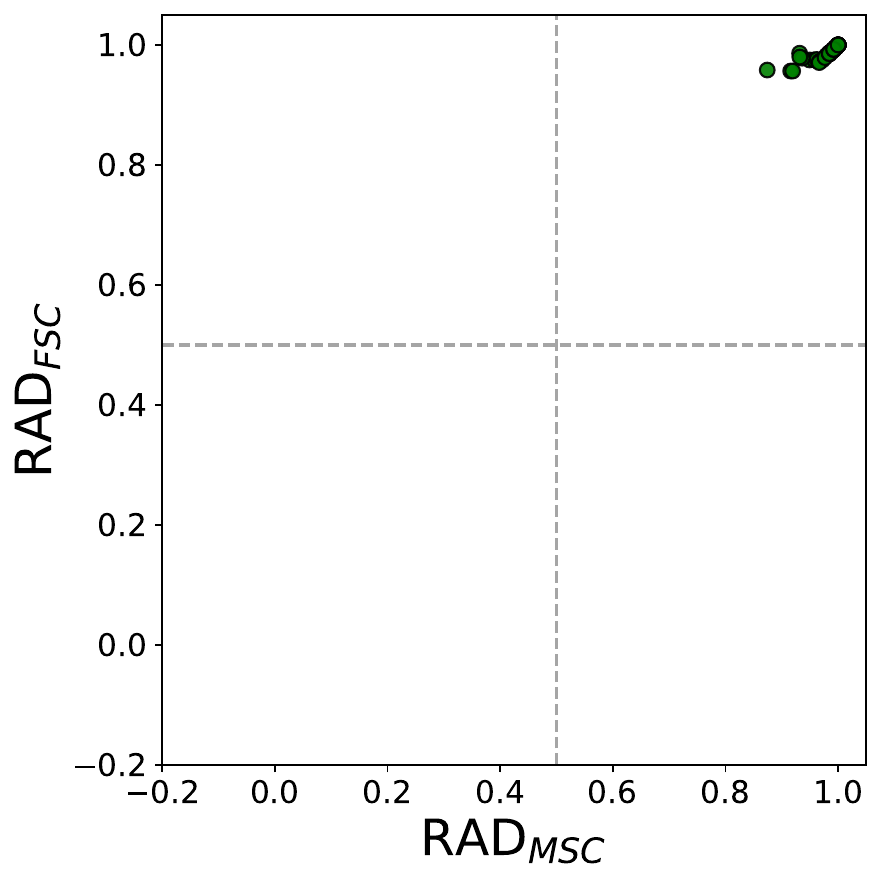}}
\subfloat[\textit{Class 4}]{\includegraphics[width=0.2\textwidth]{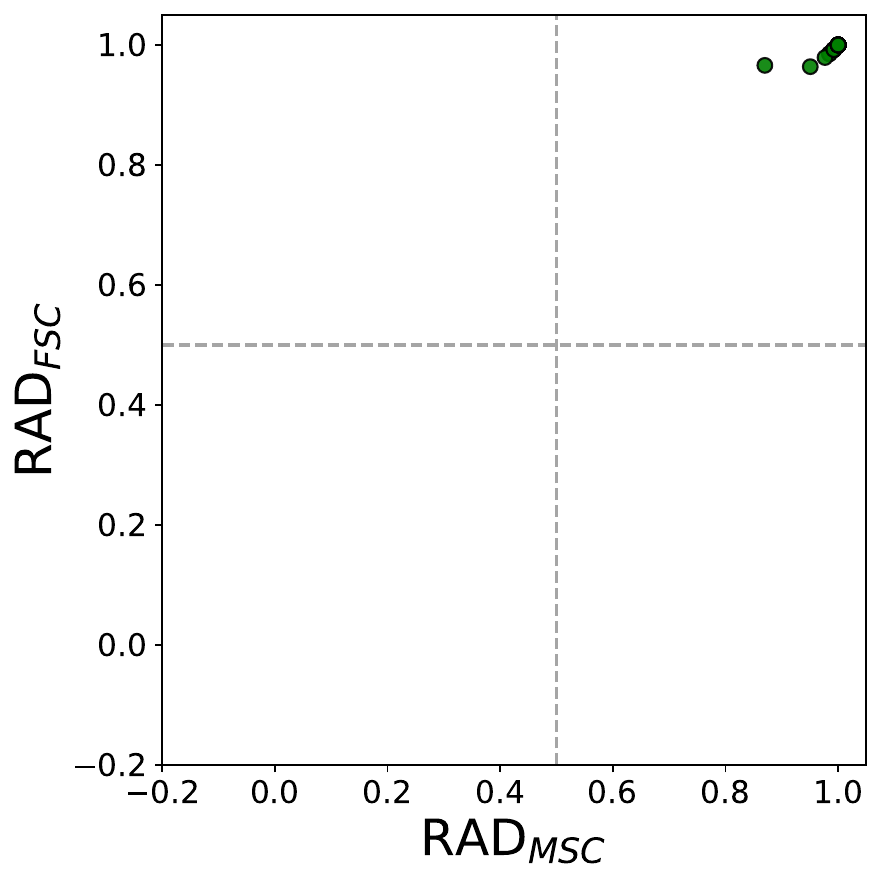}}

\caption{UCI Multi-class dataset \textit{Heart Disease} RAD Plots.
}
\label{fig:heart_disease}
\end{figure}

\begin{figure}[htbp]
\centering
\subfloat[\textit{Class 0}]{\includegraphics[width=0.17\textwidth]{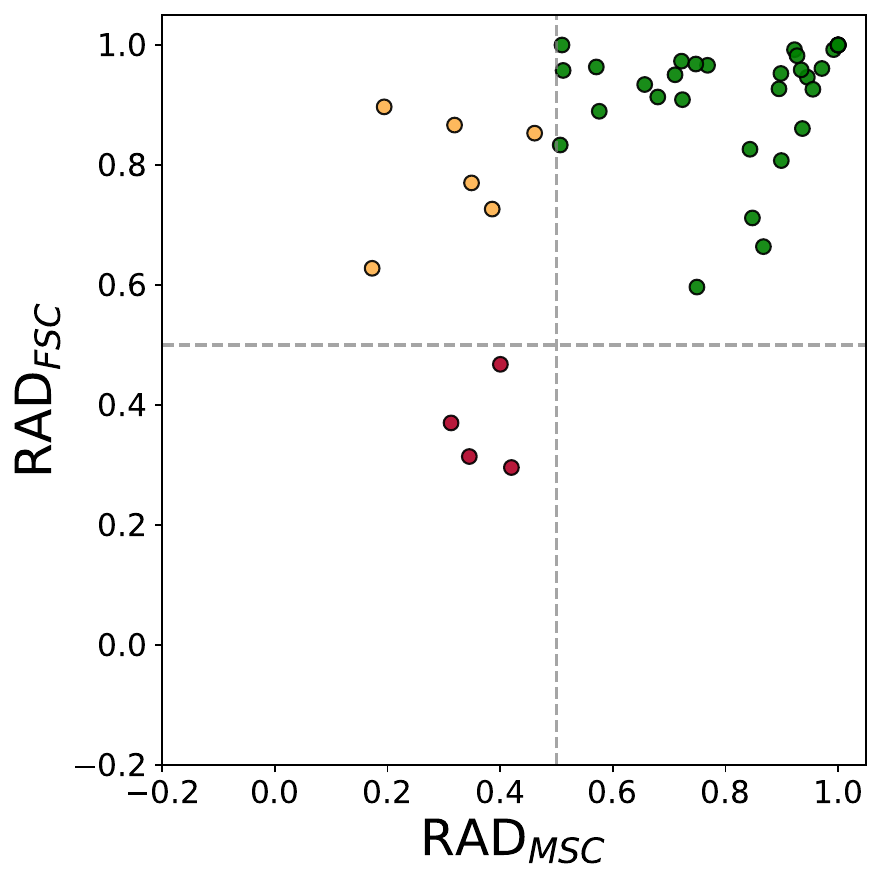}}
\subfloat[\textit{Class 1}]{\includegraphics[width=0.17\textwidth]{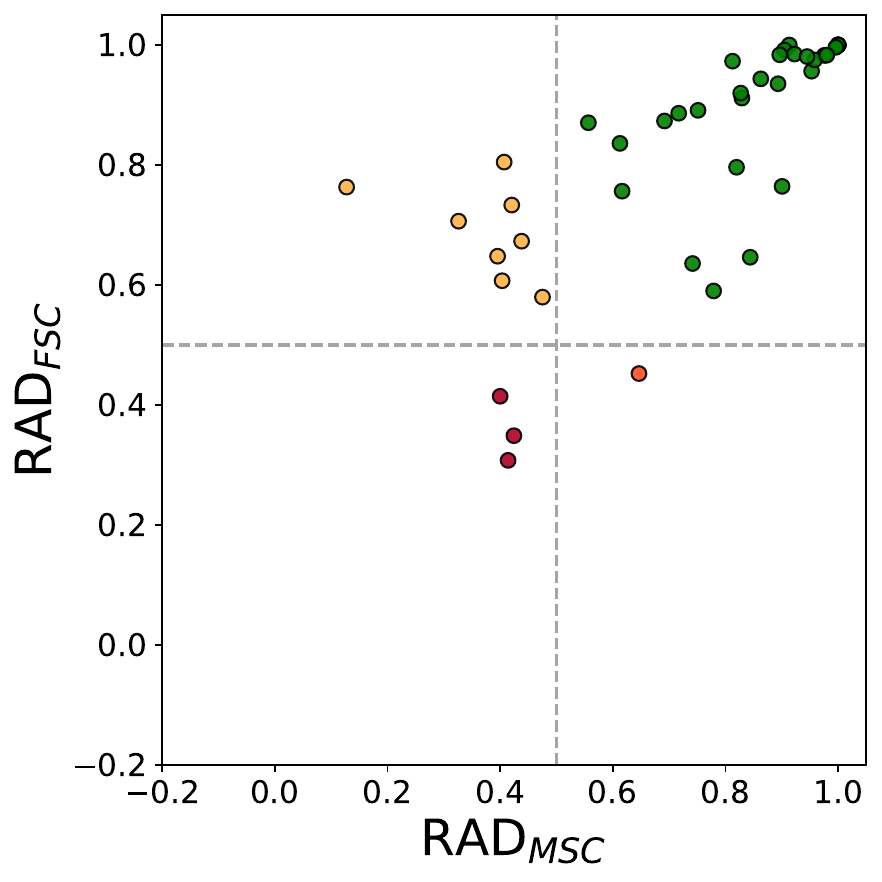}}
\subfloat[\textit{Class 2}]{\includegraphics[width=0.17\textwidth]{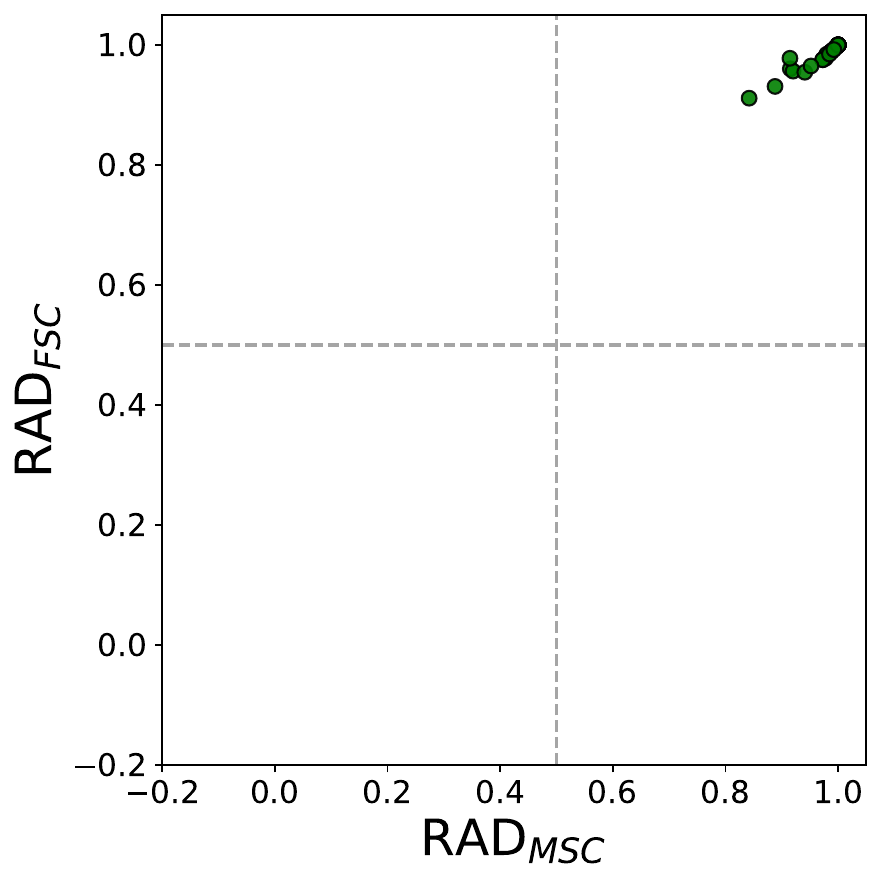}}
\subfloat[\textit{Class 3}]{\includegraphics[width=0.17\textwidth]{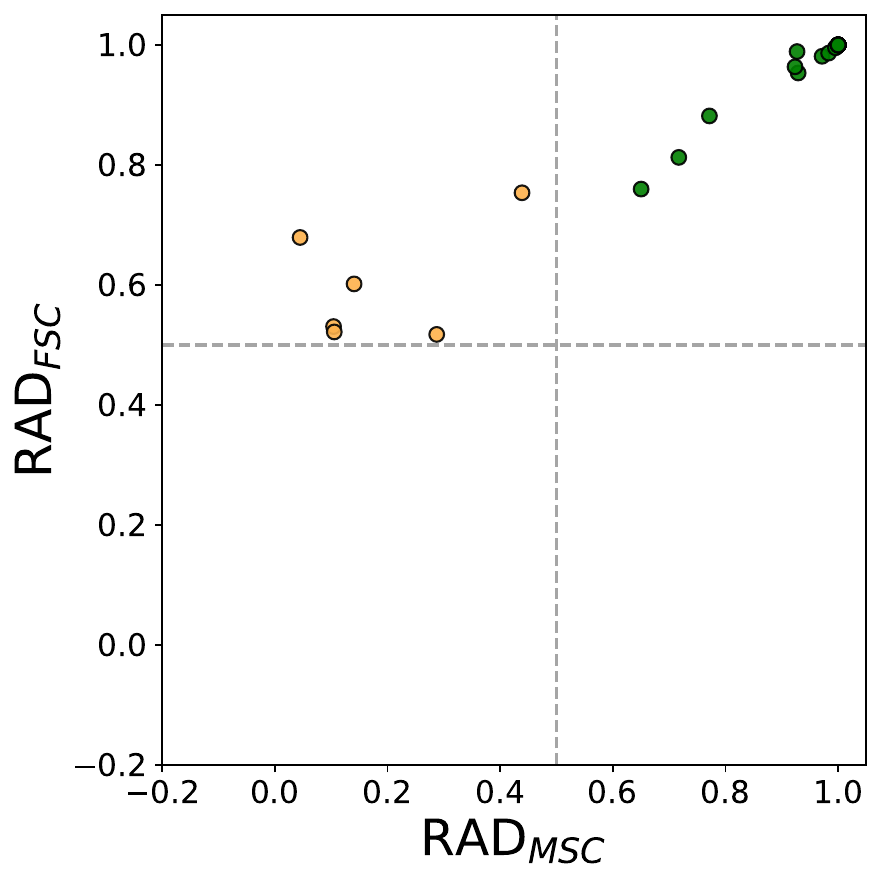}}
\subfloat[\textit{Class 4}]{\includegraphics[width=0.17\textwidth]{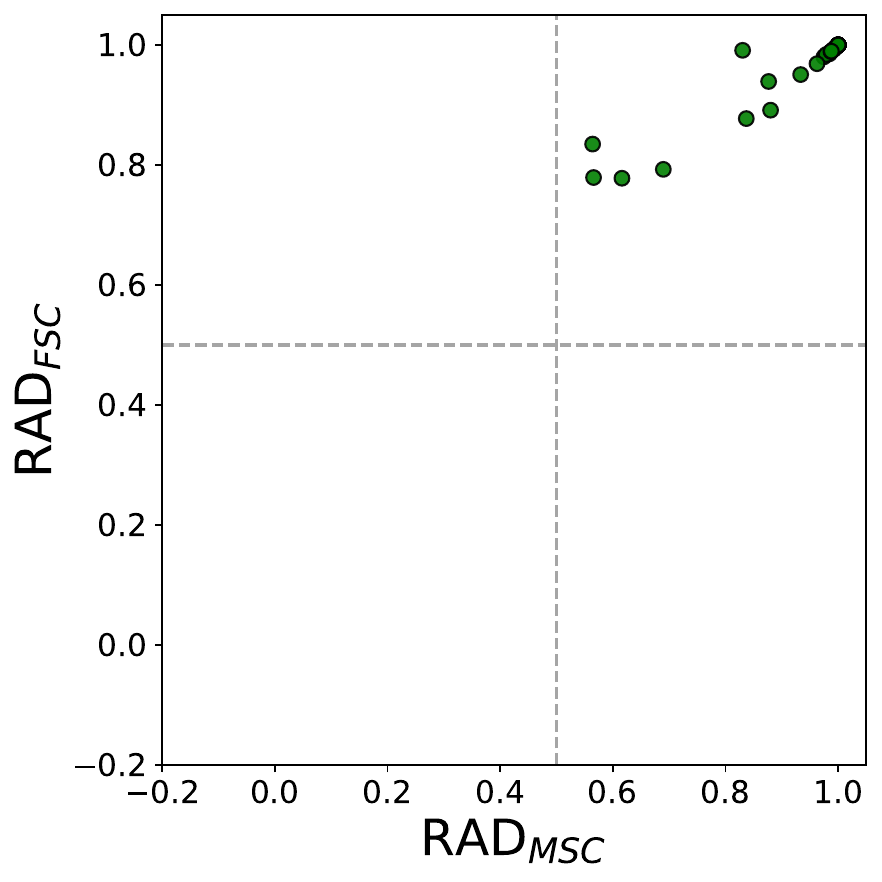}}
\subfloat[\textit{Class 5}]{\includegraphics[width=0.17\textwidth]{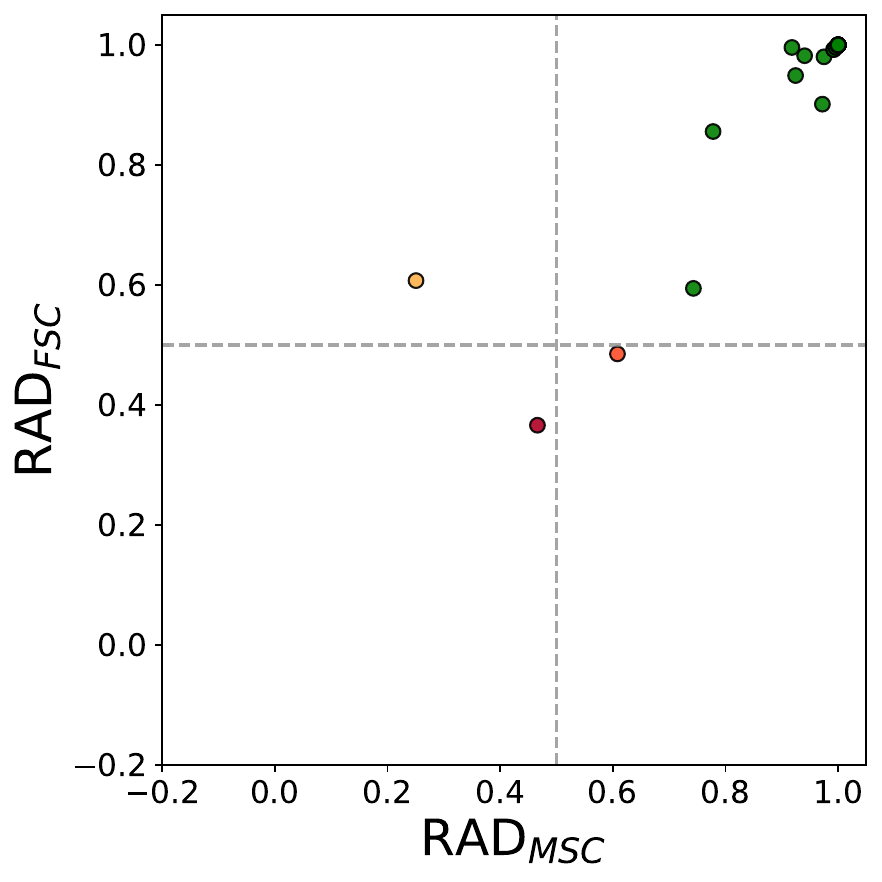}}

\caption{UCI Multi-class dataset \textit{Glass Identification} RAD Plots.
}
\label{fig:heart_disease}
\end{figure}

\begin{figure}[htbp]
\centering

\subfloat[\textit{Class 0}]{\includegraphics[width=0.2\textwidth]{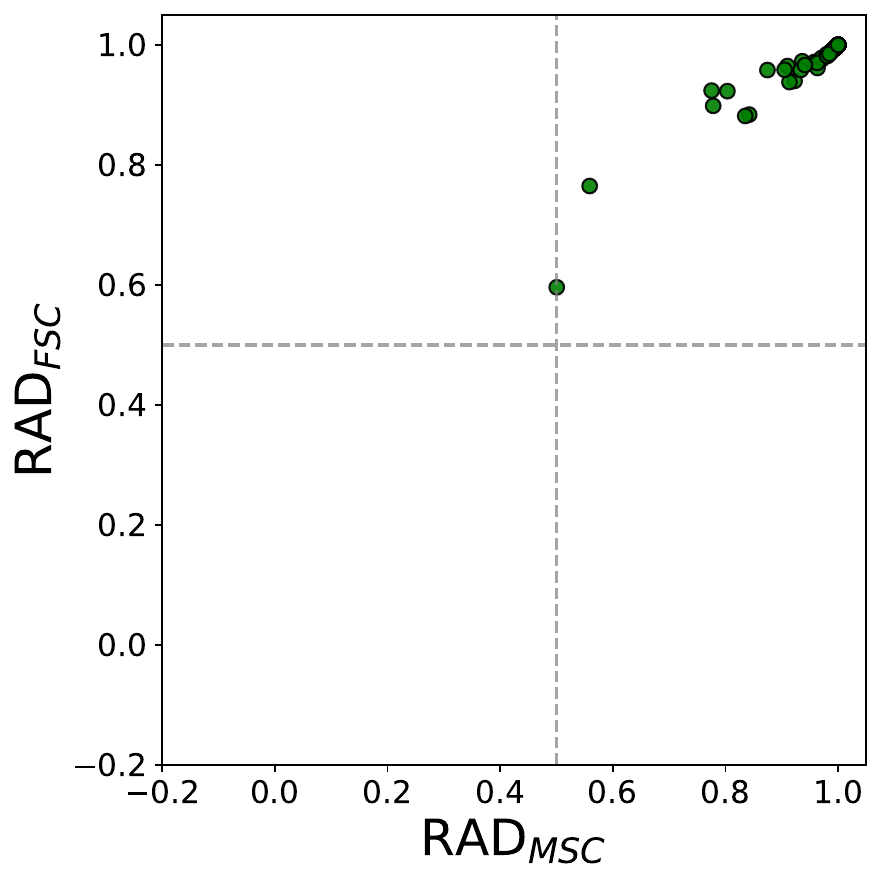}}
\subfloat[\textit{Class 1}]{\includegraphics[width=0.2\textwidth]{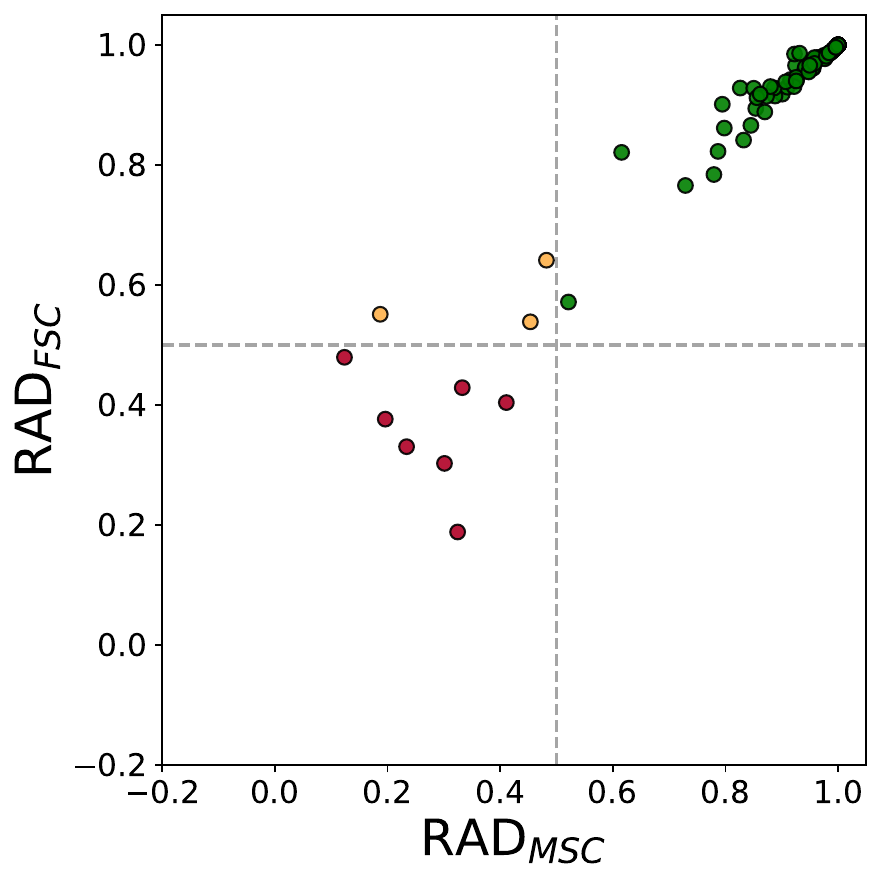}}
\subfloat[\textit{Class 2}]{\includegraphics[width=0.2\textwidth]{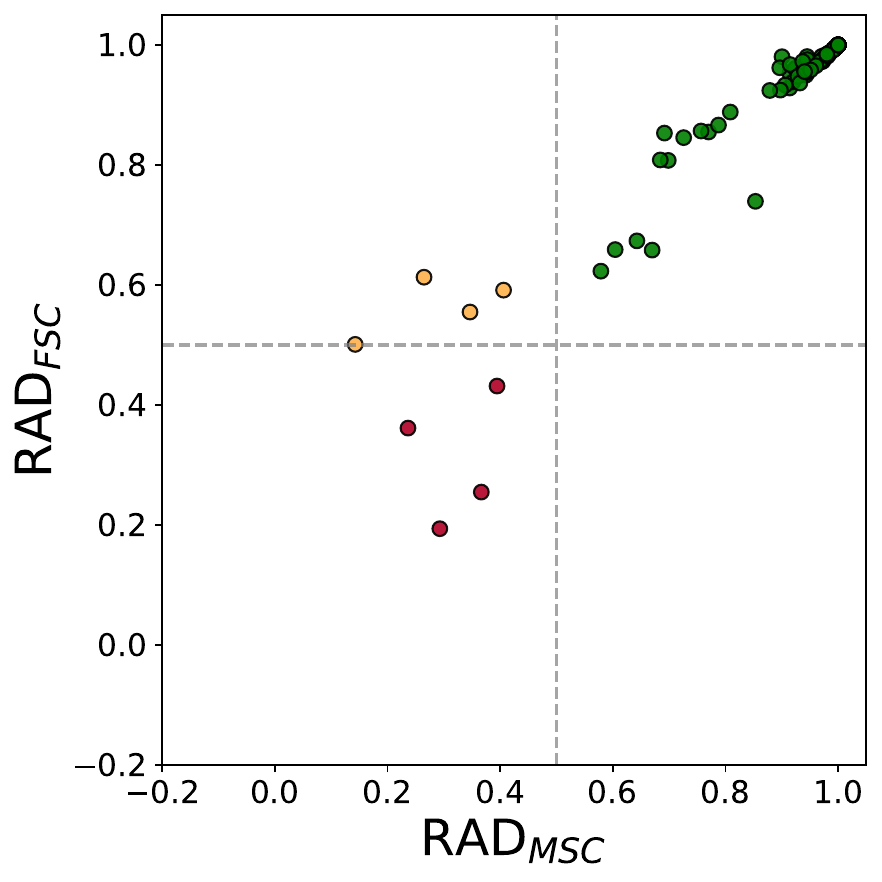}}
\subfloat[\textit{Class 3}]{\includegraphics[width=0.2\textwidth]{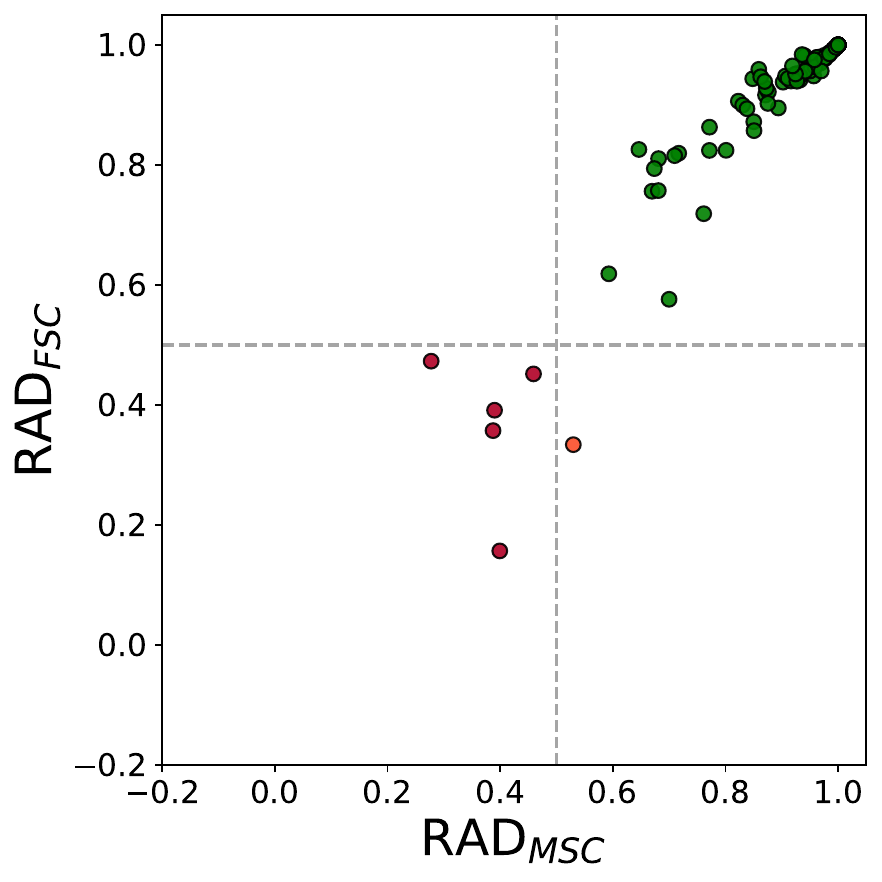}}
\subfloat[\textit{Class 4}]{\includegraphics[width=0.2\textwidth]{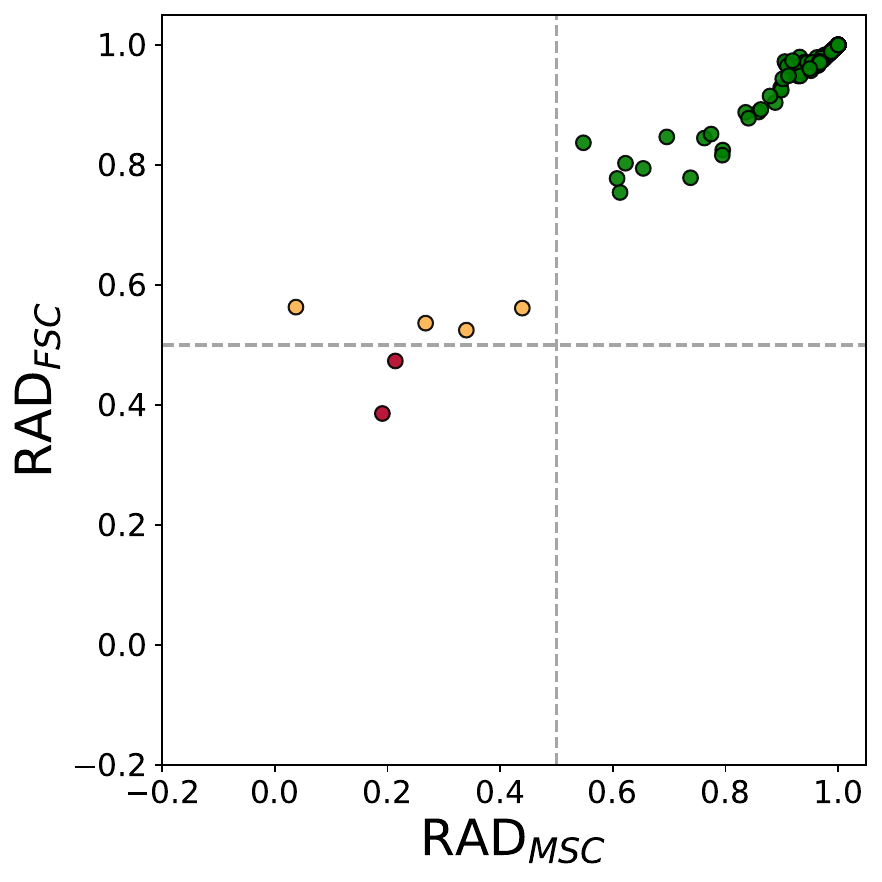}}

\subfloat[\textit{Class 5}]{\includegraphics[width=0.2\textwidth]{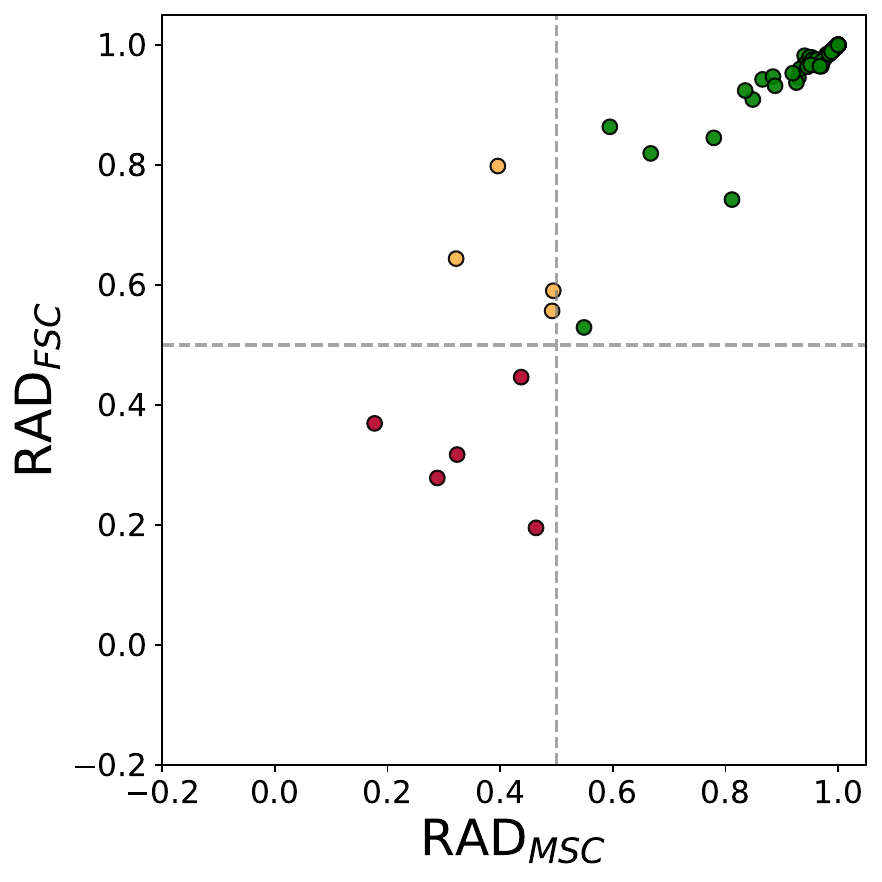}}
\subfloat[\textit{Class 6}]{\includegraphics[width=0.2\textwidth]{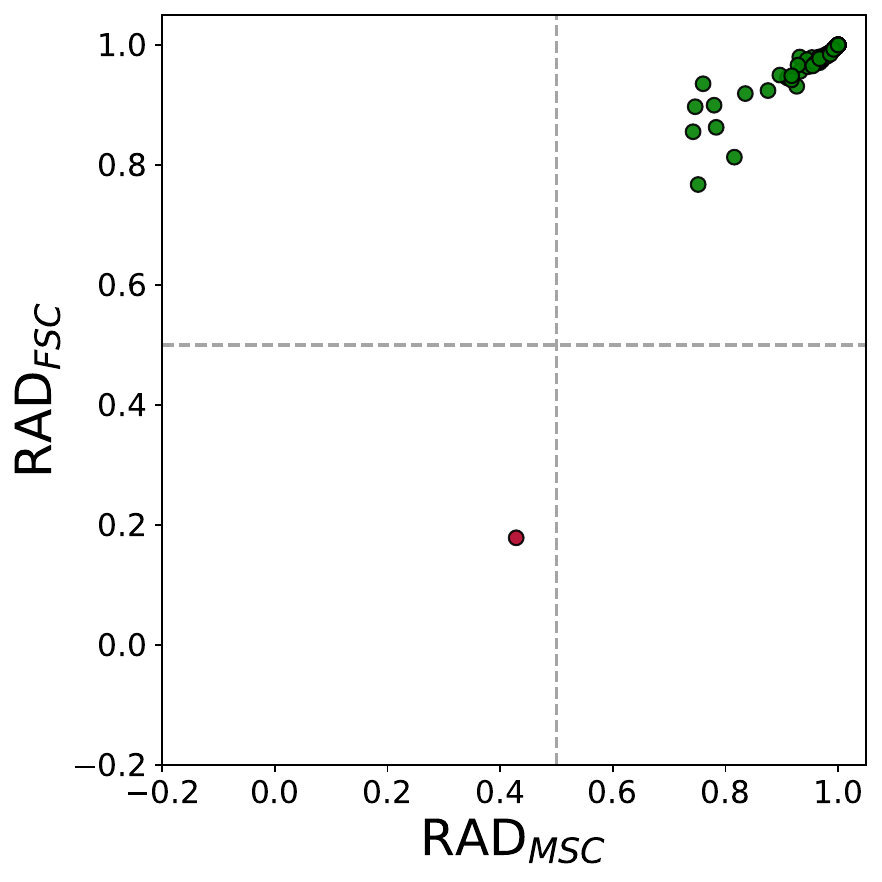}}
\subfloat[\textit{Class 7}]{\includegraphics[width=0.2\textwidth]{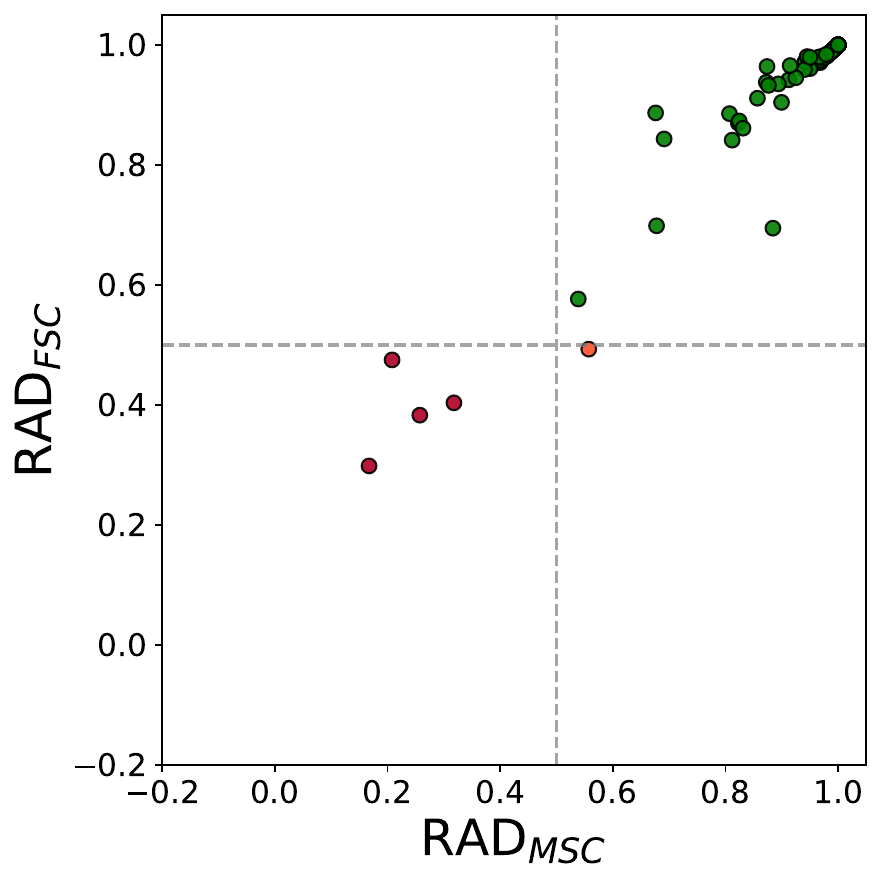}}
\subfloat[\textit{Class 8}]{\includegraphics[width=0.2\textwidth]{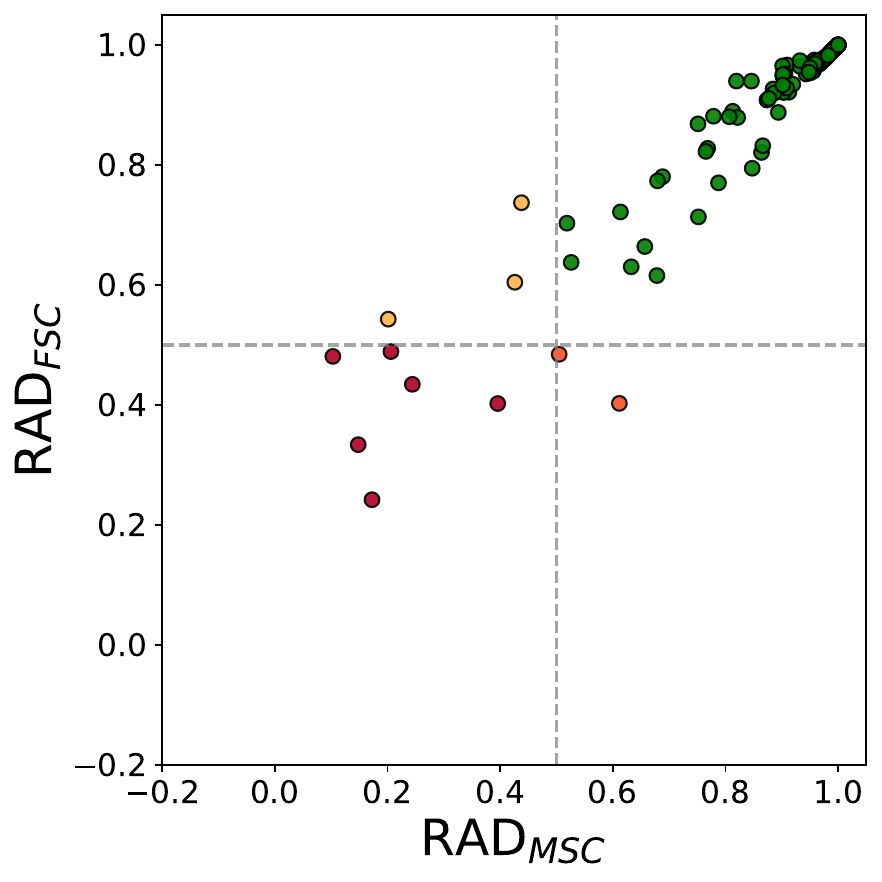}}
\subfloat[\textit{Class 9}]{\includegraphics[width=0.2\textwidth]{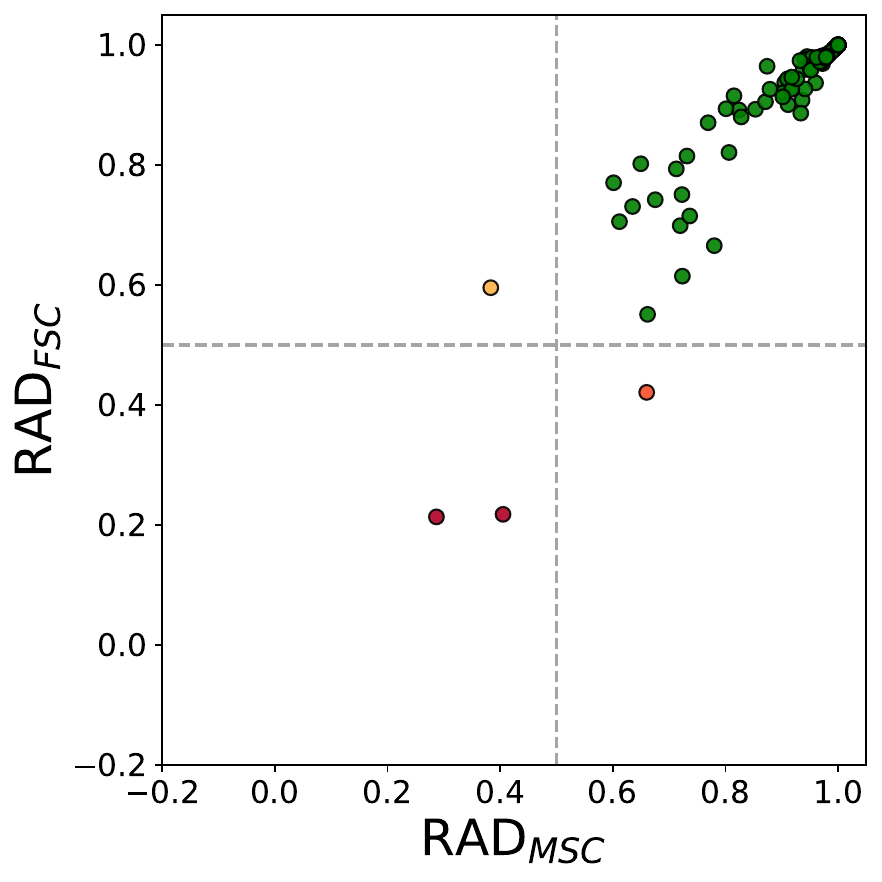}}

\caption{UCI Multi-class dataset \textit{Optical  Digit Recognition} RAD Plots.
}
\label{fig:80_ODR}
\end{figure}

\begin{figure}[htbp]
\centering

\subfloat[\textit{Class 0}]{\includegraphics[width=0.2\textwidth]{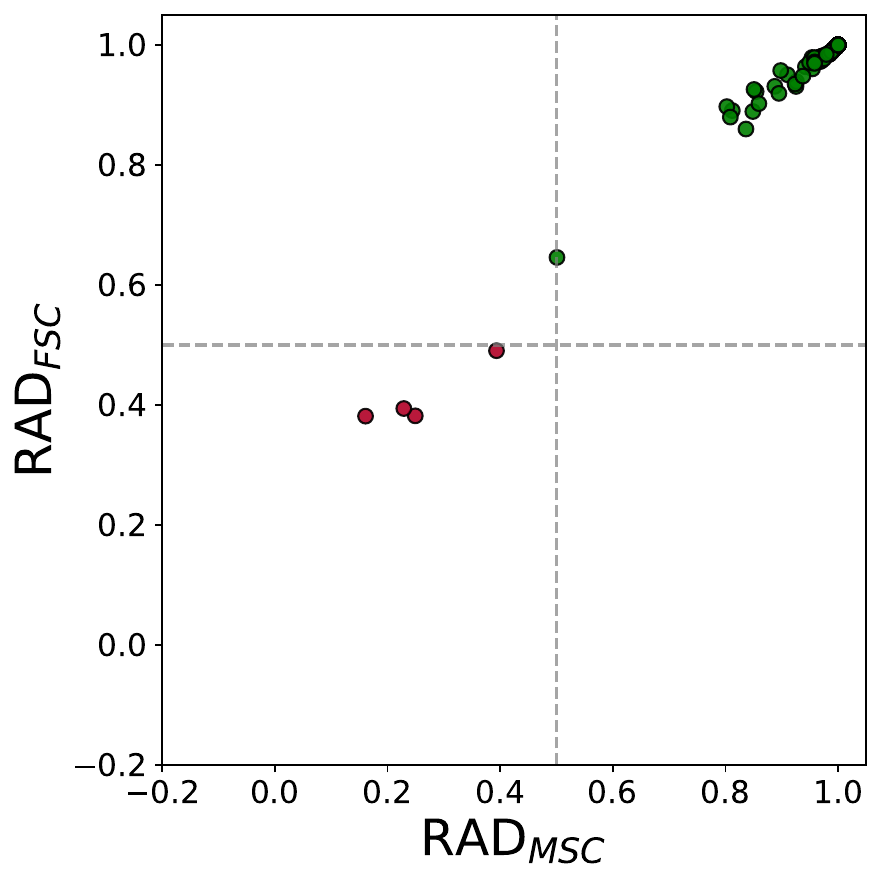}}
\subfloat[\textit{Class 1}]{\includegraphics[width=0.2\textwidth]{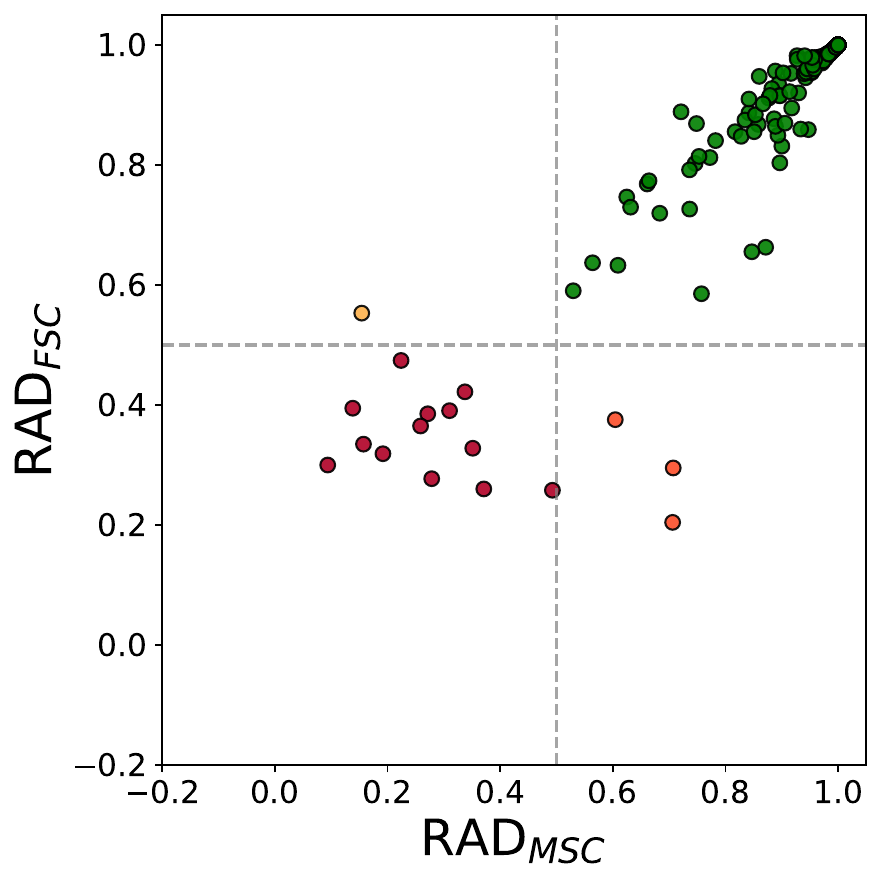}}
\subfloat[\textit{Class 2}]{\includegraphics[width=0.2\textwidth]{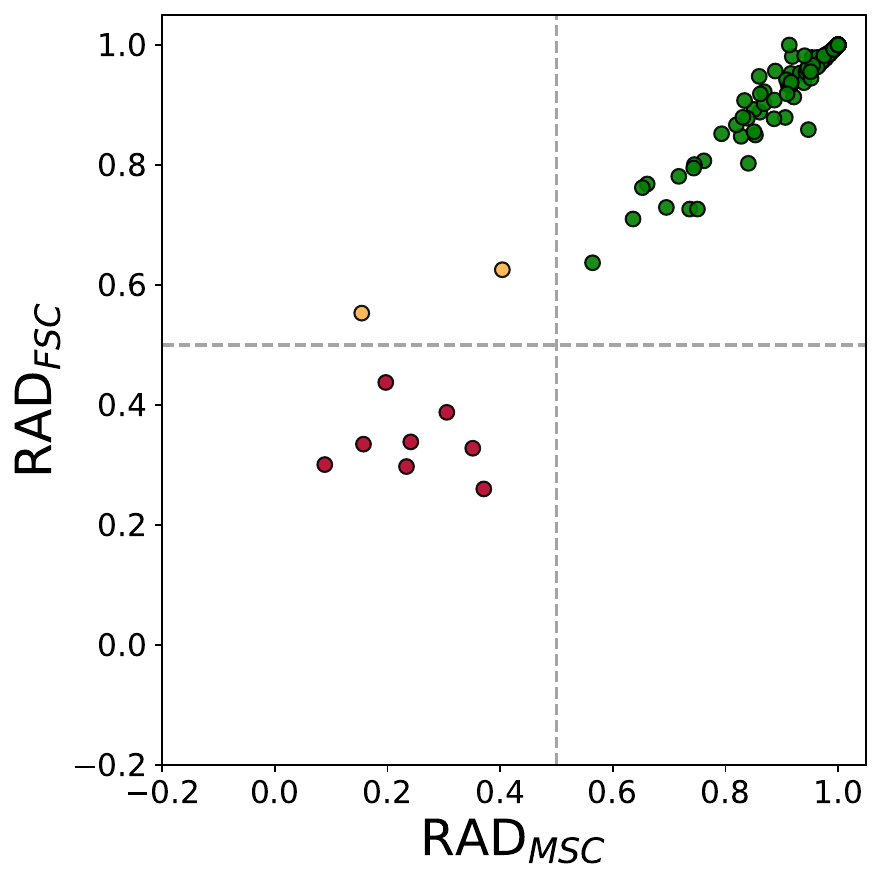}}
\subfloat[\textit{Class 3}]{\includegraphics[width=0.2\textwidth]{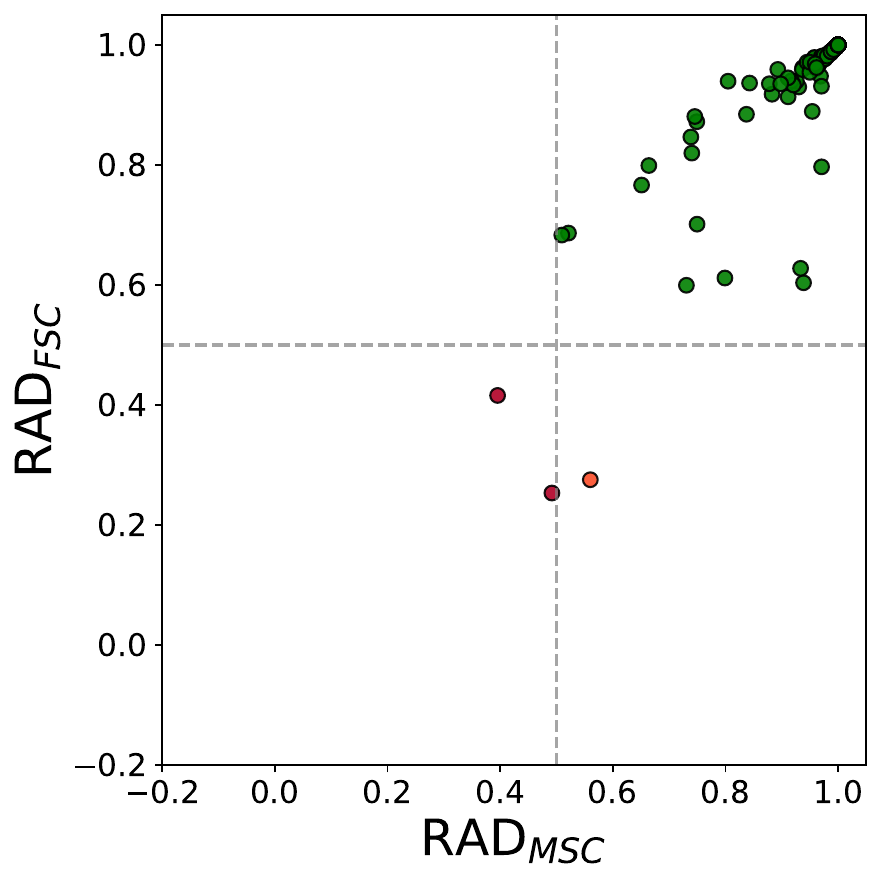}}
\subfloat[\textit{Class 4}]{\includegraphics[width=0.2\textwidth]{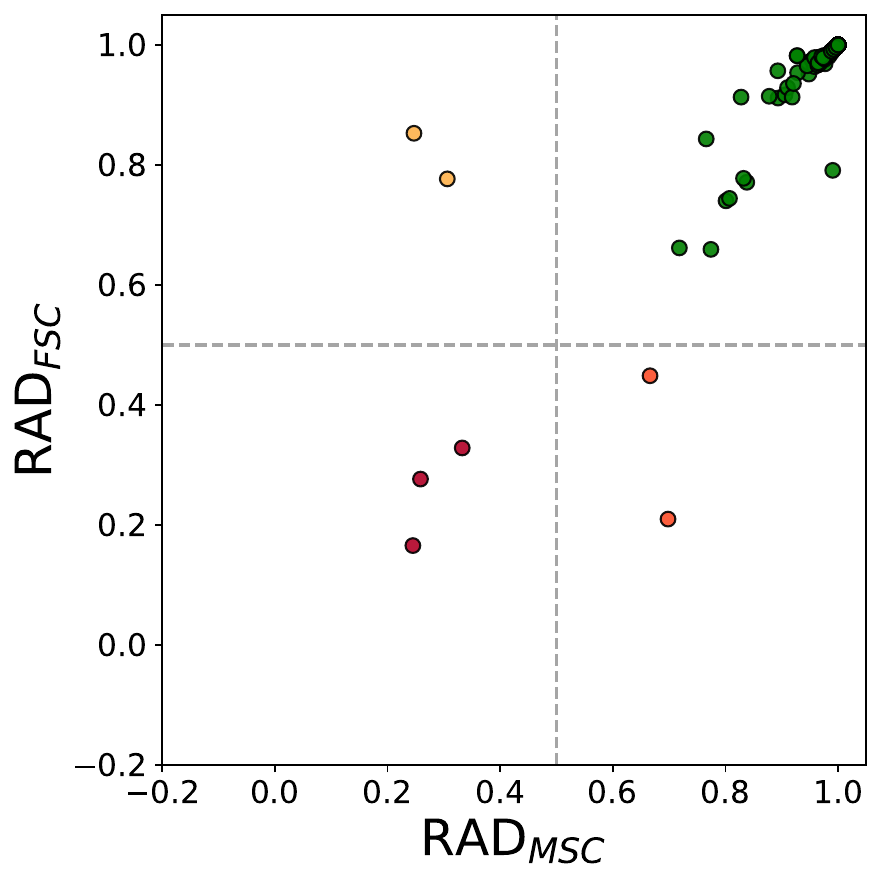}}

\subfloat[\textit{Class 5}]{\includegraphics[width=0.2\textwidth]{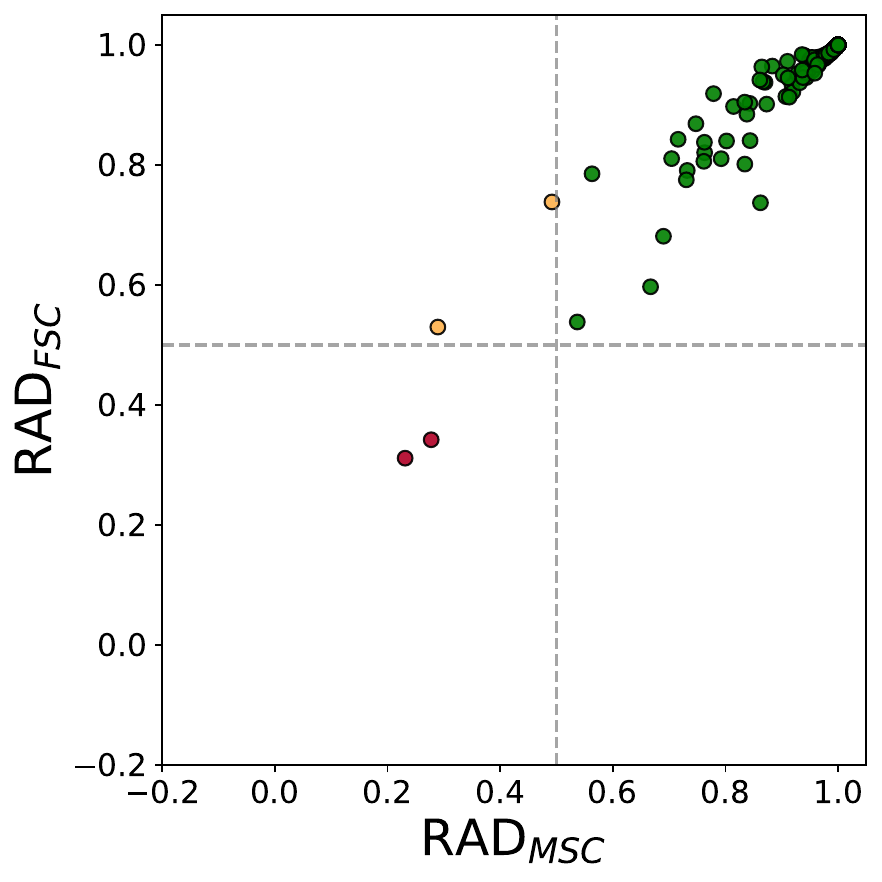}}
\subfloat[\textit{Class 6}]{\includegraphics[width=0.2\textwidth]{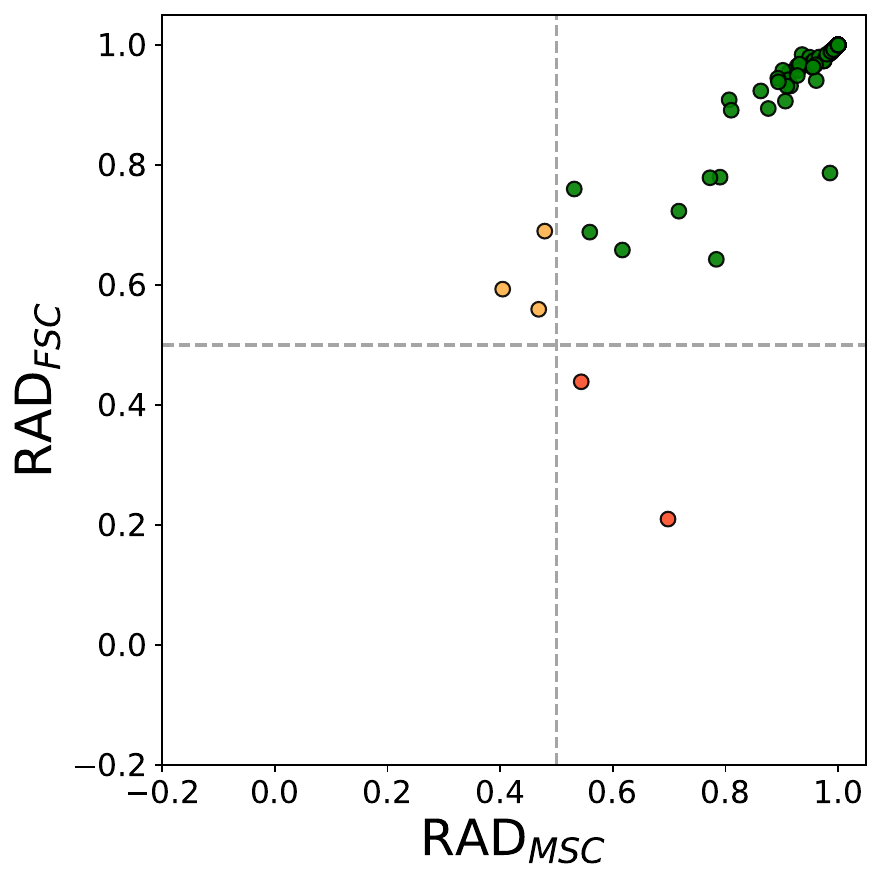}}
\subfloat[\textit{Class 7}]{\includegraphics[width=0.2\textwidth]{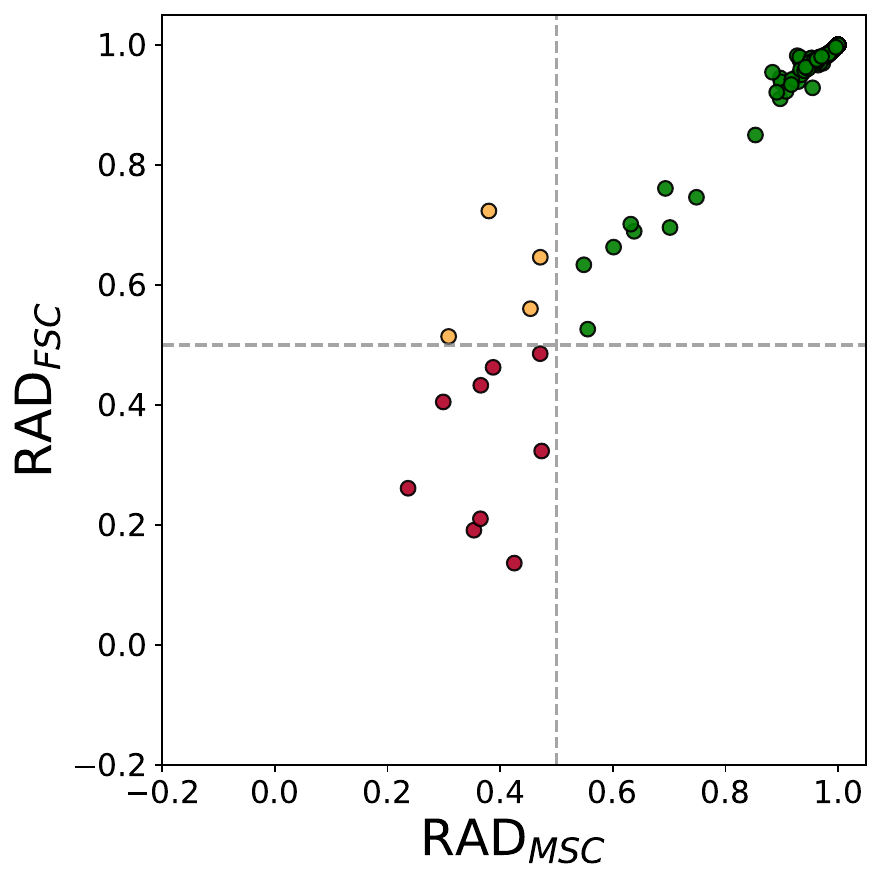}}
\subfloat[\textit{Class 8}]{\includegraphics[width=0.2\textwidth]{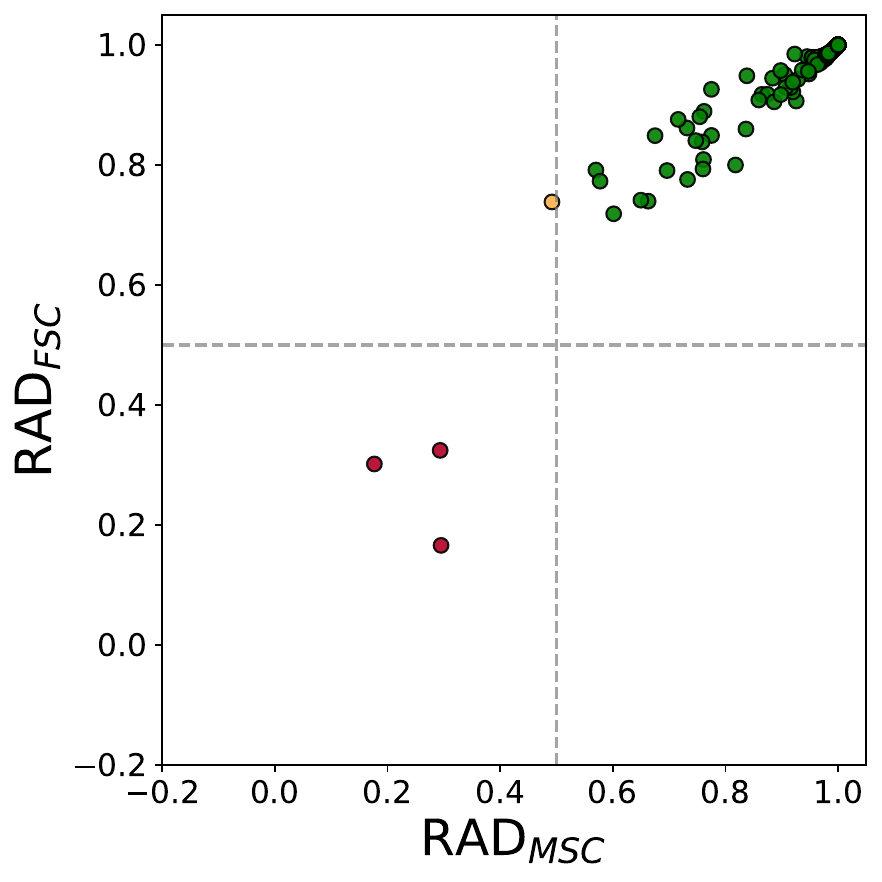}}
\subfloat[\textit{Class 9}]{\includegraphics[width=0.2\textwidth]{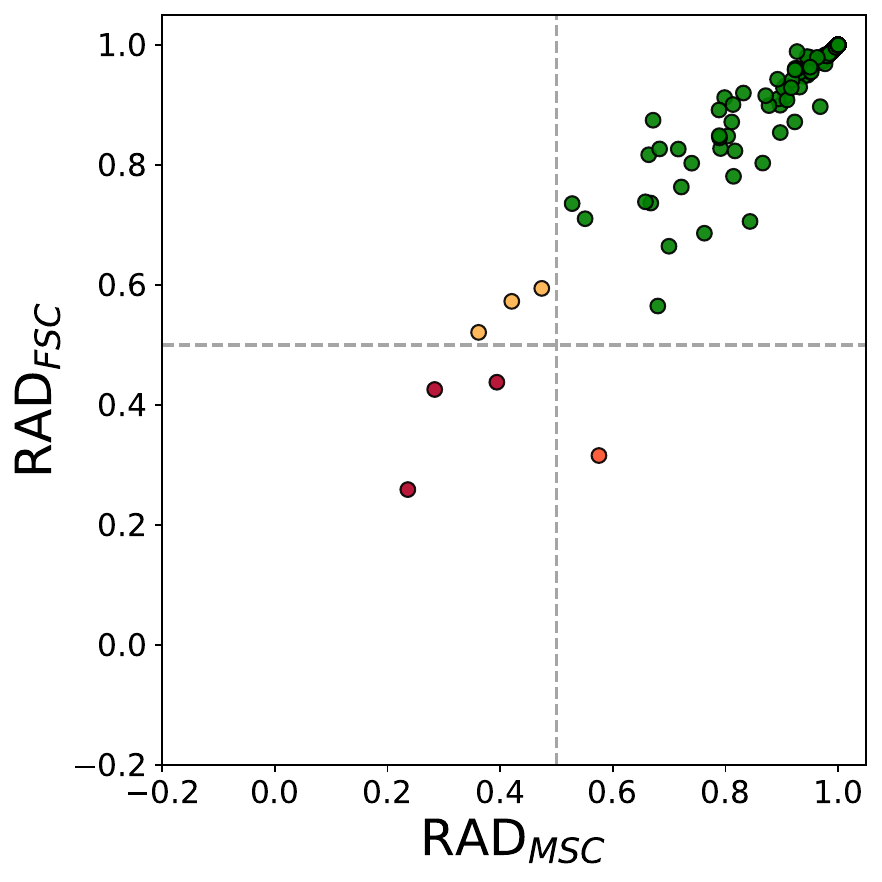}}

\caption{UCI Multi-class dataset \textit{Handwritten Digit Recognition} RAD Plots.
}
\label{fig:81_ODR}
\end{figure}

\begin{figure}[htbp]
\centering

\subfloat[\textit{Class 0}]{\includegraphics[width=0.2\textwidth]{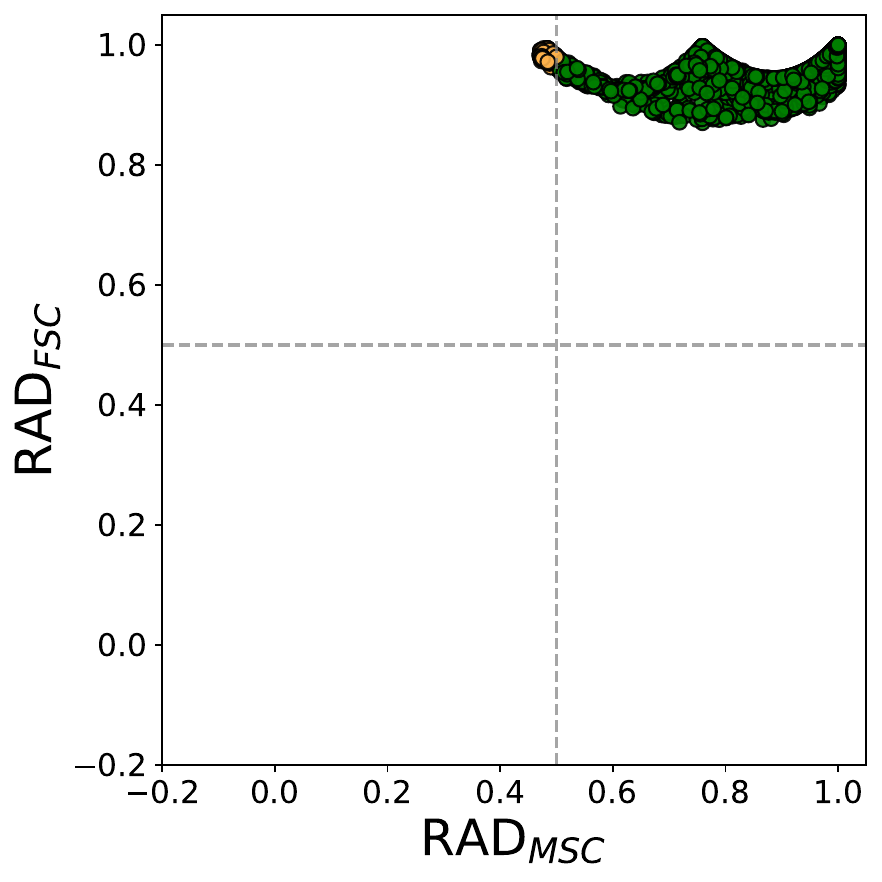}}
\subfloat[\textit{Class 1}]{\includegraphics[width=0.2\textwidth]{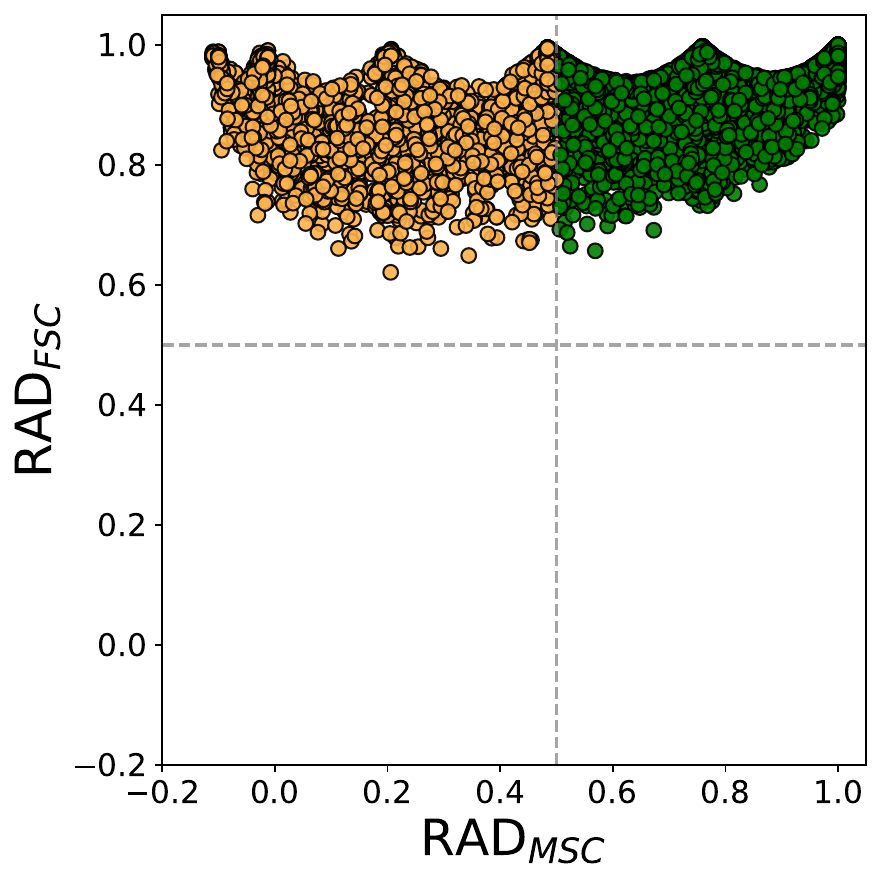}}
\subfloat[\textit{Class 2}]{\includegraphics[width=0.2\textwidth]{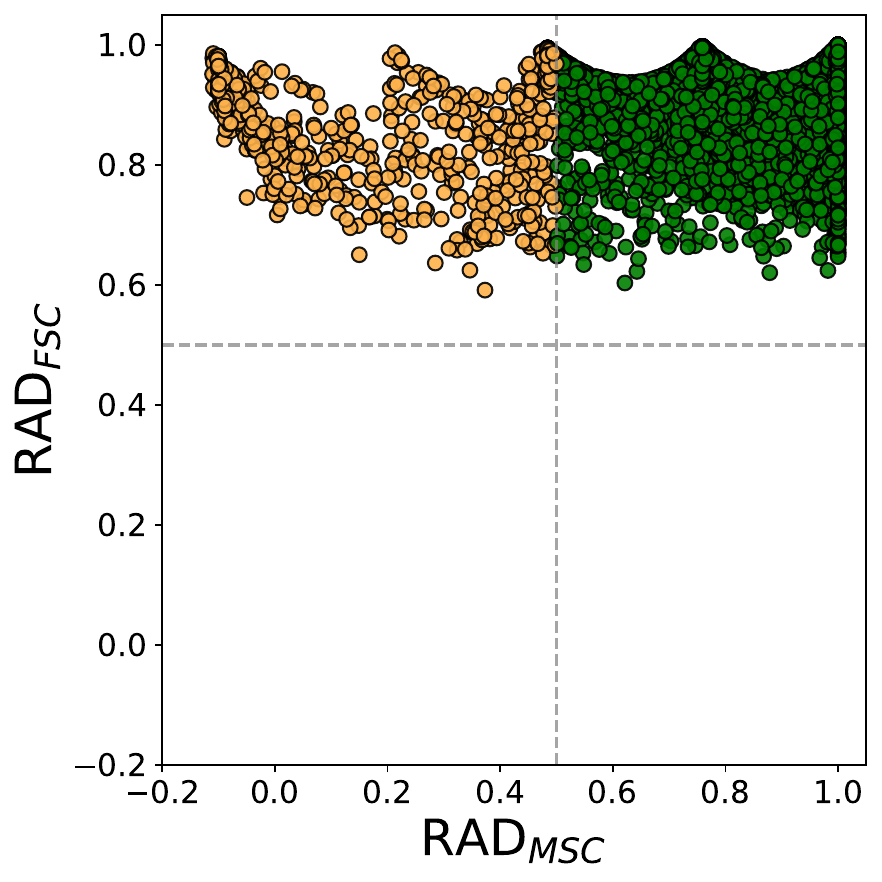}}
\subfloat[\textit{Class 3}]{\includegraphics[width=0.2\textwidth]{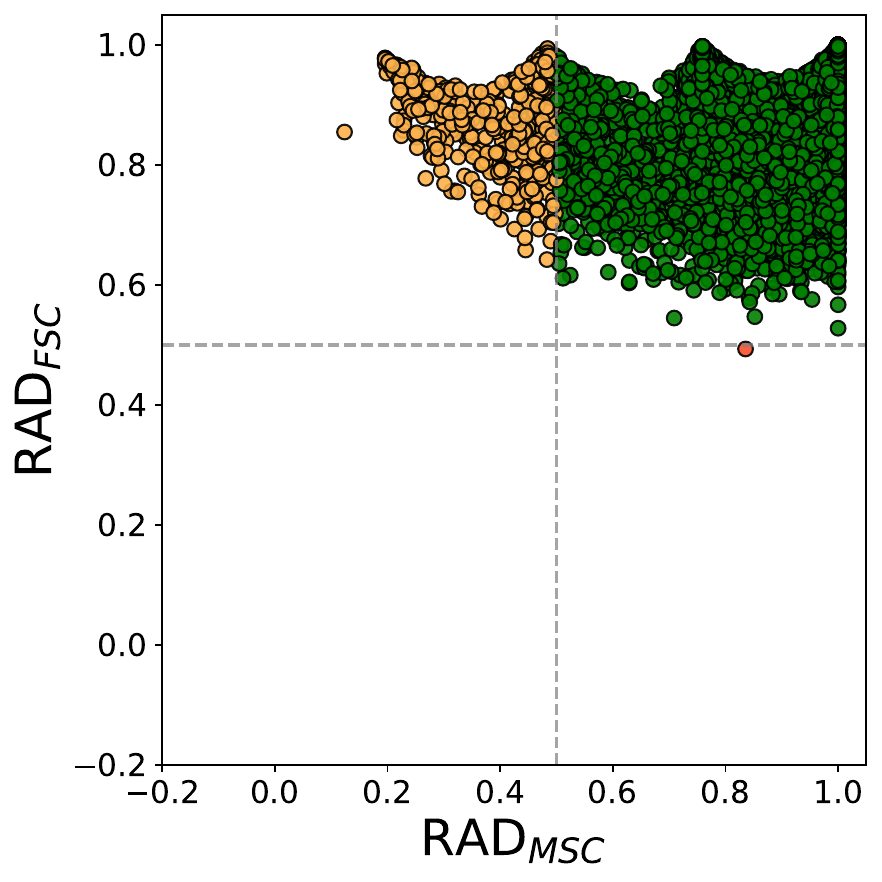}}
\subfloat[\textit{Class 4}]{\includegraphics[width=0.2\textwidth]{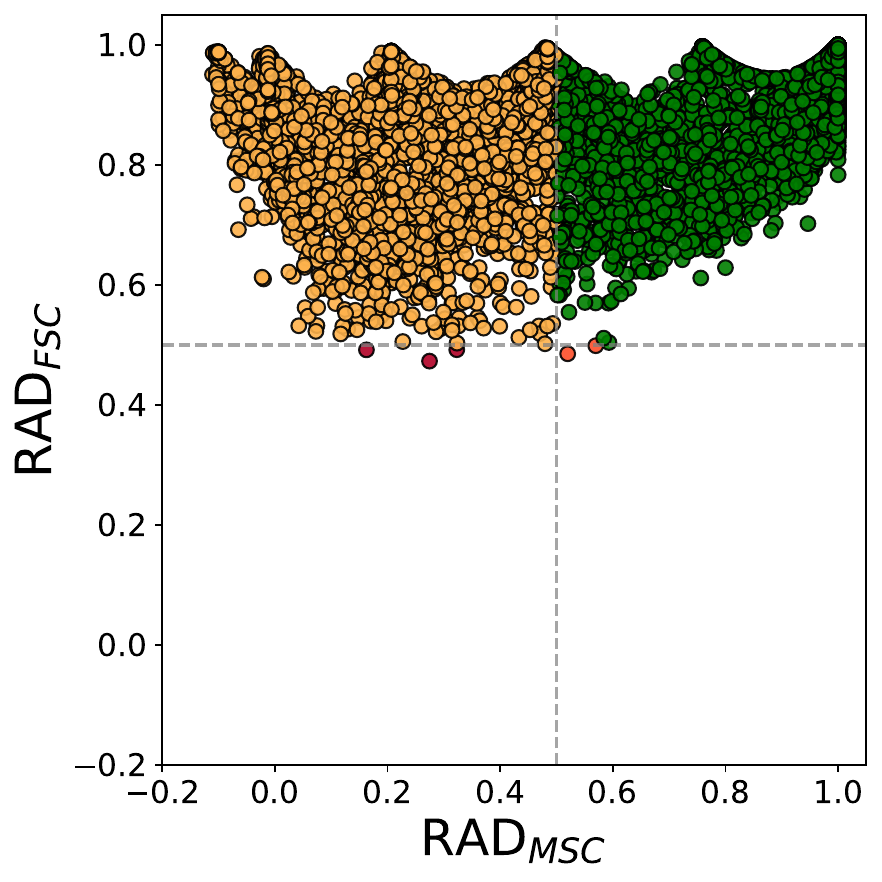}}

\subfloat[\textit{Class 5}]{\includegraphics[width=0.2\textwidth]{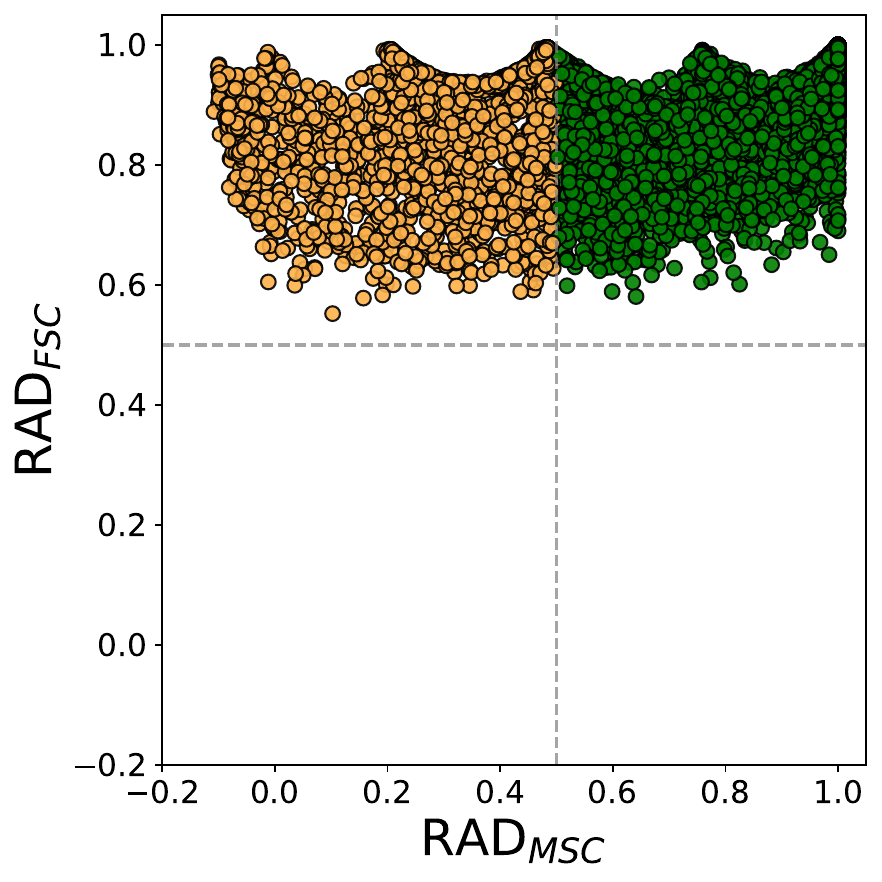}}
\subfloat[\textit{Class 6}]{\includegraphics[width=0.2\textwidth]{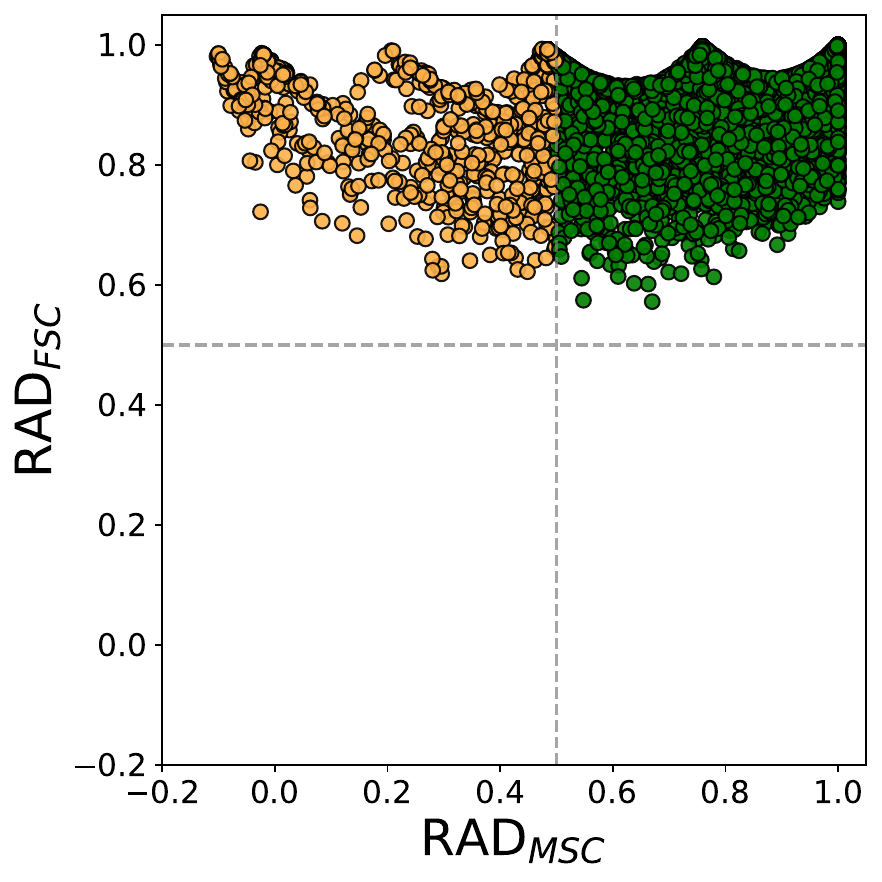}}
\subfloat[\textit{Class 7}]{\includegraphics[width=0.2\textwidth]{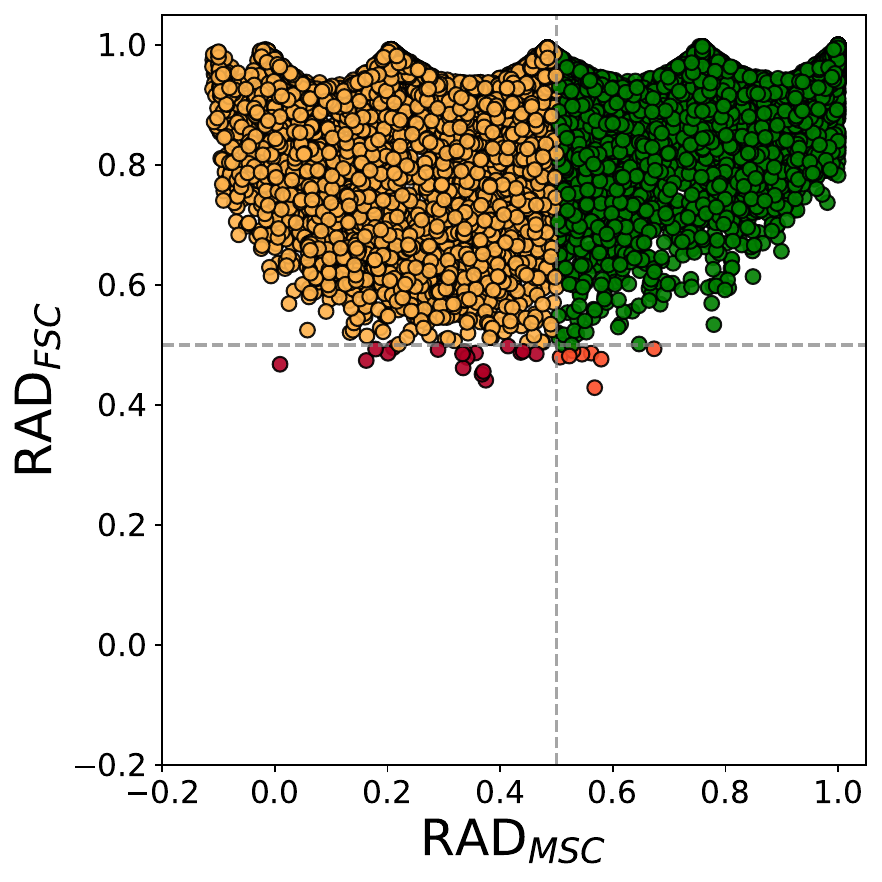}}
\subfloat[\textit{Class 8}]{\includegraphics[width=0.2\textwidth]{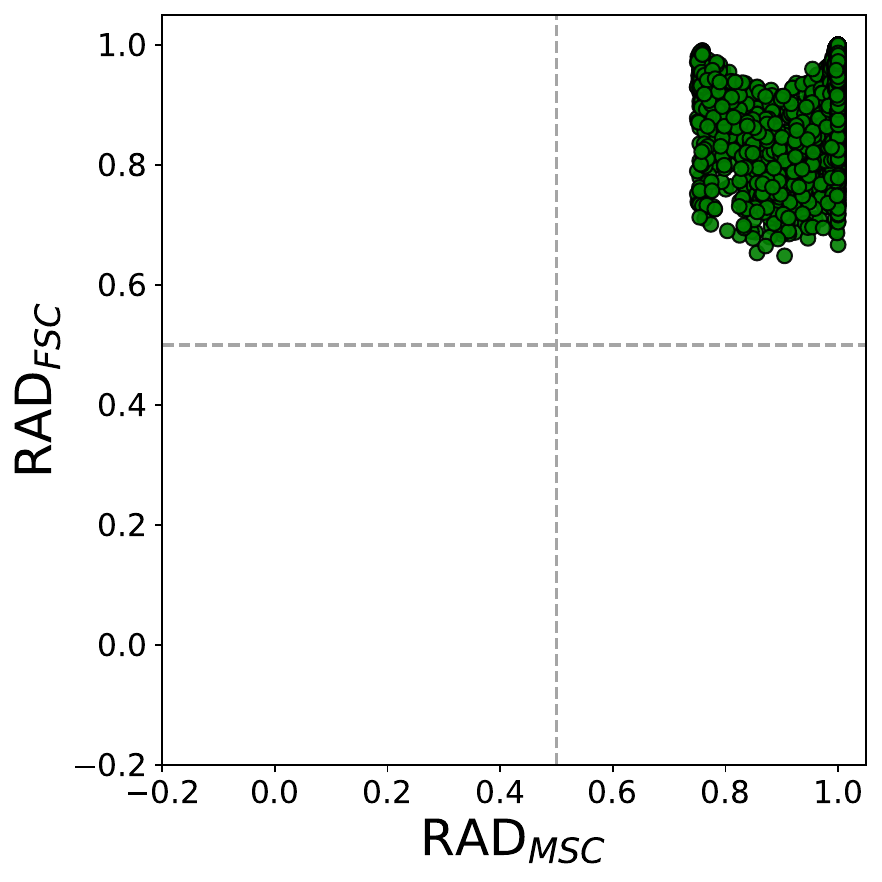}}
\subfloat[\textit{Class 9}]{\includegraphics[width=0.2\textwidth]{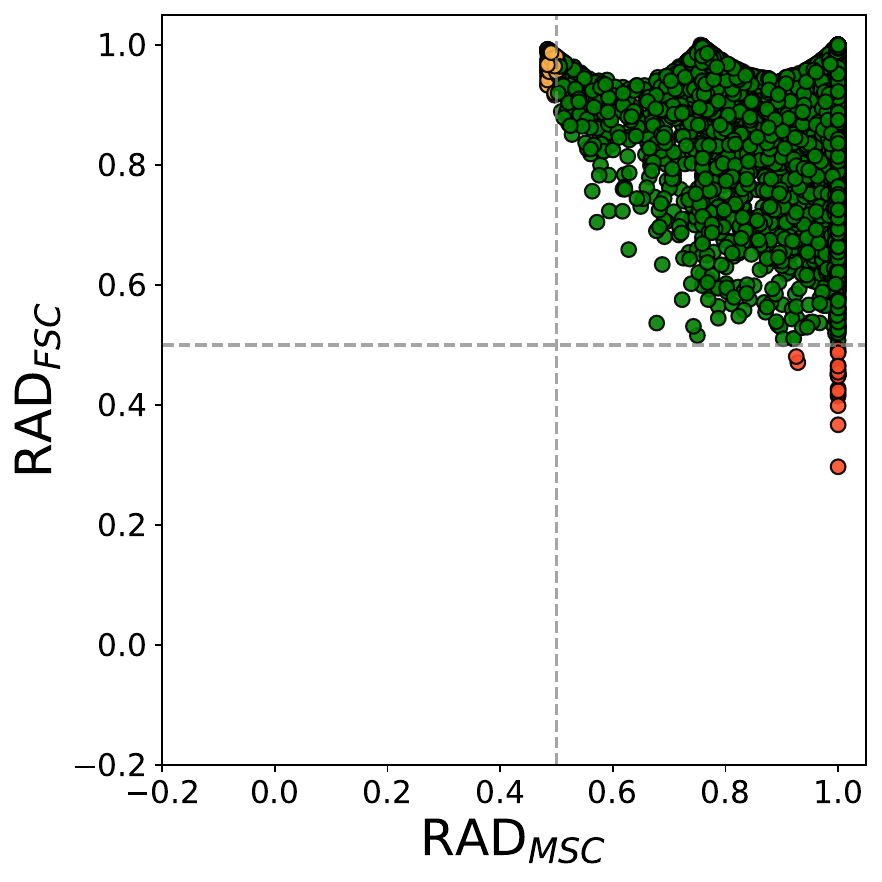}}

\caption{\textit{MNIST} RAD Plots.
}
\label{fig:mnist}
\end{figure}

\begin{landscape}
\subsection{Abstention Results}
\small
\setlength{\tabcolsep}{4pt}
\begin{longtable}{@{}lllll|lll@{}}
\caption{Rejection performance measured via AUC of accuracy vs.\ coverage (AURC).
Values in parentheses denote improvement over the random rejection baseline (\%).
Synthetic datasets are grouped into Low/Medium/High difficulty levels based on
ascending parameter values. The vertical bar separates baseline rejection methods
(left) from the RAD individual components and family (right). \textbf{Bold} indicates the best score on each row across all comparison methods (Random excluded as baseline, all method ties are not made bold). RAD Scores and RAD-Pareto is best or tied for best on 28 of 32 datasets, and performs better compared to Self-Consistency on 17 of the 22 non-tied rows.}
\label{tab:abstention}\\
\toprule
Dataset & Setting & Random & Entropy & Self-Consistency
        & $\mathrm{RAD}_{\mathrm{MSC}}$
        & $\mathrm{RAD}_{\mathrm{FSC}}$
        & RAD-Pareto \\
\midrule
\endfirsthead

\multicolumn{8}{@{}l}{\textit{(continued from previous page)}}\\
\toprule
Dataset & Setting & Random & Entropy & Self-Consistency
        & $\mathrm{RAD}_{\mathrm{MSC}}$
        & $\mathrm{RAD}_{\mathrm{FSC}}$
        & RAD-Pareto \\
\midrule
\endhead

\midrule
\multicolumn{8}{r@{}}{\textit{continued on next page}}\\
\endfoot

\bottomrule
\endlastfoot

\multicolumn{8}{@{}l}{\textit{Synthetic datasets (controlled overlap)}} \\
\midrule
\multirow{3}{*}{Blobs}
  & Low & 1.0000 & 1.0000 (0.00) & 1.0000 (0.00) & 1.0000 (0.00) & \textbf{1.000} (0.00) & 1.0000 (0.00) \\
  & Medium & 1.0000 & 1.0000 (0.00) & 1.0000 (0.00) & 1.0000 (0.00) & 1.0000 (0.00) & 1.0000 (0.00) \\
  & High & 0.8013 & 0.8227 (2.68) & 0.8743 (9.11) & 0.8754 (9.25) & 0.8833 (10.25) & \textbf{0.8903 (11.11)} \\
\midrule
\multirow{3}{*}{Circles}
  & Low & 1.0000 & 1.0000 (0.00) & 1.0000 (0.00) & 1.0000 (0.00) & 1.0000 (0.00) & 1.0000 (0.00) \\
  & Medium & 0.9869 & 0.9869 (0.01) & 0.9867 (-0.01) & 0.9994 (1.27) & \textbf{0.9997 (1.30)} & 0.9996 (1.30) \\
  & High & 0.8353 & 0.8968 (7.37) & 0.9030 (8.11) & 0.9324 (11.62) & 0.9326 (11.66) & \textbf{0.9362 (12.09)} \\
\midrule
\multirow{3}{*}{Moons}
  & Low & 1.0000 & 1.0000 (0.00) & 1.0000 (0.00) & 1.0000 (0.00) & 1.0000 (0.00) & 1.0000 (0.00) \\
  & Medium & 0.9892 & 0.9927 (0.36) & \textbf{1.0000 (1.10)} & \textbf{1.0000 (1.10)} & \textbf{1.0000 (1.10)} & \textbf{1.0000 (1.10)} \\
  & High & 0.9367 & 0.9906 (5.76) & \textbf{0.9959 (6.33)} & 0.9956 (6.29) & 0.9936 (6.09) & 0.9948 (6.21) \\
\midrule
\multirow{3}{*}{Spirals}
  & Low & 0.9918 & 0.9930 (0.12) & 1.0000 (0.82) & 0.9995 (0.78) & \textbf{1.0000 (0.83)} & 0.9999 (0.82) \\
  & Medium & 0.9784 & 0.9798 (0.14) & 0.9986 (2.06) & 0.9995 (2.16) & \textbf{0.9997 (2.18)} & 0.9996 (2.16) \\
  & High & 0.9652 & 0.9767 (1.20) & 0.9975 (3.35) & 0.997 (3.29) & \textbf{0.9982 (3.42)} & 0.9981 (3.42) \\
\midrule
\multirow{3}{*}{Checkerboard}
  & Low & 0.9948 & 0.9995 (0.48) & \textbf{1.0000 (0.53)} & 1.0000 (0.53) & 0.9999 (0.52) & 0.9999 (0.52) \\
  & Medium & 0.8970 & 0.9591 (6.93) & 0.9689 (8.03) & 0.971 (8.26) & 0.9404 (4.84) & \textbf{0.9719 (8.35)} \\
  & High & 0.7345 & 0.7817 (6.43) & 0.7986 (8.73) & \textbf{0.8582 (16.85)} & 0.8553 (16.45) & 0.8569 (16.67) \\

\pagebreak

\multicolumn{8}{@{}l}{\textit{Real-world datasets (Multi-Class (MC) / Binary classification (Bin) settings)}} \\
\midrule
Magic Gamma Telescope               & Bin  & 0.8256 & 0.9005 (9.07) & 0.9281 (12.42) & 0.9330 (13.01) & 0.9275 (12.34) & \textbf{0.9344 (13.18)} \\
Default of Credit Card Clients      & Bin  & 0.8224 & 0.8237 (0.17) & 0.8135 (-1.07) & 0.8135 (-1.07) & 0.8411 (2.28) & \textbf{0.8573 (4.24)} \\
Rice                                & Bin  & 0.9170 & 0.9120 (-0.53) & 0.9222 (0.57) & 0.9848 (7.41) & 0.9870 (7.65) & \textbf{0.9875 (7.70)} \\
Banknote Authentication             & Bin  & 0.9943 & 0.9929 (-0.14) & 0.9938 (-0.05) & 0.9996 (0.53) & \textbf{0.9998 (0.55)} & 0.9997 (0.55) \\
Heart Failure Risk                  & Bin  & 0.8351 & 0.7982 (-4.41) & 0.8061 (-3.47) & 0.8202 (-1.78) & 0.8724 (4.48) & \textbf{0.9011 (7.91)} \\
Mammographic Mass                   & Bin  & 0.8421 & 0.8669 (2.95) & 0.9105 (8.12) & \textbf{0.9194 (9.18)} & 0.9028 (7.21) & 0.8827 (4.82) \\
Statlog (Landsat Satellite)         & MC   & 0.8550 & 0.8413 (-1.60) & 0.9630 (12.64) & 0.9689 (13.33) & \textbf{0.9702 (13.49)} & 0.9696 (13.41) \\
Glass Identification                & MC   & 0.7293 & 0.7415 (1.68) & 0.7295 (0.03) & 0.8241 (13.01) & \textbf{0.8371 (14.79)} & 0.8305 (13.88) \\
E Coli                              & MC   & 0.7968 & 0.7425 (-6.81) & 0.9434 (18.41) & \textbf{0.9556 (19.94)} & 0.9517 (19.45) & 0.9542 (19.77) \\
Iris                                & MC   & 0.9263 & 0.9941 (7.32) & 0.9804 (5.84) & 0.9966 (7.59) & 0.9927 (7.18) & \textbf{0.9966 (7.60)} \\
Optical Digit Recognition           & MC   & 0.7521 & 0.8715 (15.88) & 0.9275 (23.32) & \textbf{0.9311 (23.80)} & 0.9244 (22.92) & 0.9301 (23.67) \\
Handwritten Digit Recognition       & MC   & 0.9422 & 0.9359 (-0.67) & \textbf{0.9924 (5.33)} & 0.9913 (5.22) & 0.9911 (5.19) & 0.9913 (5.21) \\
Heart Disease                       & MC   & 0.5390 & 0.6652 (23.42) & \textbf{0.8199 (52.14)} & 0.8062 (49.59) & 0.7893 (46.45) & 0.8110 (50.47) \\
Breast Cancer (Wisconsin)           & MC   & 0.9331 & 0.9067 (-2.83) & \textbf{0.9923 (6.35)} & 0.9918 (6.30) & 0.9908 (6.19) & 0.9912 (6.24) \\
COMPAS                              & MC   & 0.9910 & 0.9991 (0.82) & 0.9916 (0.06) & 0.9916 (0.06) & 0.9995 (0.86) & \textbf{0.9995 (0.86)} \\
MNIST                               & MC   & 0.8622 & 0.8034 (-6.81) & \textbf{0.9652 (11.96)} & 0.8788 (1.93) & 0.9236 (7.13) & 0.9095 (5.49) \\
Wine Quality                        & MC   & 0.5913 & 0.5977 (1.10) & \textbf{0.7803 (31.97)} & 0.7307 (23.58) & 0.6888 (16.49) & 0.7105 (20.17) \\
\end{longtable}
\end{landscape}
\end{document}